\RequirePackage{fix-cm}
\documentclass[twocolumn]{svjour3}          
\smartqed  
\usepackage{graphicx}
\usepackage{textcomp}
\usepackage{natbib} 
\usepackage[colorlinks,linkcolor=black,anchorcolor=black,citecolor=black,CJKbookmarks=True]{hyperref}
\usepackage{booktabs}   %
\usepackage{multirow,tabularx}  %
\usepackage{diagbox}
\usepackage{bm}  %
\usepackage[caption=false,font=footnotesize]{subfig}  %
\usepackage[usenames,dvipsnames]{color}
\usepackage{colortbl}
\definecolor{darkgray}{gray}{.8}
\definecolor{lightgray}{gray}{.9}
\definecolor{electricpurple}{rgb}{0.75, 0.0, 1.0}

 \journalname{International Journal of Computer Vision}
\begin{document}

\title{Unified Quality Assessment of In-the-Wild Videos with Mixed Datasets Training
}

\author{Dingquan Li$^{1,2,4,5}$ \and 
        Tingting Jiang$^{1,3,6}$ \and
        Ming Jiang$^{4}$        
}

\authorrunning{Dingquan Li et al.} 

\institute{Dingquan Li \at
           \email{dingquanli@pku.edu.cn}    
           \and
           Tingting Jiang \at
           \email{ttjiang@pku.edu.cn (Corresponding Author)}           
           \and
           Ming Jiang \at
           \email{ming-jiang@pku.edu.cn}  
           \and 
           \at
              $^1$ National Engineering Laboratory for Video Technology, Peking University, Beijing, China \\
              $^2$ Advanced Institute of Information Technology, Peking University, Hangzhou, China \\
              $^3$ Department of Computer Science, Peking University, Beijing, China \\
              $^4$ Laboratory of Mathematics and Its Applications, School of Mathematical Sciences, Peking University, Beijing, China \\  
              $^5$ Beijing International Center for Mathematical Research, Peking University, Beijing, China \\
              $^6$ Advanced Innovation Center for Future Visual Entertainment, Beijing Film Academy, Beijing, China
}

\date{Received: 20 December 2019  / Accepted: 4 November 2020}

\maketitle

\begin{abstract}
Video quality assessment (VQA) is an important problem in computer vision.
The videos in computer vision applications are usually captured in the wild.
We focus on automatically assessing the quality of in-the-wild videos, which is a challenging problem due to the absence of reference videos, the complexity of distortions, and the diversity of video contents. 
Moreover, the video contents and distortions among existing datasets are quite different, which leads to poor performance of data-driven methods in the cross-dataset evaluation setting.
To improve the performance of quality assessment models, we borrow intuitions from human perception, specifically, content dependency and temporal-memory effects of human visual system.
To face the cross-dataset evaluation challenge, we explore a mixed datasets training strategy for training a single VQA model with multiple datasets.
The proposed unified framework explicitly includes three stages: relative quality assessor, nonlinear mapping, and dataset-specific perceptual scale alignment, to jointly predict relative quality, perceptual quality, and subjective quality.
Experiments are conducted on four publicly available datasets for VQA in the wild, \textit{i.e.}, LIVE-VQC, LIVE-Qualcomm, KoNViD-1k, and CVD2014. 
The experimental results verify the effectiveness of the mixed datasets training strategy and prove the superior performance of the unified model in comparison with the state-of-the-art models.
For reproducible research, we make the PyTorch implementation of our method available at \url{https://github.com/lidq92/MDTVSFA}.
\end{abstract}

\section{Introduction}
\label{sec:introduction}

\begin{figure*}
\begin{center}
  \subfloat[Three representative frames of the video on CVD2014~\citep{nuutinen2016cvd2014} with the worst quality]{\includegraphics[width=.86\linewidth]{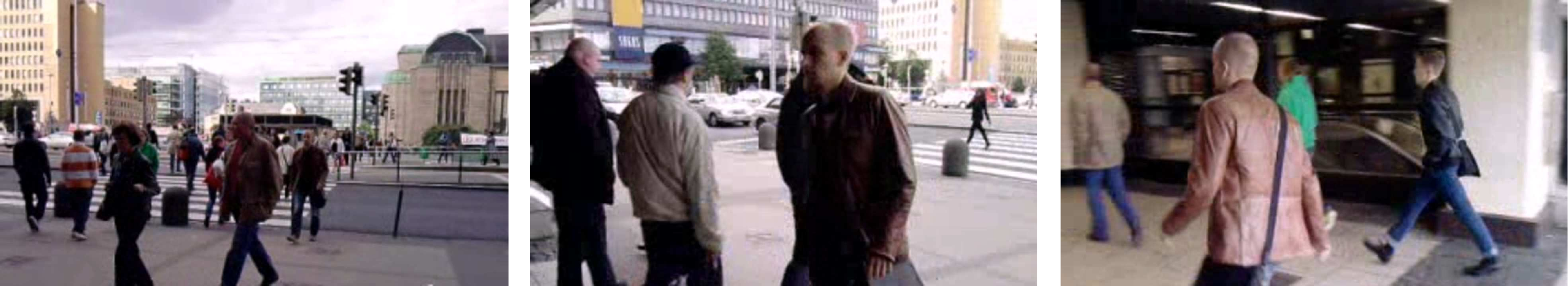}
  \label{fig:W2}}
\hfill
 \subfloat[Three representative frames of the video on LIVE-VQC~\citep{sinno2019large} with the worst quality]{\includegraphics[width=.86\linewidth]{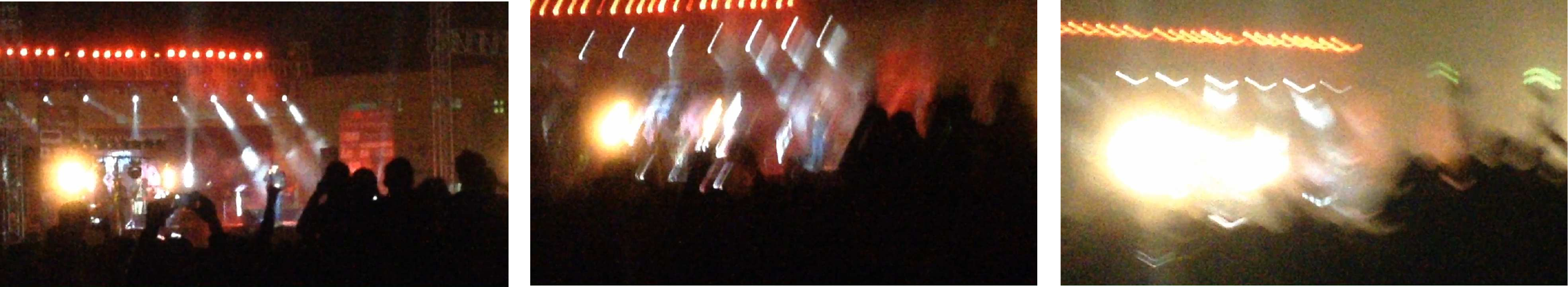}
 \label{fig:W4}}
\end{center}
  \caption{An illustration of the videos with the worst quality on CVD2014 and LIVE-VQC, respectively (Full videos are provided in \url{https://bit.ly/3csmHYk}). The upper video has a better quality in comparison with the lower video. However, linear re-scaling leads to the same quality labels for them. Such ``inconformity" will disturb the training process, and lead to a poor performance.} 
 \label{fig:scale}
\end{figure*}

Image/Video quality assessment (I/VQA) is a fundamental and longstanding problem in the image processing and computer vision community.
It is involved in benchmarking and optimizing many vision applications, such as image classification~\citep{dodge2016understanding}, object tracking~\citep{nieto2019video}, video compression~\citep{rippel2019learned}, image inpainting~\citep{isogawa2019better}, and super resolution~\citep{zhang2019ranksrgan}. 
Because of its importance, I/VQA has attracted significant attention in the past two decades~\citep{wang2004image,mittal2012no,zhang2014vsi,kang2014convolutional,ma2016group,liu2017rankiqa,kim2018deep,lin2018hallucinated}. 
Videos obtained in the wild are often in low-quality because of many factors, such as  out of focus, object motion, camera shakiness, under-/over- exposure, and adverse weather, etc. 
With the guidance of VQA in the wild, one can automatically identify, cull, repair or enhance low-quality videos before sending them to the subsequent vision applications, so that the applications can work in the real scenario.
Thus, VQA in the wild is necessary for computer vision in the wild, but few attention is paid to this task.

VQA in the wild is a challenging task for the reason that the pristine videos are not available, the distortions are complex, and the contents are diverse. 
Compared to synthetically-distorted videos, in-the-wild videos contain huge amount of contents and may be infected with mixed real-world distortions, especially some of which are temporally heterogeneous (\textit{e.g.}, temporary auto-focus blurs and exposure adjustments).
Consequently, modern advanced I/VQA methods, \textit{e.g.}, BRISQUE~\citep{mittal2012no} and VBLIINDS~\citep{saad2014blind}, validated on synthetically-distorted video datasets~\citep{seshadrinathan2010study,moorthy2012video}, do a poor job in predicting the quality of in-the-wild videos~\citep{men2017empirical,ghadiyaram2018capture,nuutinen2016cvd2014,sinno2019large} (See Table~\ref{tab:overall performance} and Table~\ref{tab:performance}). 

Some efforts have been made to generate a better feature for VQA in the wild~\citep{you2019deep,korhonen2019two,li2019quality}. 
\citet{korhonen2019two} obtains well-behaved low-complexity features for all frames and high-complexity features for representative frames, so that good quality predictions can be achieved by the support vector regression or the random forest regression.
\citet{you2019deep} learn effective spatio-temporal features with 3D convolutional neural network (3D-CNN) and predict the video quality by a long-short term memory (LSTM) network.
Our previous work~\citep{li2019quality} borrows intuitions from human visual system (HVS), which extracts content-aware and distortion-sensitive features.
Although the above mentioned methods achieve superior performances on the benchmark VQA datasets individually, their performances are poor in cross-dataset evaluation setting (See Table~\ref{tab:vs individual}).  
For example, when the model is trained on KoNViD-1k~\citep{hosu2017konstanz}, the test performance on LIVE-Qualcomm~\citep{ghadiyaram2018capture}  or CVD2014~\citep{nuutinen2016cvd2014} drops sharply~\citep{korhonen2019two}.  
This may be caused by the over-fitting problem in the training process and the discrepancy of data distribution among the datasets.

\begin{figure*}[!htb]
\begin{center}
  \includegraphics[width=.9\textwidth]{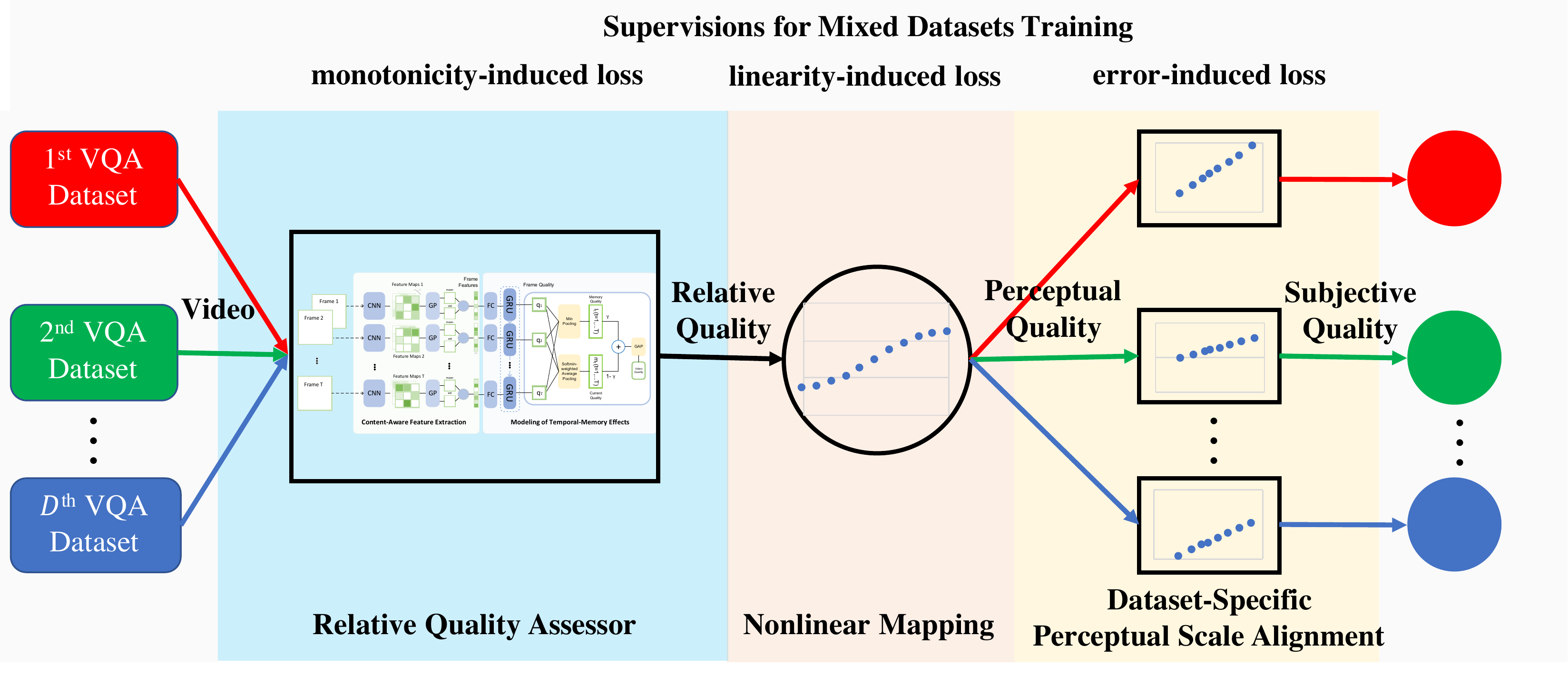}%
\end{center}
  \caption{An overview of the proposed unified framework. It consists of three stages: relative quality assessor, nonlinear mapping, and dataset-specific perceptual scale alignment for predicting relative quality, perceptual quality, and subjective quality, respectively. The supervisions for mixed datasets training at the three stages are monotonicity-induced loss, linearity-induced loss, and error-induced loss, respectively. $D$ is the number of datasets.} 
  \label{fig:framework}
\end{figure*}

To face this cross-dataset evaluation challenge, one possible solution is to mix multiple datasets during the training phase, so that the data-driven model can learn the characteristics of video contents and distortions from all these datasets. 
Mixed datasets training provides two advantages. 
First, it provides a single unified model for all datasets/applications instead of multiple models for different datasets. 
Second, it makes the utmost of existing relevant data for VQA model training, since the largest size of current in-the-wild VQA datasets is only 1200 and acquiring new annotations is time-consuming. 
However, mixed datasets training is not trivial, since the ranges of subjective quality scores among different datasets are inconsistent. 
A na\"ive strategy is the ``linear re-scaling", which maps all subjective score ranges of different datasets to the same range.
Nevertheless, the ranges of the inherent video quality among these datasets are not equal in most circumstances. 
For instance, in Fig.~\ref{fig:scale}, both the two videos are the worst in their corresponding datasets. 
The video in Fig.~\ref{fig:scale}\subref{fig:W2} has better quality in comparison with the video in Fig.~\ref{fig:scale}\subref{fig:W4}, since the latter one contains more complicated distortions, including motion blur, under-/over-exposure, and grainy noise.
However, linear re-scaling leads to the same quality labels for them. 
Such ``inconformity" will disturb the training process, thus a good performance is hard to achieve (See Fig.~\ref{fig:vs naive} and Table~\ref{tab:overall performance}).

To tackle the above inconformity problem, we should align subjective quality scores for different datasets.
One way is conducting an additional subjective study to re-align the subjective quality scores.
The other way is to learn the alignment of subjective quality scores for these datasets.
As the first way is time-consuming and impracticable when more and more datasets are considered, we choose the second one.
Before introducing our method, we first introduce three important quality concepts: perceptual quality, subjective quality, and relative quality. 
\begin{itemize}
\item[] \textbf{Perceptual quality}: Perceptual quality is an ideal concept that is related to human perception of video quality, and only if we gather all the videos in the wild and conduct the largest scale subjective study can we get the ground-truth of perceptual quality.
Perceptual quality can be used for benchmarking and optimizing video processing systems/algorithms, but its ground-truth is impossible to obtain since we cannot conduct such a large-scale subjective study on all videos in the wild.
\vspace{2mm}
\item[] \textbf{Subjective quality}: As an ``approximation'' of perceptual quality, subjective quality is considered, whose ground-truth can be accessed by conducting a subjective study on a video dataset of limited size.
Although subjective quality is designed to reflect perceptual quality, it may have different ranges for different datasets.
In terms of this fact, we can assume the subjective quality to be linearly correlated with the perceptual quality for a single dataset, but the linear transformations between subjective quality and perceptual quality are not necessarily the same for different datasets.
Subjective quality can be used as a supervised signal for the prediction of perceptual quality.
\vspace{2mm}
\item[] \textbf{Relative quality}: Compared to directly rating the quality of a video in the subjective study, it is easier for humans to choose a video with better quality from two videos.
In terms of this fact, we define the concept of relative quality, which can be accessed by ranking the quality of videos.
Relative quality can be used for benchmarking video processing algorithms.
However, due to its nonlinearity to perceptual quality, it might not be directly used for optimization.
For example, the optimization might be early stopped when the relative quality is approaching the perfect value while the perceptual quality is far from the perfect one.
\end{itemize}

With the above three concepts, we show our solution.
We decompose the VQA problem into three sub-problems, \textit{i.e.}, predicting relative quality, perceptual quality, and subjective quality in turn (See Fig.~\ref{fig:framework}).
For details, our proposed model contains three stages to solve these three sub-problems. 
First, to predict the relative quality, we use our previous HVS-inspired VQA model~\citep{li2019quality} as the backbone.
The relative quality assessor takes the video as input and outputs a relative quality score.
This stage focuses on prediction monotonicity, which describes the ability to provide the quality ranking for any list of videos that is consistent with subjective quality. 
Correspondingly, we propose a monotonicity-induced loss for this stage.
Second, to predict the perceptual quality, we adopt the well-known 4-parameter logistic function for characterizing the nonlinearity of human perception on video quality~\citep{vqeg2000fr}.
We reformulate this function and design it as a network module.
The nonlinear mapping module maps the relative quality of a video to the perceptual quality of a video.
The predicted perceptual quality is expected to be linearly correlated with the subjective quality, and we propose a linearity-induced loss as the supervision for this stage.
Third, we learn a dataset-specific perceptual scale alignment for each dataset, which tries to map the perceptual quality of a video to the subjective quality of the video on its belonging dataset.
With this dataset-specific alignment, an error-induced loss can be used as the supervision without disturbing the training. 
Under this model, we can use the above three losses for mixed datasets training to solve the ``inconformity" problem.

To verify the effectiveness of the proposed unified model with the mixed datasets training strategy, we conduct comparative experiments on four publicly available datasets for VQA in the wild, \textit{i.e.},  KoNViD-1k~\citep{hosu2017konstanz}, CVD2014~\citep{nuutinen2016cvd2014}, LIVE-Qualcomm~\citep{ghadiyaram2018capture}, and LIVE-VQC~\citep{sinno2019large}. 
Our method is compared with several modern advanced methods.
In terms of prediction monotonicity and prediction accuracy, the superior performances of our method across datasets are verified by the experimental results. 

Lastly, we highlight the relationship and difference between our previous work~\citep{li2019quality} and this work.
The model design in this work is build upon the model in our previous work.
However, there are two major differences between our previous work and this work.
First, this work focuses on model optimization with mixed datasets training while our previous work does not consider mixed datasets training.
Second, in this work, it is the first time to decompose the VQA problem into three sub-problems: predicting relative quality, perceptual quality, and subjective quality, and we propose a unified VQA framework that explicitly designs three stages to tackle these three sub-problems. 



\section{Related Work}
\label{sec:related}
This section reviews some related work. 
Sec.~\ref{sec:vqa} overviews several representative VQA methods, especially the VQA methods for in-the-wild videos. 
Sec.~\ref{sec:mdt} introduces mixed datasets training in computer vision, especially in the tasks of quality assessment.

\subsection{Video Quality Assessment}
\label{sec:vqa}
\indent\indent Classical VQA methods are grounded on different cues, such as structures~\citep{wang2004video,wang2012novel}, motion~\citep{seshadrinathan2010motion,manasa2016optical-NR}, energy~\citep{li2016spatiotemporal}, saliency~\citep{zhang2017study,you2014attention}, gradients~\citep{lu2019spatiotemporal}, or natural video statistics (NVS)~\citep{mittal2016completely,saad2014blind,sinno2019spatio}.
Besides, some VQA methods focus on the fusion of primary features~\citep{freitas2018using,li2016no1}. 
Recently, several VQA methods exploit the use of deep learning in this task~\citep{zhang2018blind,liu2018end,kim2018deep,zhang2019objective}.
\citet{kim2018deep} obtain the spatio-temporal sensitivity maps by a CNN model. 
\citet{liu2018end} exploit the 3D-CNN model for multi-task learning of codec classification and quality assessment for compressed videos. 
\citet{zhang2018blind,zhang2019objective} make use of both video and image data with transfer learning.
However, all these methods are proposed for quality assessment of synthetically-distorted videos, and they are not applicable to in-the-wild videos or their performances are poor on in-the-wild datasets. 
Note that the relevant concept ``streaming video quality-of-experience (QoE)" is beyond the scope of this work, and the interested reader can refer to these two good surveys~\citep{seufert2014survey,juluri2015measurement}.

Quality assessment of in-the-wild videos has been attracting significant attention in recent years. 
Four relevant datasets have been constructed with corresponding subjective studies, \textit{i.e.}, CVD2014~\citep{nuutinen2016cvd2014}, KoNViD-1k~\citep{hosu2017konstanz}, LIVE-Qualcomm~\citep{ghadiyaram2018capture}, and LIVE-VQC~\citep{sinno2019large}.
Since we cannot access the pristine reference videos in this situation, only no-reference VQA (NR-VQA) methods are applicable. 
The deep learning-based VQA models described in the last paragraph are unfeasible in this problem since they either need the reference information~\citep{zhang2018blind,kim2018deep,zhang2019objective} or only suit for compression artifacts~\citep{liu2018end}.
Thus, in our previous work~\citep{li2019quality}, we propose a deep learning-based model for predicting the quality of in-the-wild videos. 
The model extracts content-aware distortion-sensitive features from CNN models trained for image classification tasks, and uses a gated recurrent unit (GRU) followed by a subjectively-inspired temporal pooling layer for modeling the temporal-memory effect.
Concurrent works to our previous work are~\citet{you2019deep,varga2019no1,varga2019no2}. 
Although all of these methods achieve a good performance, they do not enable mixing multiple datasets during the training phase. 
As a result, their performances are poor in the cross-dataset evaluation setting. 
The main purpose of this paper is to propose an elegant mixed datasets training strategy. 
With this strategy, we can train a unified model that learns the characteristics of videos from all datasets and thus further improve the overall performance over the datasets.
\subsection{Mixed Datasets Training}
\label{sec:mdt}
\indent\indent Mixed datasets training has two advantages. 
One is to provide a unified model for all datasets. 
The other is to take full advantage of existing relevant datasets for improving the model learning.
Therefore, many computer vision tasks consider mixed datasets training, such as person re-identification~\citep{lv2018unsupervised,li2019cross}, monocular depth estimation~\citep{lasinger2019towards}, and human parsing~\citep{he2019grapy}.

There are some relevant works in quality assessment tasks that consider mixed datasets training.
The challenge is that ranges of subjective quality scores are inconsistent across datasets.
\citet{korhonen2019two} uses a na\"ive method to handle this challenge, \textit{i.e.}, linearly re-scaling the subjective quality scores of different datasets to the same range. 
Pair-wise learning considers only the relative quality score instead of the absolute subjective quality scores, and thus can bypass the ``inconformity" problem.
Therefore, several I/VQA works consider pair-wise learning for mixed datasets training, while they use different loss functions for training~\citep{yang2019cnn,zhang2019learning,krasula2019training}. 
\citet{yang2019cnn} use the margin ranking loss and the Euclidean loss.
\citet{zhang2019learning} consider the cross entropy loss and the fidelity loss.
\citet{krasula2019training} determine different and similar pairs based on statistical analysis on the mean and standard deviation of subjective ratings, and then define the training objective as the correct classification rate of these pairs. 
However, pair-wise learning will increase the training time. In the next section, we show that our proposed monotonicity-induced loss can be regarded as an extension of the losses in~\citet{yang2019cnn} and~\citet{zhang2019learning} with a more efficient implementation. Besides the monotonicity-induced loss, we also propose a linearity-induced loss and assign a dataset-specific perceptual scale alignment to enable mixing multiple datasets during the training phase.

\section{Proposed Method}
\label{sec:method}
\subsection{Overview}
Fig.~\ref{fig:framework} shows the overview of the proposed unified VQA framework for quality assessment of in-the-wild videos. 
Our VQA model consists of three stages: relative quality assessor, nonlinear mapping, and dataset-specific perceptual scale alignment for predicting relative quality, perceptual quality, and subjective quality, respectively. 

The flow of our proposed unified framework is as follows. 
At the first stage, to predict the relative quality, we learn a relative quality assessor with the supervision of a monotonicity-inspired loss, where the monotonicity-induced loss is derived from the monotonicity condition and it is the sum of all pair-wise losses. 
To account for the content dependency and temporal-memory effects of human perception, we design our relative quality assessor as a deep neural network that includes two key modules: content-aware feature extraction and modeling of temporal-memory effect. 
At the second stage, to predict the perceptual quality, a nonlinear mapping module is added after the relative quality assessor, to explicitly account for the nonlinearity of human perception. 
The parameters in this module are learned with the supervision of a linearity-induced loss based on Pearson's linear correlation.
At the third stage, to predict the subjective quality, a dataset-specific perceptual scale alignment layer is added to map the predicted perceptual quality to the subjective quality of a video on each dataset. 
After the alignment, the widely-used error-induced loss is used as the supervision.

Thus, in our mixed datasets training strategy, three kinds of losses are involved.
For each dataset, the overall loss is the sum of these three kinds of losses on the dataset.
To emphasize the datasets with larger loss values, our final training loss is a softmax-weighted loss over all training datasets.
With this strategy, we can learn a single unified VQA model for multiple datasets by mixing them all during the training phase.

\begin{figure*}[!htb]
    \centering
    \includegraphics[width=.8\textwidth]{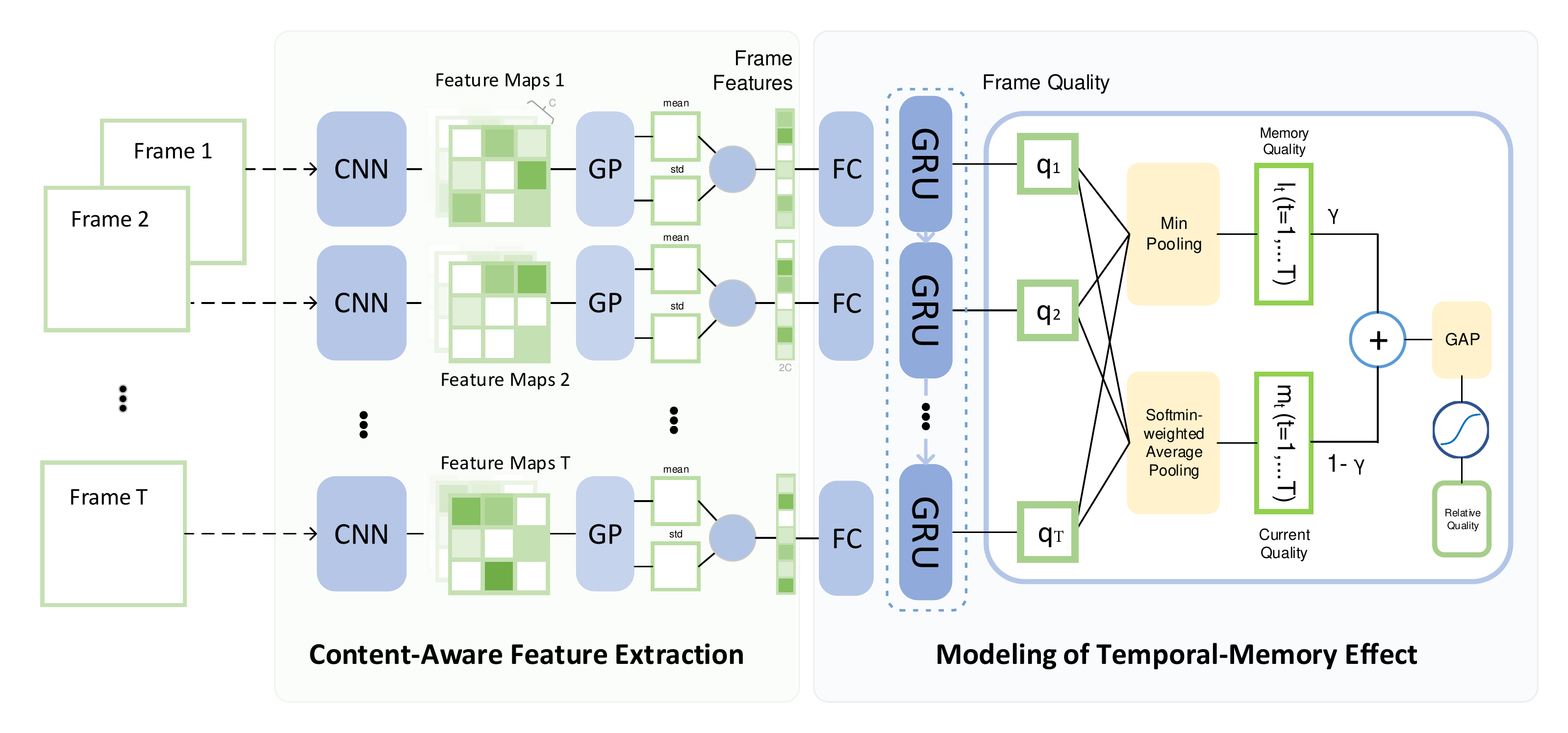}
    \caption{Relative Quality Assessor. It mainly consists of two modules. Module I includes a pre-trained CNN with effective global pooling (GP) serving as a content-aware feature extractor. Module II models temporal-memory effect and it includes two sub-modules: a GRU network and a subjectively-inspired temporal pooling layer. Note that the GRU network is the unrolled version of one GRU and the parallel CNNs/FCs share weights.}
    \label{fig:rqa}
\end{figure*}
 
\subsection{Relative Quality Assessor}
This subsection describes the design of the relative quality assessor. 
The framework of our relative quality assessor is shown in Fig.~\ref{fig:rqa}. 
We adopt the model in our previous work~\citep{li2019quality} as the backbone of the relative quality assessor. 
It integrates the two eminent effects of human perception into the assessor. 
One is the content dependency effect, which guides us introducing the content-aware feature extraction module. 
The other is the temporal-memory effect, which is modeled in the feature level and the quality score level.

\subsubsection{Content-Aware Feature Extraction}
In the visual quality assessment task, human perception is content dependent~\citep{siahaan2018semantic,triantaphillidou2007image,wang2017videoset,bampis2017study,zhang2018unreasonable,li2019quality,li2019has}. 
This can be attributed to the fact that, the complexity of distortions, the human tolerance thresholds for distortions, and the human preferences could vary a lot in different contents/scenes.
Since there are diverse contents in the in-the-wild scenario, a relative quality assessor which mimics human perception, should take this effect into accounts. 
So we need to extract features that are not only distortion-sensitive but also content-aware.
The image classification models pre-trained on ImageNet~\citep{deng2009imagenet} using CNN possess the discriminatory power of different content information.
Thus, the deep features extracted from these models, \textit{e.g.}, ResNet~\citep{he2016deep}, are expected to be content-aware. 
Meanwhile, \citet{dodge2016understanding} point out that the deep features are distortion-sensitive. 
So it is reasonable to extract content-aware and distortion-sensitive features from pre-trained image classification models.

Assuming the video is a stack of $T$ frames $\mathbf{I}_t (t=1,2,\dots,T)$, we feed each video frame into a pre-trained CNN model and get the corresponding deep feature maps $\mathbf{M}_t$ from its top convolutional layer:
\begin{equation}\label{eq:CNN}
\mathbf{M}_t = \mathrm{CNN}(\mathbf{I}_t).
\end{equation}

$\mathbf{M}_{t}$ contains a total of $C$ feature maps.
Then, we apply spatial global pooling (GP) for each feature map of $\mathbf{M}_{t}$. 
Applying only the spatial global average pooling operation ($\mathrm{GP}_{\mathrm{mean}}$) to $\mathbf{M}_t$ discards much information of $\mathbf{M}_t$. 
We further consider the spatial global standard deviation pooling operation ($\mathrm{GP}_{\mathrm{std}}$) to preserve the variation information in $\mathbf{M}_t$. 
The output feature vectors of $\mathrm{GP}_{\mathrm{mean}}, \mathrm{GP}_{\mathrm{std}}$ are $\mathbf{f}_{t}^{\mathrm{mean}}, \mathbf{f}_{t}^{\mathrm{std}}$ respectively.
\begin{equation}\label{eq:GP}
\mathbf{f}_{t}^{\mathrm{mean}}  = \mathrm{GP}_{\mathrm{mean}}(\mathbf{M}_t),\quad
\mathbf{f}_{t}^{\mathrm{std}}  = \mathrm{GP}_{\mathrm{std}}(\mathbf{M}_t).
\end{equation}

After that, $\mathbf{f}_{t}^{\mathrm{mean}}$ and $\mathbf{f}_{t}^{\mathrm{std}}$ are concatenated to serve as content-aware and distortion-sensitive features $\mathbf{f}_t$:
\begin{equation}\label{eq:concatenation}
\mathbf{f}_t = \mathbf{f}_{t}^{\mathrm{mean}}\oplus\mathbf{f}_{t}^{\mathrm{std}},
\end{equation}
where \(\oplus\) is the concatenation operator and the length of $\mathbf{f}_t$ is $2C$.

\subsubsection{Modeling of Temporal-Memory Effect}
Temporal-memory effect is another important clue for designing objective VQA models~\citep{park2013video,seshadrinathan2011temporal, xu2014no,choi2018video,kim2018deep}. 
It induces that video quality rating is influenced by historic memory.
We model the temporal-memory effect in two aspects. 
In the feature integration aspect, we adopt a GRU network for modeling the long-term dependencies in our method. 
In the quality pooling aspect, we propose a subjectively-inspired temporal pooling model and embed it into the network.

\textbf{Long-term dependencies modeling}.
Existing NR-VQA methods cannot well model the long-term dependencies in the VQA task. 
To handle this issue, we resort to GRU~\citep{cho2014learning}, a recurrent neural network model with gates control. 

The dimension of the extracted content-aware features is very high, which is not suitable for GRU training. 
Therefore, we perform dimension reduction using a single fully-connected (FC) layer before feeding them into GRU,  that is:
\begin{equation}\label{eq:linear DR}
\mathbf{x}_t = \mathbf{W}_{fx}\mathbf{f}_t+\mathbf{b}_{fx},
\end{equation}
where $\mathbf{W}_{fx}$ and $\mathbf{b}_{fx}$ are the parameters in the single FC layer. 
Without the bias term, it acts as a linear dimension reduction model.

After dimension reduction, the reduced features $\mathbf{x}_t (t=1, 2, \cdots, T)$ are sent to GRU. 
We consider the hidden states of GRU as the integrated features $\mathbf{h}_t$, whose initial values are $\mathbf{h}_0$.  $\mathbf{h}_t$ is calculated as follow.
\begin{equation}\label{eq:GRU}
\mathbf{h}_t = \mathrm{GRU}(\mathbf{x}_t, \mathbf{h}_{t-1}).
\end{equation}

With the integrated features $\mathbf{h}_t$, we predict the frame quality score $q_t$ by adding a single FC layer:
\begin{equation}
q_t = \mathbf{W}_{hq}\mathbf{h}_t+\mathbf{b}_{hq},
\end{equation}
where $\mathbf{W}_{hq}$ and $\mathbf{b}_{hq}$ are the weight and bias parameters.

\textbf{Subjectively-inspired temporal pooling}.
In subjective experiments, subjects are intolerant of poor quality video events~\citep{park2013video}. 
More specifically, temporal hysteresis effect is found in the subjective experiments~\citep{seshadrinathan2011temporal}.
That is, subjects react sharply to drops in video quality and provide poor quality for such time interval, but react dully to improvements in video quality thereon. 
Inspired by these observations, to connect the predicted frame-level quality to the video-level quality, we put forward a new differentiable temporal pooling model. Details are as follows.

To mimic the human's intolerance to poor quality events, we define a memory quality element $l_t$ at the $t$-th frame as the minimum of quality scores over the previous several frames:
\begin{equation}\label{eq:memory score}
l_t =\left\{
\begin{array}{rcl}
q_t & &\mbox{for}\, \ t=1, \\
\min_{k\in V_{\mathrm{prev}}}{q_{k}}  & & \mbox{for}\, \ t>1, 
\end{array} \right.
\end{equation}
where $V_{\mathrm{prev}}=\{\max{(1,t-\tau)},\cdots, t-2, t-1\}$ is the index set of the considered frames, and $\tau$ is a hyper-parameter relating to the temporal duration.

Accounting for the fact that subjects react sharply to the drops in quality but react dully to the improvements in quality, we construct a current quality element $m_t$ at the $t$-th frame, using the weighted quality scores over the next several frames, where larger weights are assigned for worse quality frames. 
Specifically, we define the weights $w_t^k$ by a differentiable softmin function, \textit{i.e.}, a composition of the negative linear function and the softmax function.
\begin{equation}\label{eq:current score}
m_t  = \sum_{k\in V_{\mathrm{next}}}q_{k}w_t^{k},\quad
w_t^k  = \frac{e^{-q_k}}{\sum_{j\in V_{\mathrm{next}}}e^{-q_j}}, k\in V_{\mathrm{next}}, 
\end{equation}
where $V_{\mathrm{next}}=\{t,t+1,\cdots, \min{(t+\tau,T)}\}$ is the index set of the related frames.

In the end, we approximate the subjective frame quality scores by linearly combining the memory quality and current quality elements. 
The relative quality score $Q_r$ is then calculated by temporal global average pooling (GAP) of the approximate scores and bounded by a sigmoid function:
\begin{equation}\label{eq:approximate quality}
q'_t  = \gamma l_t + (1-\gamma)m_t,
\end{equation}
\begin{equation}\label{eq:video quality}
Q_r  = \sigma\left(\frac{1}{T}\sum_{t=1}^{T}q'_t\right),
\end{equation}
where $\gamma$ is a hyper-parameter to balance the contributions of memory and current elements to the approximate score, and $\sigma(\cdot)$ is the sigmoid function.

\subsection{Nonlinear Mapping}
For predicting the perceptual quality, we add a nonlinear mapping module after the relative quality assessor to explicitly account for the nonlinearity of human perception on video quality~\citep{vqeg2000fr}. 
The nonlinear mapping module can be a complex neural network with many parameters, or just a simple nonlinear function with few parameters. 

Following the suggestion by Video Quality Experts Group~\citep{vqeg2000fr}, we can use a 4-parameter logistic function for nonlinear mapping.
\begin{equation}\label{eq:5nlm}
Q_p = f(Q_r)= \frac{\beta_1-\beta_2}{1+e^{-\frac{Q_r-\beta_3}{|\beta_4|}}}+\beta_2,
\end{equation}
where $\beta_1$ to $\beta_4$ are fitting parameters, $Q_r$ is the relative quality score, and $Q_p$ is the perceptual quality score.

\begin{figure}[!htb]
\begin{center}
\includegraphics[width=\linewidth]{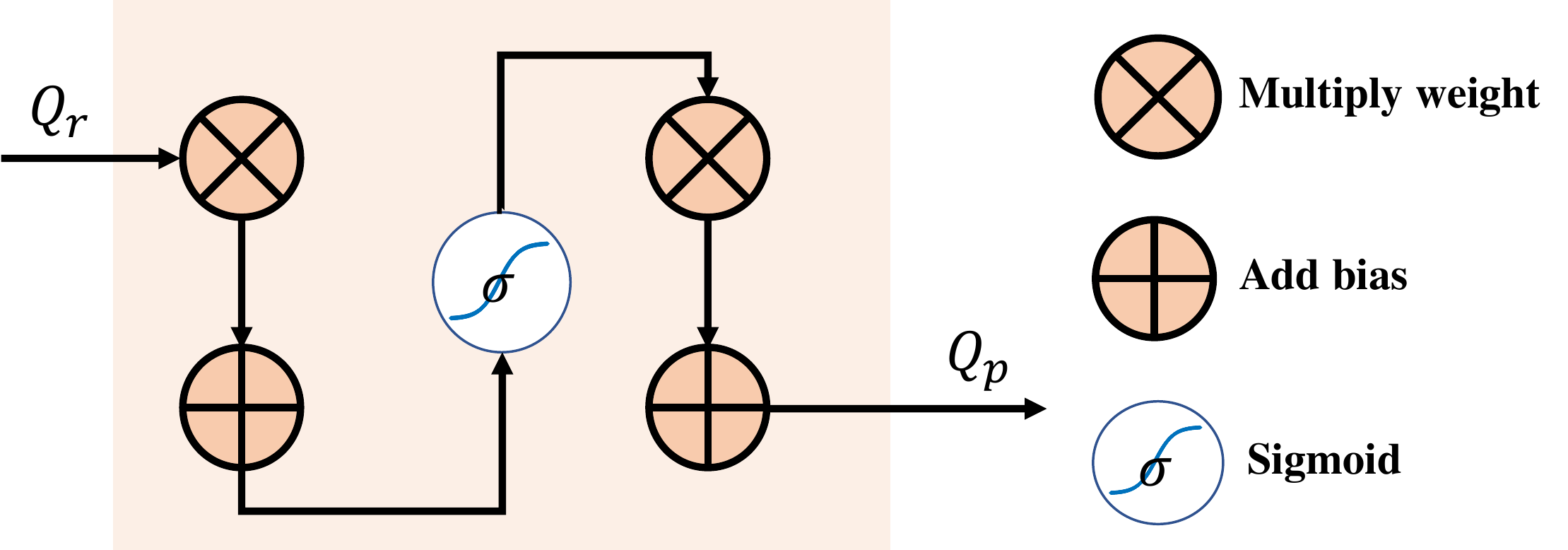}%
\end{center}
   \caption{Illustration of the nonlinear mapping module}
\label{fig:nlm}
\end{figure}

We can reformulate Eqn.~(\ref{eq:5nlm}) as the following.
\begin{equation}\label{eq:new5nlm}
Q_p = \beta'_1\sigma{(\beta'_4Q_r+\beta'_3)}+\beta'_2,
\end{equation}
where $\beta'_1\leftarrow\beta_1-\beta_2, \beta'_2\leftarrow\beta_2, \beta'_3\leftarrow-\frac{\beta_3}{|\beta_4|}$, and $\beta'_4\leftarrow \frac{1}{|\beta_4|}$. 
And $\beta'_1, \beta'_2$ are parameters to control the range of $Q_p$. $\beta'_3, \beta'_4$ are parameters to control the normalization of $Q_r$. 
Therefore, it is equivalent to ``Linear (\textit{i.e.}, Multiply Weights and Add Bias)+Sigmoid+Linear", as shown in Fig.~\ref{fig:nlm}. 

With the reformulation, we can design the 4-parameter nonlinear mapping as a network module.
Since we will handle the scale problem in the next stage, the nonlinear mapping just handles the nonlinearity and does not change the scale, \textit{i.e.}, the ranges of $Q_r$ and $Q_p$ are both $[0, 1]$. 
We need to initialize the 4 parameters in this module at the start of the training. 
Random initialization is not a good choice since we have priors of $Q_r$ and $Q_p$.
Therefore, we can have a better initialization as follows.
\begin{equation}\label{eq:nlm initialization}
\begin{array}{l}
\beta'_1\leftarrow \sup{(Q_p)}-\inf{(Q_p)} = 1, \\
\beta'_2\leftarrow \inf{(Q_p)} = 0, \\
\beta'_3\leftarrow-c*\mathrm{mean}{(Q_r)}/\mathrm{std}{(Q_r)}, c = 1, \\
\beta'_4\leftarrow c/\mathrm{std}{(Q_r)}, 
\end{array}
\end{equation}
where $\mathrm{mean}{(\cdot)}, \mathrm{std}{(\cdot)}, \inf{(\cdot)}, \sup{(\cdot)}$ indicate the mean, standard deviation, infimum, and supremum functions, respectively.

\subsection{Dataset-Specific Perceptual Scale Alignment}
Since the subjective study is designed to reflect human perception on video quality, based on the concepts of subjective quality and perceptual quality, we can assume that the subjective quality is linearly correlated with the perceptual quality.
Thus, the perceptual scale alignment can be simply set as a specific FC layer.
\begin{equation}\label{eq:aligned}
Q_s  = \xi_1 Q_p + \xi_2,
\end{equation}
where $Q_s$ is the predicted subjective quality score, and $\xi_1, \xi_2$ are the scale and shift parameters. 

Since different datasets have different ranges of subjective quality scores, we need a dataset-specific alignment of perceptual scale on each dataset. Eqn.~(\ref{eq:aligned}) is then modified as follows.
\begin{equation}\label{eq:specific aligned}
Q^{(d)}_s  = \xi_1^{(d)} Q_p + \xi_2^{(d)} (d=1, \cdots, D), 
\end{equation}
where $Q^{(d)}_s$ is the predicted subjective quality score on the $d$-th dataset, $\xi_1^{(d)}, \xi_2^{(d)}$ are the scale and shift parameters for the $d$-th dataset, and $D$ is the number of considered datasets. 
These parameters can be determined by least square regression (LSR) or just jointly learned with other parameters by iterative stochastic gradient decent (SGD) algorithm. 
The latter way can provide supervision for end-to-end network training and it is adopted in our mixed datasets training strategy.


\subsection{Mixed Datasets Training Strategy}
We have introduced the unified VQA model in the above. 
In this subsection, we show how we can enable mixed datasets training when the ranges of subjective quality scores are not consistent among the VQA datasets. 
For the first and second stages, the relative quality and perceptual quality are not involved with the ranges of subjective quality.
We bypass the inconformity problem by designing two losses to supervise the training process of predicting relative quality and perceptual quality. 
For the third stage, to predict subjective quality of videos on each dataset, we learn a dataset-specific perceptual scale alignment for each dataset to avoid the inconformity caused by the na\"ive linear re-scaling.
Such dataset-specific alignment enables mixing multiple datasets during the training without disturbing the process. 
Specifically, monotonicity-induced loss is proposed for Stage 1 ``relative quality assessor", and linearity-induced loss is adopted for Stage 2 ``nonlinear mapping". 
As for Stage 3 ``dataset-specific perceptual scale alignment", we can just use the widely-used error-induced (\textit{i.e.}, normalized $L_1$) loss as the supervision.

Assume we have $D$ datasets of VQA, and the $d$-th dataset contains $N_d$ videos ($d=1,\cdots, D$). 
For the $i$-th video of the $d$-th dataset, we denote its predicted relative quality score as $Q^{d, i}_r$, the predicted perceptual quality score as $Q^{d, i}_p$, the predicted subjective quality score as $Q^{d, i}_s$, and ground-truth subjective quality score as $Q^{d, i}$.

\subsubsection{Monotonicity-Induced Loss}
The goal of relative quality assessor is to achieve the best prediction monotonicity. 
That is, it aims to give a quality ranking for any list/pair of videos from the same dataset, that is consistent with subjective quality. 
A natural objective is to maximize the Spearman's rank-order correlation coefficient (SROCC) or Kendall's rank-order correlation coefficient (KROCC). 
However, they are not applicable to back-propagation based neural network optimization due to their non-differentiable property. 

Let us take a close look at the monotonicity condition. 
For all  $i, j = 1, \cdots, N_d$, $d=1, \cdots, D$, 
\begin{equation}\label{eq:monotonicity}
(Q^{d, i}_r-Q^{d, j}_r)(Q^{d, i}-Q^{d, j})\ge 0.
\end{equation}

So we can consider the sum of the pair-wise losses as a surrogate.  We call this monotonicity-induced loss, which is defined as follows.
\begin{equation}\label{eq:mloss}
\begin{array}{l}
L^{(d)}_{\mathrm{rel}} =\frac{2}{N_d(N_d-1)}\sum_{i<j} E^{d,(i,j)}_r,\\[2mm]
E^{d,(i,j)}_r=\max\{(Q^{d, i}_r-Q^{d, j}_r)*\mathrm{sign}(Q^{d, j}-Q^{d, i}), 0\},
\end{array}
\end{equation}
where $E^{d,(i,j)}_r$ is the error term induced by the monotonicity condition, \textit{i.e.}, Eqn.~(\ref{eq:monotonicity}). 
Here, we choose the error term as the margin ranking loss used in~\citet{yang2019cnn}. 
It can also be in the form of the fidelity loss or the cross entropy loss as described in~\citet{zhang2019learning}. 
Note that compared to pair-wise learning, the number of forward operations is reduced from $\mathrm{C}_{N_d}^2$ to $N_d$ in our list-wise learning setting. 
Together with the vectorization form, we provide a much more efficient implementation and save more training time than the pair-wise learning used in image quality assessment~\citep{yang2019cnn,zhang2019learning}.

\subsubsection{Linearity-Induced Loss}
The goal of the nonlinear mapping module is to achieve the best prediction linearity between the predicted perceptual quality scores and the subjective quality scores. 
Pearson's linear correlation coefficient (PLCC) is a good objective for characterizing linearity. 
And it is differentiable, so we can define our linearity-induced loss for nonlinear mapping module as follow.
\begin{equation}\label{eq:aloss}
\begin{array}{l}
L^{(d)}_{\mathrm{lin}} =(1-\mathrm{PLCC}_d)/2, \\[2mm]
\mathrm{PLCC}_d=\frac{\sum_{i}(Q^{d, i}_p-\bar{Q}_p^{(d)})(Q^{d, i}-\bar{Q}^{(d)})}{\sqrt{\sum_{i}{(Q^{d, i}_p-\bar{Q}_p^{(d)})}^2\sum_{i}{(Q^{d, i}-\bar{Q}^{(d)})}^2}},
\end{array}
\end{equation}
where $\bar{Q}_p^{(d)}=\frac{1}{N_d}\sum_{i}Q^{d, i}_p$ and $\bar{Q}^{(d)}=\frac{1}{N_d}\sum_{i}Q^{d, i}$. Note that PLCC-induced loss is also considered in~\citep{ma2018geometric,liu2018end,li2020norm}.

\subsubsection{Error-Induced Loss}
After dataset-specific perceptual scale alignment, our goal is to minimize the absolute prediction error. In this stage, any regression error can be used as the loss function. We simply choose the widely-used error-induced (\textit{i.e.}, normalized $L_1$) loss in this work. More general and robust regression losses may be explored to further improve the optimization performance~\citep{barron2019general}. To balance the losses among different datasets, we consider the inverse scale on each dataset as a normalization factor.
\begin{equation}\label{eq:l1}
L^{(d)}_{\mathrm{err}} =\sum_{i} \frac{1}{N_d}\frac{\left|Q^{d, i}_s-Q^{d, i}\right|}{S_d},
\end{equation}
where $S_d=\max{(Q^{d,i})}-\min{(Q^{d,i})}$ is the range of the subjective quality scores on the $d$-th dataset.

\subsubsection{Final Loss for The Whole Model}
We obtain the loss for the $d$-th dataset $(d=1,\cdots,D)$ from the above three losses $L^{(d)}_{\mathrm{rel}}, L^{(d)}_{\mathrm{lin}}, L^{(d)}_{\mathrm{err}}$.
\begin{equation}\label{eq:loss on d}
L^{(d)} = L^{(d)}_{\mathrm{rel}} + L^{(d)}_{\mathrm{lin}} + L^{(d)}_{\mathrm{err}}, 
\end{equation}
and the overall final loss for training a single unified model from multiple datasets is defined as a softmax-weighted average of the losses over all datasets.
\begin{equation}\label{eq:overall loss}
\begin{array}{l}
L= \sum_{d} w^{(d)}L^{(d)}, \\[2mm]
w^{(d)}=  e^{L^{(d)}}/\sum_{d} e^{L^{(d)}},
\end{array}
\end{equation}
where $w^{(d)}$ is the weight of $L^{(d)}$ ($d=1,\cdots,D$).

\subsection{Implementation Details}
We choose ResNet-50~\citep{he2016deep} pre-trained on ImageNet~\citep{deng2009imagenet} for the content-aware feature extraction, and the feature maps are extracted from its top convolutional layer ``res5c". 
In this instance, the dimension of $\mathbf{f}_t$ is $2C=4096$.
The feature dimension is then reduced from 4096 to 128, followed by a single-layer GRU network with hidden size 32. 
$\tau$ and $\gamma$ in the temporal pooling layer are set as 12 and 0.5, respectively. 
We choose the 4-parameter nonlinear mapping, and the parameters in the module are initialized based on Eqn.~(\ref{eq:nlm initialization}).
We freeze the parameters in the pre-trained ResNet-50 to ensure that the content-aware property is not altered, and we train the other part of the network in an end-to-end manner. 
We train our model using Adam optimizer~\citep{kingma2014adam} for 40 epochs with an initial learning rate 1e-4, a training batch size 32 for each dataset.
The proposed model is implemented with PyTorch~\citep{paszke2019pytorch}.
To support reproducible scientific research, we release the code at \url{https://github.com/lidq92/MDTVSFA}.

\section{Experiments}
\label{sec:experiments}
This section reports and analyzes the experimental results. 
We first describe the experimental setup, including the benchmark datasets, compared methods and basic evaluation criteria. 
Next, we study the effectiveness of our mixed datasets training strategy. 
After that, the performance comparison is carried out between our method and the state-of-the-art methods. 
Finally, the computational efficiency is briefly discussed.

\begin{table*}[!hbt]
    \centering
    \caption{Brief information of the four benchmark datasets, including the information of the videos and the information of the corresponding subjective study.}
    \label{tab:datasets}
    \resizebox{\textwidth}{!}{
    \begin{tabular}{lcccc}
    \toprule
    \multirow{2}{*}{Dataset} & CVD2014 & KoNViD-1k & LIVE-Qualcomm & LIVE-VQC \\
    & \citep{nuutinen2016cvd2014} & \citep{hosu2017konstanz} & \citep{ghadiyaram2018capture} & \citep{sinno2019large}\\
    \midrule
    Number of Videos & 234 & 1200 & 208 & 585 \\
    Number of Scenes & 5 & $\approx$1200 & 54 & $\approx$585 \\
    Number of Devices & 78 & - & 8 & 101 \\
    Number of Users & - & 480 & - & 80 \\
    Video Orientations & Landscape & Landscape & Landscape & Portrait, Landscape \\
    Video Resolutions & 1280$\times$720 or 640$\times$480 & 960$\times$540 & 1920$\times$1080 & 1920$\times$1080 to 320$\times$240\\
    Number of Resolutions & 2 & 1 & 1 & 18\\
    Frames Per Second & 11 to 31 & 24, 25 or 30 & 30 & 19-30 (one 120)\\
    Time Span & 10-25s & 8s & 15s & 10s \\
    Max Video Length & 830 frames & 240 frames & 526 frames & 1202 frames\\
    \midrule
    Test Methodology &  Single stimulus &  Single stimulus &  Single stimulus &  Single stimulus \\
    Lab or Crowdsourcing & Lab & Crowdsourcing & Lab & Crowdsourcing\\
    Number of Participants & 210 & 642 & 39 & 4776 \\
    Number of Ratings & 28-33 & $>$50, average 114 & 18 & $>$200, average 240\\
    Raw Ratings Provided & Yes & Yes & No & No\\
    Mean Opinion Score & [-6.50, 93.38] & [1.22, 4.64] & [16.5621, 73.6428] & [6.2237, 94.2865]\\
    \bottomrule
    \end{tabular}
    }
\end{table*}

\subsection{Experimental Setup}
\indent\indent\textbf{Benchmark datasets}. 
Currently, there are four datasets for quality assessment of in-the-wild videos, including CVD2014~\citep{nuutinen2016cvd2014}, KoNViD-1k~\citep{hosu2017konstanz}, LIVE-Qualcomm~\citep{ghadiyaram2018capture}, and LIVE-VQC~\citep{sinno2019large}. 
We summarize their brief information in Table~\ref{tab:datasets}. 
We can see that the four datasets have different characteristics and the ranges of mean opinion score (MOS) are different among these datasets. 
In the default setting, each dataset is split into 80\%, and 20\% for training and testing, respectively. 
No overlap is among training and testing data. 
And 25\% of the training data are used for validation.
We repeat this procedure 10 times to avoid performance bias.

\textbf{Compared methods}. 
Only NR methods are applicable for quality assessment of in-the-wild videos.
We select five state-of-the-art NR methods for comparison, whose original codes are released by the authors, icluding VBLIINDS~\citep{saad2014blind}, VIIDEO~\citep{mittal2016completely}, BRISQUE~\citep{mittal2012no}\footnote{Video-level features of BRISQUE are the average pooling of its frame-level features.}, NIQE~\citep{mittal2013making}, and CORNIA~\citep{ye2012unsupervised}. 
Besides, we also show some relevant results reported from previous arts, \textit{e.g.}, TLVQM~\citep{korhonen2019two}. 
Note that the method in~\citet{zhang2018blind} needs scores of full-reference methods, methods in~\citet{kim2018deep} and~\citet{zhang2019objective} are full-reference methods, and thus they are unfeasible for this problem.

\textbf{Basic evaluation criteria}. 
We follow the suggestion from Video Quality Experts Group~\citep{vqeg2000fr}, and report SROCC and PLCC as the criteria of prediction monotonicity and prediction accuracy, respectively. 
Better VQA methods should have larger values of SROCC and PLCC. 
When the predicted quality scores are not the same scale as the subjective scores, PLCC is calculated after nonlinear mapping with a 4-parameter logistic function as suggested by VQEG.

\begin{table*}[!hbt]
    \centering
    \caption{Performance gain in terms of median SROCC when one more dataset is added into the training data. $D_{+}$ is the added dataset for training, $D_B$ indicates the base datasets for training before adding $D_{+}$, and $D_T$ indicates the dataset for testing. ``Overall Performance'' is indicated by the dataset-size weighted average of median SROCC. Positive gain is shown in blue, while negative gain is shown in red. The performance values in the scenario $D_T\subseteq D_B$ are marked in a light gray background, and the performance values in the scenario $D_T\cap (D_B\cup D_{+})=\emptyset$ are marked in a dark gray background.}
    \label{tab:mixing more datasets}

    \begin{small}
    
    \resizebox{\textwidth}{!}{
    \begin{tabular}{ccccccc}
    \toprule
    \multicolumn{2}{c}{Train data} & \multicolumn{4}{c}{Test set of $D_T$} & Overall \\
    train sets of $D_B$ &train set of $D_{+}$ & CVD2014 & KoNViD-1k & LIVE-Qualcomm & LIVE-VQC & Performance\\
    \midrule
KoNViD-1k & \multirow{7}{*}{CVD2014}
& 0.6474({\color{blue}+0.2078})
& \multicolumn{1}{>{\columncolor{lightgray}}c}{0.7809({\color{red}-0.0067})}
&  \multicolumn{1}{>{\columncolor{darkgray}}c}{0.6732({\color{red}-0.0248})}
&  \multicolumn{1}{>{\columncolor{darkgray}}c}{0.7160({\color{red}-0.0227})}
& 0.7398({\color{blue}+0.0100})\\

LIVE-Qualcomm & 
& 0.5879({\color{blue}+0.2757})
&  \multicolumn{1}{>{\columncolor{darkgray}}c}{0.6128({\color{blue}+0.0563})}
&  \multicolumn{1}{>{\columncolor{lightgray}}c}{0.7538({\color{blue}+0.0344})}
&  \multicolumn{1}{>{\columncolor{darkgray}}c}{0.6214({\color{red}-0.0061})}
& 0.6256({\color{blue}+0.0609})\\

LIVE-VQC & 
& 0.4819({\color{blue}+0.3556})
&  \multicolumn{1}{>{\columncolor{darkgray}}c}{0.7059({\color{blue}+0.0319})}
&  \multicolumn{1}{>{\columncolor{darkgray}}c}{0.6550({\color{blue}+0.0246})}
&  \multicolumn{1}{>{\columncolor{lightgray}}c}{0.7470({\color{red}-0.0193})}
& 0.6884({\color{blue}+0.0518})\\

KoNViD-1k+LIVE-Qualcomm & 
& 0.6933({\color{blue}+0.1479})
&  \multicolumn{1}{>{\columncolor{lightgray}}c}{0.7836({\color{red}-0.0177})}
&  \multicolumn{1}{>{\columncolor{lightgray}}c}{0.8170({\color{red}-0.0012})}
&  \multicolumn{1}{>{\columncolor{darkgray}}c}{0.6969({\color{red}-0.0118})}
& 0.7544({\color{blue}+0.0028})\\

KoNViD-1k+LIVE-VQC & 
& 0.6325({\color{blue}+0.1978})
& \multicolumn{1}{>{\columncolor{lightgray}}c}{0.7974({\color{red}-0.0115})}
& \multicolumn{1}{>{\columncolor{darkgray}}c}{0.6995({\color{blue}+0.0018})}
& \multicolumn{1}{>{\columncolor{lightgray}}c}{0.7461({\color{red}-0.0018})}
& 0.7575({\color{blue}+0.0143})\\

LIVE-Qualcomm+LIVE-VQC & 
& 0.5849({\color{blue}+0.2516})
& \multicolumn{1}{>{\columncolor{darkgray}}c}{0.6843({\color{blue}+0.0310})}
& \multicolumn{1}{>{\columncolor{lightgray}}c}{0.8010({\color{blue}+0.0108})}
& \multicolumn{1}{>{\columncolor{lightgray}}c}{0.7434({\color{red}-0.0222})}
& 0.7002({\color{blue}+0.0383})\\

KoNViD-1k+LIVE-Qualcomm+LIVE-VQC & 
& 0.6422({\color{blue}+0.1870})
& \multicolumn{1}{>{\columncolor{lightgray}}c}{0.7906({\color{red}-0.0113})}
& \multicolumn{1}{>{\columncolor{lightgray}}c}{0.8003({\color{blue}+0.0042})}
& \multicolumn{1}{>{\columncolor{lightgray}}c}{0.7476({\color{red}-0.0124})}
& 0.7646({\color{blue}+0.0107})\\

    \midrule 
CVD2014 & \multirow{7}{*}{KoNViD-1k}
& \multicolumn{1}{>{\columncolor{lightgray}}c}{0.8747({\color{red}-0.0195})}
& 0.6051({\color{blue}+0.1692})
& \multicolumn{1}{>{\columncolor{darkgray}}c}{0.3919({\color{blue}+0.2565})}
& \multicolumn{1}{>{\columncolor{darkgray}}c}{0.4950({\color{blue}+0.1983})}
& 0.5846({\color{blue}+0.1652})\\

LIVE-Qualcomm & 
& \multicolumn{1}{>{\columncolor{darkgray}}c}{0.5879({\color{blue}+0.1054})}
& 0.6128({\color{blue}+0.1708})
& \multicolumn{1}{>{\columncolor{lightgray}}c}{0.7538({\color{blue}+0.0631})}
& \multicolumn{1}{>{\columncolor{darkgray}}c}{0.6214({\color{blue}+0.0755})}
& 0.6256({\color{blue}+0.1288})\\

LIVE-VQC & 
& \multicolumn{1}{>{\columncolor{darkgray}}c}{0.4819({\color{blue}+0.1506})}
& 0.7059({\color{blue}+0.0915})
& \multicolumn{1}{>{\columncolor{darkgray}}c}{0.6550({\color{blue}+0.0445})}
& \multicolumn{1}{>{\columncolor{lightgray}}c}{0.7470({\color{red}-0.0008})}
& 0.6884({\color{blue}+0.0691})\\

CVD2014+LIVE-Qualcomm & 
& \multicolumn{1}{>{\columncolor{lightgray}}c}{0.8636({\color{red}-0.0224})}
& 0.6691({\color{blue}+0.0968})
& \multicolumn{1}{>{\columncolor{lightgray}}c}{0.7883({\color{blue}+0.0275})}
& \multicolumn{1}{>{\columncolor{darkgray}}c}{0.6153({\color{blue}+0.0698})}
& 0.6865({\color{blue}+0.0707})\\

CVD2014+LIVE-VQC & 
& \multicolumn{1}{>{\columncolor{lightgray}}c}{0.8375({\color{red}-0.0072})}
& 0.7378({\color{blue}+0.0482})
& \multicolumn{1}{>{\columncolor{darkgray}}c}{0.6795({\color{blue}+0.0217})}
& \multicolumn{1}{>{\columncolor{lightgray}}c}{0.7276({\color{blue}+0.0167})}
& 0.7401({\color{blue}+0.0316})\\

LIVE-Qualcomm+LIVE-VQC & 
& \multicolumn{1}{>{\columncolor{darkgray}}c}{0.5849({\color{blue}+0.0574})}
& 0.6843({\color{blue}+0.1063})
& \multicolumn{1}{>{\columncolor{lightgray}}c}{0.8010({\color{red}-0.0007})}
& \multicolumn{1}{>{\columncolor{lightgray}}c}{0.7434({\color{blue}+0.0042})}
& 0.7002({\color{blue}+0.0644})\\

CVD2014+LIVE-Qualcomm+LIVE-VQC & 
& \multicolumn{1}{>{\columncolor{lightgray}}c}{0.8364({\color{red}-0.0072})}
& 0.7152({\color{blue}+0.0641})
& \multicolumn{1}{>{\columncolor{lightgray}}c}{0.8118({\color{red}-0.0073})}
& \multicolumn{1}{>{\columncolor{lightgray}}c}{0.7211({\color{blue}+0.0141})}
& 0.7385({\color{blue}+0.0368})\\
    \midrule 
CVD2014 & \multirow{7}{*}{LIVE-Qualcomm}
& \multicolumn{1}{>{\columncolor{lightgray}}c}{0.8747({\color{red}-0.0111})}
& \multicolumn{1}{>{\columncolor{darkgray}}c}{0.6051({\color{blue}+0.0640})}
& 0.3919({\color{blue}+0.3963})
&\multicolumn{1}{>{\columncolor{darkgray}}c}{0.4950({\color{blue}+0.1203})}
& 0.5846({\color{blue}+0.1019})\\

KoNViD-1k &
& \multicolumn{1}{>{\columncolor{darkgray}}c}{0.6474({\color{blue}+0.0459})}
& \multicolumn{1}{>{\columncolor{lightgray}}c}{0.7809({\color{blue}+0.0026})}
& 0.6732({\color{blue}+0.1437})
& \multicolumn{1}{>{\columncolor{darkgray}}c}{0.7160({\color{red}-0.0192})}
& 0.7398({\color{blue}+0.0146})\\

LIVE-VQC &
& \multicolumn{1}{>{\columncolor{darkgray}}c}{0.4819({\color{blue}+0.1030})}
& \multicolumn{1}{>{\columncolor{darkgray}}c}{0.7059({\color{red}-0.0216})}
& 0.6550({\color{blue}+0.1460})
& \multicolumn{1}{>{\columncolor{lightgray}}c}{0.7470({\color{red}-0.0036})}
& 0.6884({\color{blue}+0.0119})\\

CVD2014+KoNViD-1k &
& \multicolumn{1}{>{\columncolor{lightgray}}c}{0.8552({\color{red}-0.0140})}
& \multicolumn{1}{>{\columncolor{lightgray}}c}{0.7743({\color{red}-0.0084})}
& 0.6484({\color{blue}+0.1673})
& \multicolumn{1}{>{\columncolor{darkgray}}c}{0.6934({\color{red}-0.0083})}
& 0.7498({\color{blue}+0.0074})\\

CVD2014+LIVE-VQC &
& \multicolumn{1}{>{\columncolor{lightgray}}c}{0.8375({\color{red}-0.0010})}
& \multicolumn{1}{>{\columncolor{darkgray}}c}{0.7378({\color{red}-0.0226})}
& 0.6795({\color{blue}+0.1322})
& \multicolumn{1}{>{\columncolor{lightgray}}c}{0.7276({\color{red}-0.0065})}
& 0.7401({\color{red}-0.0016})\\

KoNViD-1k+LIVE-VQC &
& \multicolumn{1}{>{\columncolor{darkgray}}c}{0.6325({\color{blue}+0.0098})}
& \multicolumn{1}{>{\columncolor{lightgray}}c}{0.7974({\color{red}-0.0068})}
& 0.6995({\color{blue}+0.1008})
& \multicolumn{1}{>{\columncolor{lightgray}}c}{0.7461({\color{blue}+0.0014})}
& 0.7575({\color{blue}+0.0072})\\

CVD2014+KoNViD-1k+LIVE-VQC &
& \multicolumn{1}{>{\columncolor{lightgray}}c}{0.8303({\color{red}-0.0010})}
& \multicolumn{1}{>{\columncolor{lightgray}}c}{0.7860({\color{red}-0.0066})}
& 0.7012({\color{blue}+0.1032})
& \multicolumn{1}{>{\columncolor{lightgray}}c}{0.7443({\color{red}-0.0092})}
& 0.7718({\color{blue}+0.0036})\\
    \midrule 
CVD2014 & \multirow{7}{*}{LIVE-VQC}
& \multicolumn{1}{>{\columncolor{lightgray}}c}{0.8747({\color{red}-0.0372})}
& \multicolumn{1}{>{\columncolor{darkgray}}c}{0.6051({\color{blue}+0.1327})}
& \multicolumn{1}{>{\columncolor{darkgray}}c}{0.3919({\color{blue}+0.2876})}
& 0.4950({\color{blue}+0.2326})
& 0.5846({\color{blue}+0.1555})\\

KoNViD-1k &
& \multicolumn{1}{>{\columncolor{darkgray}}c}{0.6474({\color{red}-0.0149})}
& \multicolumn{1}{>{\columncolor{lightgray}}c}{0.7809({\color{blue}+0.0165})}
& \multicolumn{1}{>{\columncolor{darkgray}}c}{0.6732({\color{blue}+0.0263})}
& 0.7160({\color{blue}+0.0301})
& 0.7398({\color{blue}+0.0177})\\

LIVE-Qualcomm &
& \multicolumn{1}{>{\columncolor{darkgray}}c}{0.5879({\color{red}-0.0030})}
& \multicolumn{1}{>{\columncolor{darkgray}}c}{0.6128({\color{blue}+0.0715})}
& \multicolumn{1}{>{\columncolor{lightgray}}c}{0.7538({\color{blue}+0.0472})}
& 0.6214({\color{blue}+0.1220})
& 0.6256({\color{blue}+0.0747})\\

CVD2014+KoNViD-1k &
& \multicolumn{1}{>{\columncolor{lightgray}}c}{0.8552({\color{red}-0.0250})}
& \multicolumn{1}{>{\columncolor{lightgray}}c}{0.7743({\color{blue}+0.0117})}
& \multicolumn{1}{>{\columncolor{darkgray}}c}{0.6484({\color{blue}+0.0528})}
& 0.6934({\color{blue}+0.0509})
& 0.7498({\color{blue}+0.0220})\\

CVD2014+LIVE-Qualcomm &
& \multicolumn{1}{>{\columncolor{lightgray}}c}{0.8636({\color{red}-0.0272})}
& \multicolumn{1}{>{\columncolor{darkgray}}c}{0.6691({\color{blue}+0.0461})}
& \multicolumn{1}{>{\columncolor{lightgray}}c}{0.7883({\color{blue}+0.0235})}
& 0.6153({\color{blue}+0.1058})
& 0.6865({\color{blue}+0.0520})\\

KoNViD-1k+LIVE-Qualcomm &
& \multicolumn{1}{>{\columncolor{darkgray}}c}{0.6933({\color{red}-0.0511})}
& \multicolumn{1}{>{\columncolor{lightgray}}c}{0.7836({\color{blue}+0.0070})}
& \multicolumn{1}{>{\columncolor{lightgray}}c}{0.8170({\color{red}-0.0167})}
& 0.6969({\color{blue}+0.0507})
& 0.7544({\color{blue}+0.0102})\\

CVD2014+KoNViD-1k+LIVE-Qualcomm &
& \multicolumn{1}{>{\columncolor{lightgray}}c}{0.8412({\color{red}-0.0119})}
& \multicolumn{1}{>{\columncolor{lightgray}}c}{0.7659({\color{blue}+0.0135})}
& \multicolumn{1}{>{\columncolor{lightgray}}c}{0.8157({\color{red}-0.0113})}
& 0.6851({\color{blue}+0.0501})
& 0.7572({\color{blue}+0.0181})\\
    \bottomrule
    \end{tabular}
    }
    \end{small}
\end{table*}

\subsection{Effectiveness of Mixed Datasets Training Strategy}
\label{sec:ablation}
In this subsection, we conduct experiments to verify the effectiveness of our mixed datasets training strategy in the following four aspects. 
We first consider different loss combinations in our strategy. 
Then, we compare our strategy with the na\"ive linear re-scaling strategy. 
In the third and fourth aspects, we exploit whether our strategy helps further improving the performance with more training data available.

\begin{figure}[!thb]
    \centering
    \includegraphics[width=.95\columnwidth]{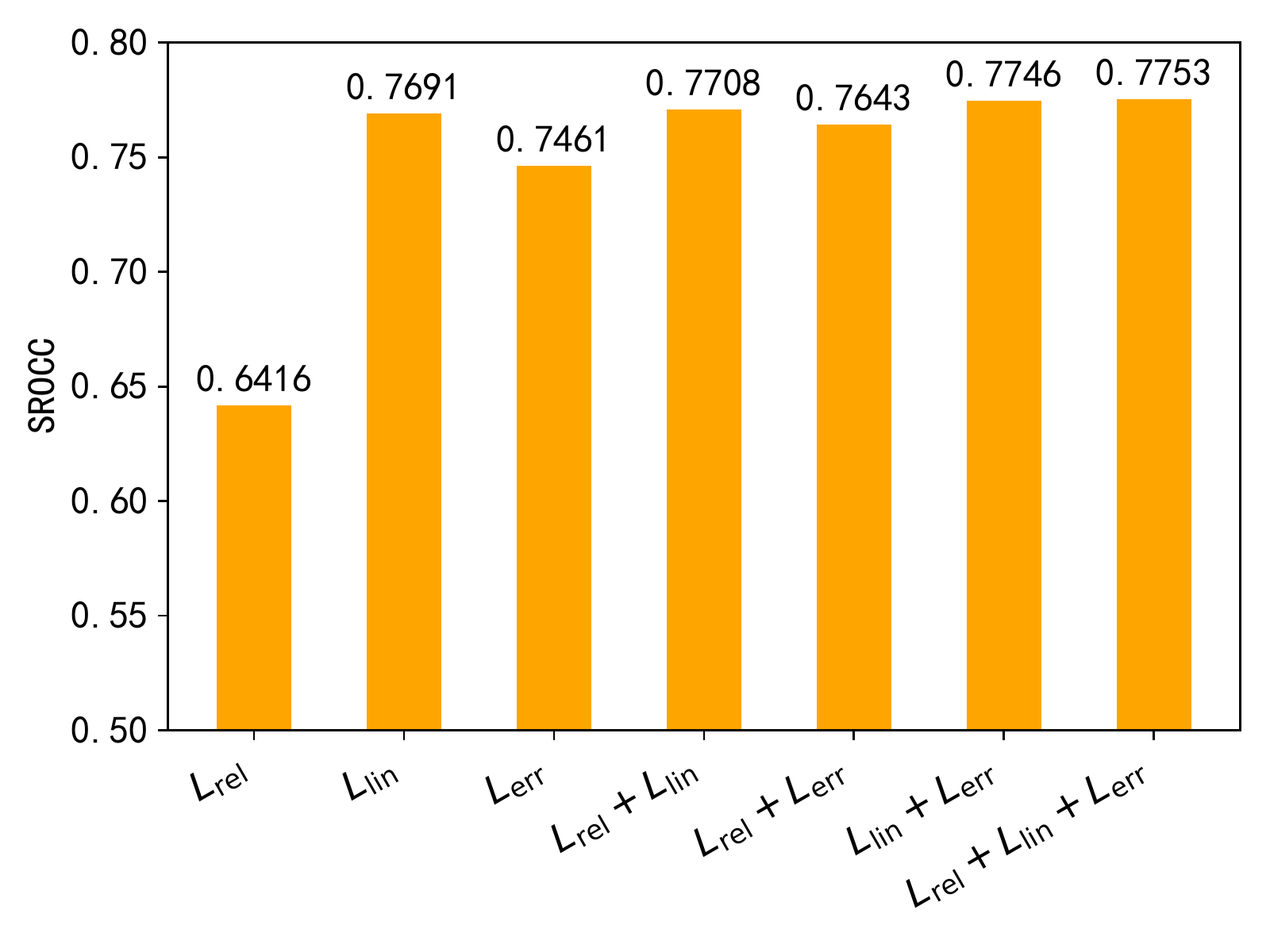}
    \caption{Median SROCC results for different losses used in our mixed datasets training strategy}
    \label{fig:losses}
\end{figure}

\textbf{Different loss combinations}.
To verify the effectiveness of the proposed losses, we compare different combinations of monotonicity-induced loss $L_{\mathrm{rel}}$, linearity-induced loss $L_{\mathrm{lin}}$, and error-induced loss $L_{\mathrm{err}}$. 
We consider mixing all the four datasets (CVD2014, KoNViD-1k, LIVE-Qualcomm, and LIVE-VQC) in this experiment. 
Fig.~~\ref{fig:losses} shows the dataset-size weighted average of median SROCC results over the four datasets. 
It can be seen that the combination of three losses is better than that of two losses, and  the combination of two losses is better than one of the two losses only.
The three losses all contribute to the performance gain, but the contribution of linearity-induced loss is the largest.

\begin{figure}[!thb]
    \centering
    \includegraphics[width=.85\columnwidth]{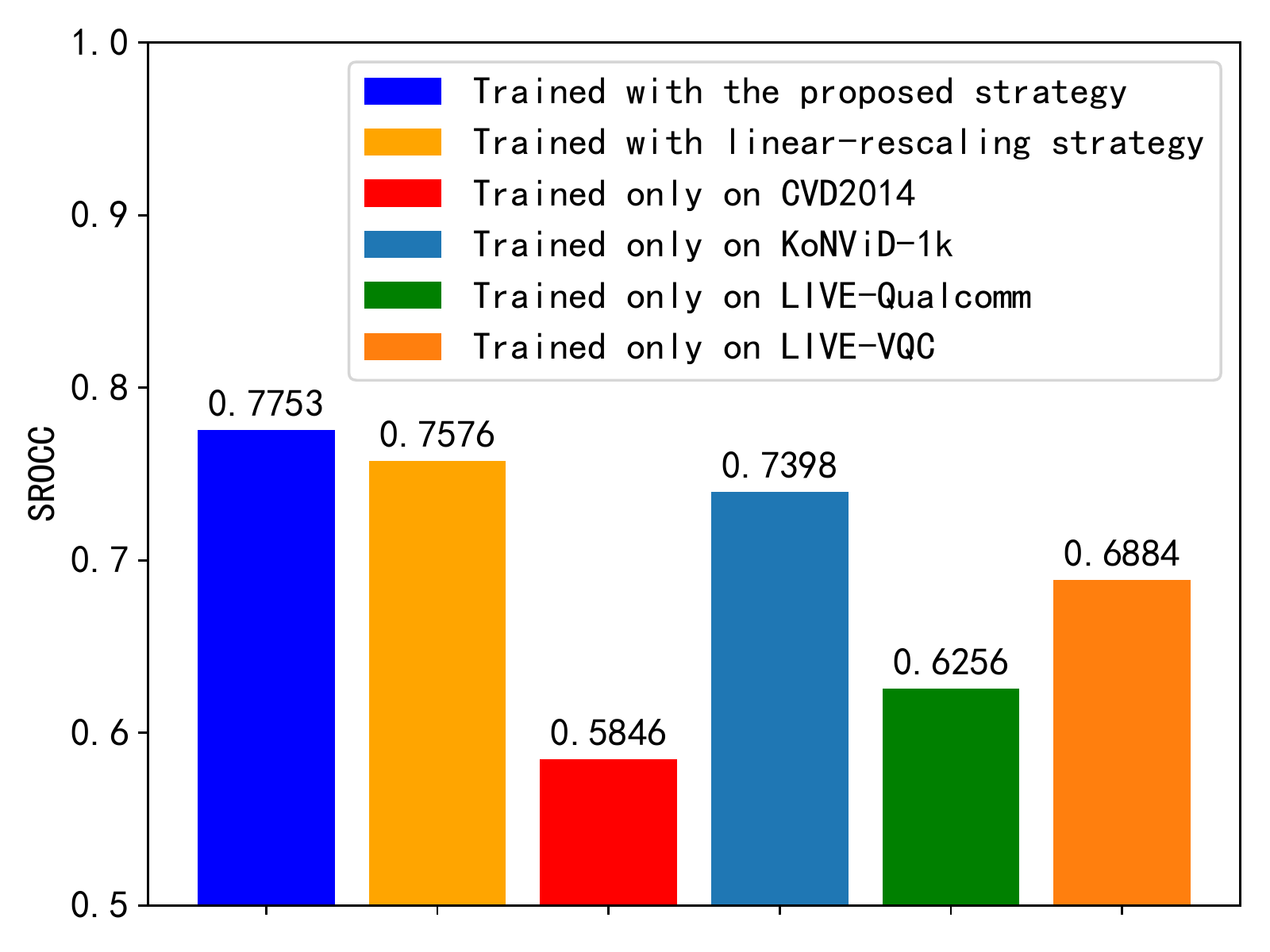}
    \caption{Median SROCC results for models trained with our strategy and the linear re-scaling strategy in comparison with the models trained only on one of the datasets}
    \label{fig:vs naive}
\end{figure}

\textbf{Comparison with linear re-scaling}.
To verify the effectiveness of our dataset-specific perceptual scale alignment, we compare it with the na\"ive linear re-scaling.
Similar to the last experiment, all the four datasets (CVD2014, KoNViD-1k, LIVE-Qualcomm, and LIVE-VQC) are considered. 
And both our strategy and the linear re-scaling strategy use all three losses.
They are compared with the models trained on one of the datasets, \textit{i.e.}, ``Trained only on CVD2014/KonViD-1k/LIVE-Qualcomm/LIVE-VQC".
Fig.~\ref{fig:vs naive} shows the dataset-size weighted average of median SROCC results over the four datasets. 
Models trained on the two larger datasets (KoNViD-1k and LIVE-VQC) achieve better performance than models trained on the two smaller datasets (CVD2014 and LIVE-Qualcomm).
Linear rescaling strategy improves the performance to 0.7576, and our mixed datasets training strategy further improves the performance to 0.7753.
The further performance gain is contributed by the dataset-specific perceptual scale alignment for avoiding the inconformity due to linear re-scaling.

\begin{table}[!thb]
    \centering
    \caption{The test performance of a model trained only on a single train set}
    \label{tab:corr}

    \begin{small}
    
    \resizebox{\columnwidth}{!}{
    \begin{tabular}{lccccc}
    \toprule
   \diagbox{Train}{SROCC}{Test}  & CVD2014 & KoNViD-1k & LIVE-Qualcomm & LIVE-VQC \\
    \midrule 
CVD2014 
& \textbf{0.8747}
& 0.6051
& 0.3919
& 0.4950\\
KoNViD-1k
& 0.6474
& \textbf{0.7809}
& 0.6732
& 0.7160\\
LIVE-Qualcomm 
& 0.5879
& 0.6128
& \textbf{0.7538}
& 0.6214\\
LIVE-VQC
& 0.4819
& 0.7059
& 0.6550
& \textbf{0.7470}\\

    \bottomrule
    \end{tabular}
    }
    \end{small}
\end{table}

\begin{table*}[!thb]
    \centering
    \caption{Cross dataset performance gain in terms of median SROCC when KoNViD-1k is added into the training data. Note that the testing is conducted on the full dataset, including its train and test sets. }
    \label{tab:mixing more datasets-cross}

    \begin{small}
    
    \begin{tabular}{lccc}
    \toprule
    \multirow{2}{*}{Train data} & \multicolumn{3}{c}{Test dataset} \\
    & CVD2014 (full) & LIVE-Qualcomm (full) & LIVE-VQC (full) \\
    \midrule 
CVD2014 ({\color{blue}+KoNViD-1k})
& -
& 0.3390({\color{blue}+0.2579})
& 0.4751({\color{blue}+0.2128})\\

LIVE-Qualcomm ({\color{blue}+KoNViD-1k})
& 0.4938({\color{blue}+0.1488})
& - 
& 0.5988({\color{blue}+0.0983})\\

LIVE-VQC ({\color{blue}+KoNViD-1k})
& 0.4662({\color{blue}+0.1584})
& 0.5888({\color{blue}+0.0521})
& - \\

CVD2014+LIVE-Qualcomm ({\color{blue}+KoNViD-1k})
& -
& - 
& 0.5984({\color{blue}+0.0821})\\

CVD2014+LIVE-VQC ({\color{blue}+KoNViD-1k})
& -
& 0.6459({\color{blue}+0.0087})
& -\\

LIVE-Qualcomm+LIVE-VQC ({\color{blue}+KoNViD-1k})
& 0.5069({\color{blue}+0.1178})
& -
& -\\

    \bottomrule
    \end{tabular}
    \end{small}
\end{table*}

\textbf{Mixing more datasets}.
In this experiment, we explore the effect of mixing more datasets into the training data.
Table~\ref{tab:mixing more datasets} shows the median SROCC results in 10 runs for mixing different datasets.
Each cell shows the base performance of the model trained on the train sets of $D_B$ and tested on a test set of $D_T$, and the value in the brackets indicates the performance gain when the train set of $D_{+}$ is added into the training data.
The ``Overall Performance'' is the dataset-size weighted average of median SROCC over these datasets.
In general, the overall performance over the four datasets is improved in most cases. 
As for the performance on a single test set, there are mainly three scenarios.
\begin{itemize}
\item $D_T = D_{+}$:
The performance values in this scenario, shown in the diagonal blocks of Table~\ref{tab:mixing more datasets}, all increase a lot. 
For example, when the train set of CVD2014 ($D_{+}$) is added into any train sets of $D_B$, the performance on the test set of CVD2014 ($D_{T}$) is improved (0.1479+ gain). 
\item $D_T\subseteq D_B$: 
The performance values in this scenario, marked in a light gray background, mostly decrease a little. 
For example, when the train set of CVD2014 ($D_{+}$) is added into the train set of KoNViD-1k ($D_B$), the performance on the test set of KoNViD-1k ($D_T$) drops 0.0067.
\item $D_T\cap (D_B\cup D_{+})=\emptyset$:
The performance values in this scenario, marked in a dark gray background, may increase or decrease.
For example, when the train set of LIVE-Qualcomm ($D_{+}$) is added into the train set of LIVE-VQC ($D_B$), the performance on the test set of CVD2014 ($D_T$) is improved while that on the test set of KoNViD-1k ($D_T$) drops.
\end{itemize}

This phenomenon is the consequence of the following two factors: (1) over-fitting problem during the training, (2) the discrepancy of data distribution between the train set and the test set.
Table~\ref{tab:corr} shows the performance of the model trained on a single train set and tested on a single test set, which can somehow reflects how well the trained dataset can represent the test set.
In Table~\ref{tab:corr}, the diagonal values are always the largest one in its column, \textit{i.e.}, the most similar data set to a test set is its corresponding train set.
Thus, adding the train set of $D_{+}$ to the train sets of $D_B$ leads to a significant performance improvement on the test set of $D_{+}$, but a minor performance drop on the test sets of $D_B$.
However, we can notice that adding one more train set to the LIVE-Qualcomm train set provides a performance gain on the LIVE-Qualcomm test set.
This might be attributed to the fact that LIVE-Qualcomm is the smallest dataset among these four dataset and overfitting is most likely to happen during model training on LIVE-Qualcomm.
Besides, the performance on the test set of an unseen dataset $D_T$ depends on whether the train set of $D_{+}$ or $D_B$ is more similar to the test set of $D_T$.
In this regard, to improve the model performance on unseen datasets, it is critical to collect similar data for training.
When datasets with similar data distribution to the test set are added into training data, it is more likely to learn the characteristics that are needed for assessing the quality of the test video in the wild. 
For example, in Table~\ref{tab:mixing more datasets-cross}, when KoNViD-1k is added into the training data, the cross-dataset evaluation performance on the unseen dataset is improved.

\begin{figure}[!htb]
    \centering
    \includegraphics[width=.85\columnwidth]{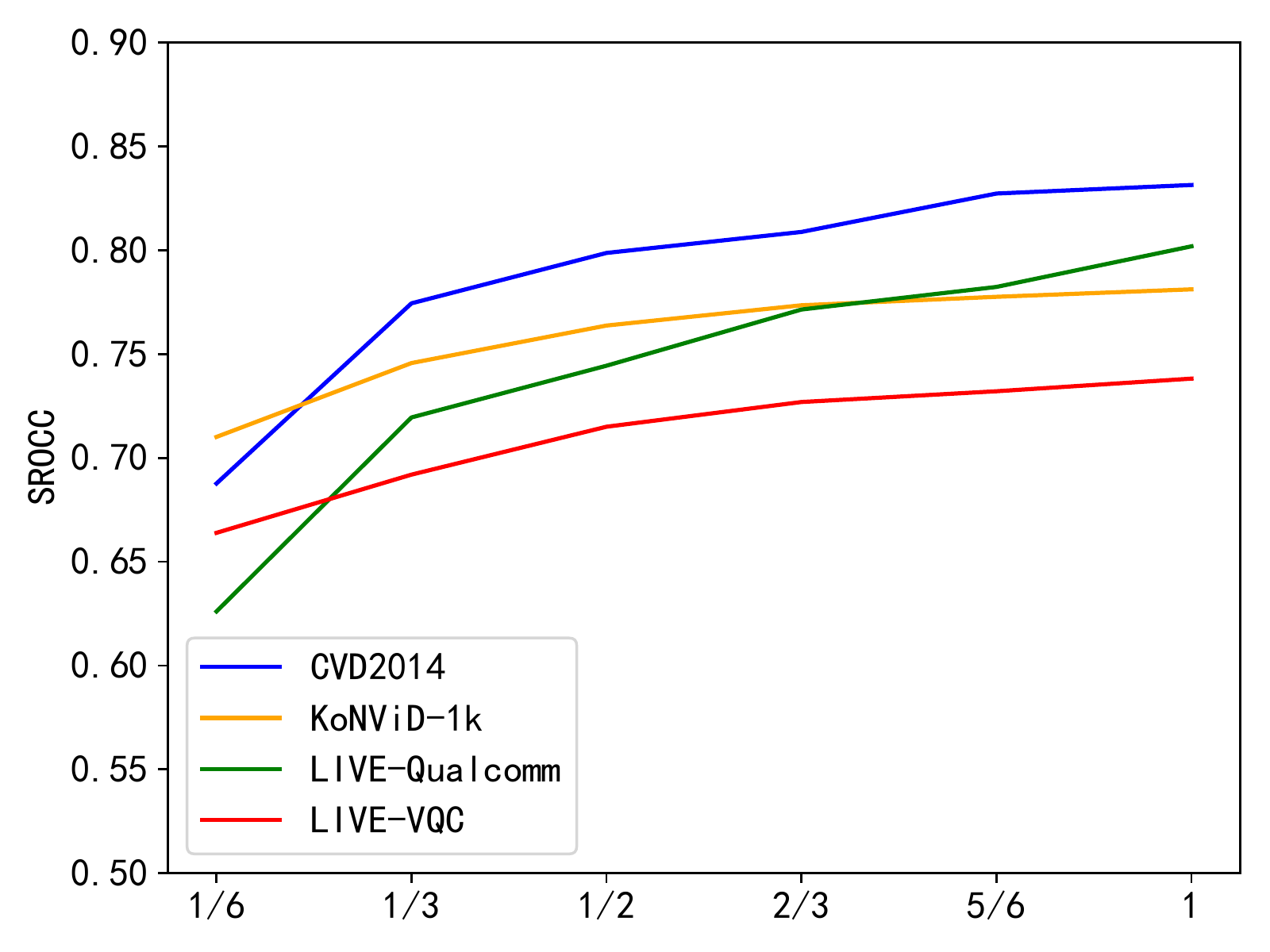}
    \caption{Mean SROCC results under different training proportions when the model is trained by mixing all datasets}
    \label{fig:train proportion}
\end{figure}

\begin{table*}[!hbt]
    \centering
    \caption{Overall performance comparison on CVD2014, KoNViD-1k, and LIVE-Qualcomm. Mean and standard deviation (std) of the dataset-size weighted performance values in 10 runs are reported, \textit{i.e.}, mean ($\pm$ std). The $p$-value is also reported, where $p<0.001$ indicates our method MDTVSFA is significantly better than the method in that row.}
    \label{tab:overall performance}

    \begin{small}
    
    \begin{tabular}{lcccc}
    \toprule
     \multirow{2}{*}{Method} & SROCC$\uparrow$ & $p$-value & PLCC$\uparrow$ & $p$-value\\
     & mean ($\pm$ std) & based on SROCC & mean ($\pm$ std) & based on PLCC \\
    \midrule
    BRISQUE~\citep{mittal2012no} & 0.6610 ($\pm$ 0.0218) & 9.6754E-09 & 0.6032 ($\pm$ 0.0144) & 4.7276E-10\\
    NIQE~\citep{mittal2013making} & 0.5255 ($\pm$ 0.0479)  & 2.3066E-09 & 0.5396 ($\pm$ 0.0430) & 6.4720E-10 \\
    CORNIA~\citep{ye2012unsupervised} &  0.5913 ($\pm$ 0.0253) & 5.4983E-10 & 0.5954 ($\pm$ 0.0240) & 5.0748E-10\\
    VIIDEO~\citep{mittal2016completely} &  0.2368 ($\pm$ 0.0595) & 7.4623E-11 & 0.2351 ($\pm$ 0.0574) & 4.4222E-11 \\
    VBLIINDS~\citep{saad2014blind} &  0.6628 ($\pm$ 0.0321) & 7.7577E-08 & 0.6127 ($\pm$ 0.0833) & 5.1515E-05 \\
    TLVQM~\citep{korhonen2019two} & 0.77 ($\pm$ 0.02)$^*$ & $^*$ & 0.77 ($\pm$ 0.02)$^*$ & $^*$\\
    \midrule
    LS-VSFA &  0.7603 ($\pm$ 0.0219) & 4.0044E-07 &  0.7662 ($\pm$ 0.0238) & 1.9500E-06 \\
    \textbf{MDTVSFA} & \textbf{0.7860} ($\pm$ 0.0202) & - & \textbf{0.7923} ($\pm$ 0.0207) & -\\
    \bottomrule
    \end{tabular}
    \vspace{1mm}
    
    $^*$The results are cited from Table VIII of the original paper~\citep{korhonen2019two}. 
    
    We can not calculate the $p$-value due to the lack of raw SROCC/PLCC values of TLVQM.
    \end{small}
\end{table*}

\textbf{Different training proportions}.
In this experiment, we utilize different proportions of training data from the four datasets (LIVE-VQC, LIVE-Qualcomm, KoNViD-1k, and CVD2014) to train our VQA model with the proposed strategy. 
Fig.~\ref{fig:train proportion} shows the test performance on the four datasets under different training proportions of the training data. 
The performance on each dataset increases as the training proportion increases. 
Our method can still achieve a good performance even when the training proportion is 1/2, which means only half of the training data are used for training. 
And the increasing trend indicates that the performance can still be improved when more training data are available.

Based on the above study, we have learned that our mixed datasets training strategy is effective. 
To sum up, it is helpful for learning characteristics from all datasets and thus improving the overall performance. 
It also has the potential benefits for cross-dataset evaluation since the characteristics of the test videos are more likely to be learned, if more datasets with similar data distribution to the testing set are added into the training data. 
Besides, the performance can be further improved with more training data available.

\subsection{Performance Comparison}
\label{sec:performance}
In this section, we compare our method with the state-of-the-art NR methods.
For VBLIINDS, BRISQUE and our method, we choose the models with the highest SROCC values on the validation set during the training phase. 
NIQE, CORNIA, and VIIDEO are tested on the same 20\% testing data after fitting the four-parameter logistic function with the training data.

\textbf{Overall performance}. 
In this part, all the methods are trained using mixed datasets. 
Similar to~\citet{korhonen2019two}, the other compared methods use the na\"ive linear re-scaling strategy. 
Our model trained with the na\"ive linear re-scaling strategy, denoted as LS-VSFA, does not learn the dataset-specific perceptual scale alignment and uses all three losses after linear re-scaling the subjective quality scores to the same range.
We denote our VQA model trained with the proposed mixed datasets training strategy as MDTVSFA. 
Table~\ref{tab:overall performance} reports the overall performance over CVD2014, KoNViD-1k, and LIVE-Qualcomm, where the overall performance is measured by the dataset-size weighted performance values over the three datasets. 
We can see that our VQA model achieves the best performance in terms of prediction monotonicity (SROCC) and prediction accuracy (PLCC). 
The last two rows show that our proposed mixed datasets training strategy can achieve better performance than the na\"ive linear re-scaling strategy.
We further carry out the statistical significance test to see whether these comparison results are statistical significant or not. 
On each dataset, the paired t-test is conducted at 1\textperthousand~significance level using the performance values (in 10 runs) of our method MDTVSFA and that of the compared one. 
The $p$-values are shown in Table~\ref{tab:overall performance}. 
All $p$-values are far smaller than 0.001 and it proves that our method is significantly better than all the other methods.

\begin{figure*}
\begin{center}
 \includegraphics[height=.25\linewidth, width=.325\linewidth]{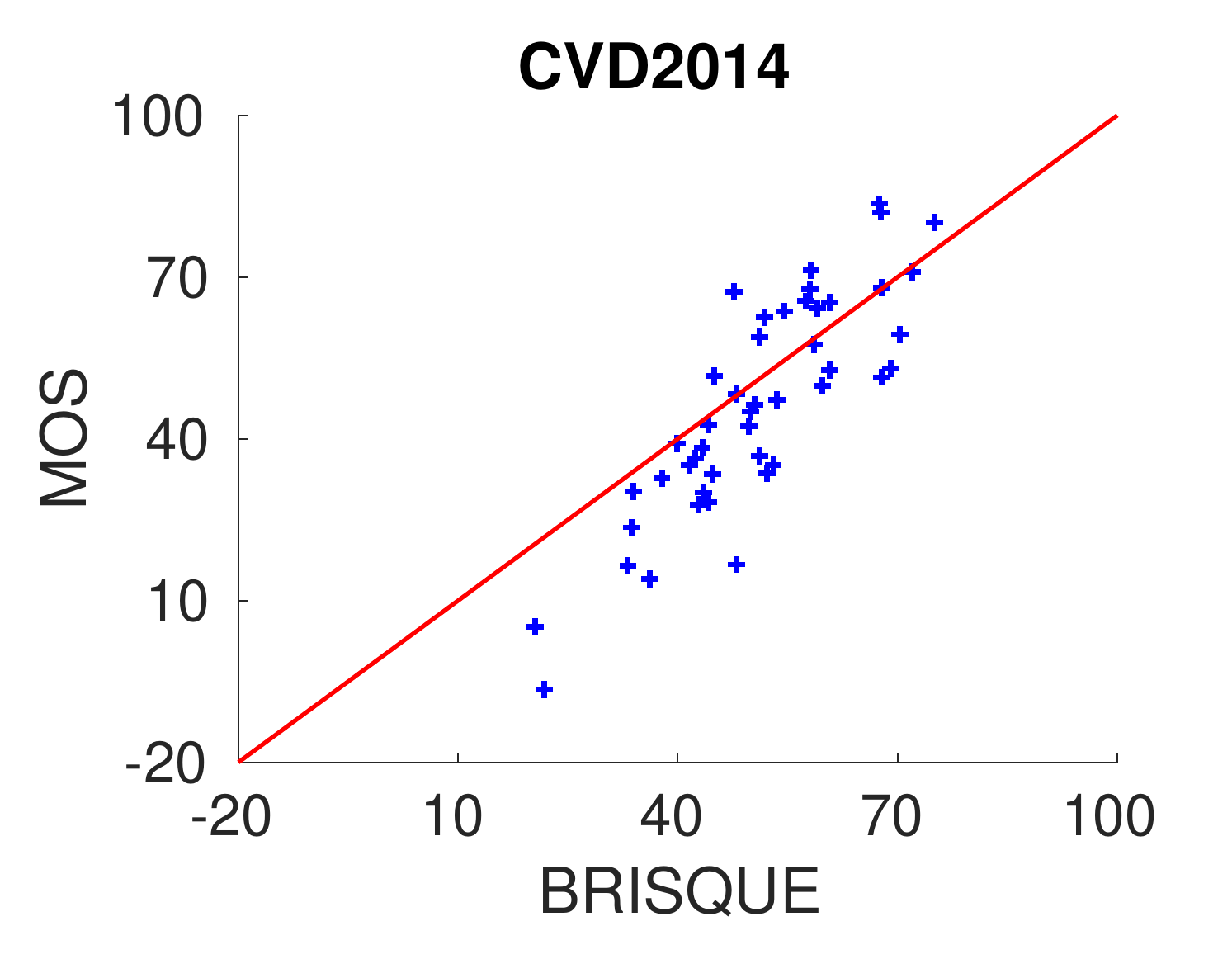}
 \hfill
 \includegraphics[height=.25\linewidth, width=.325\linewidth]{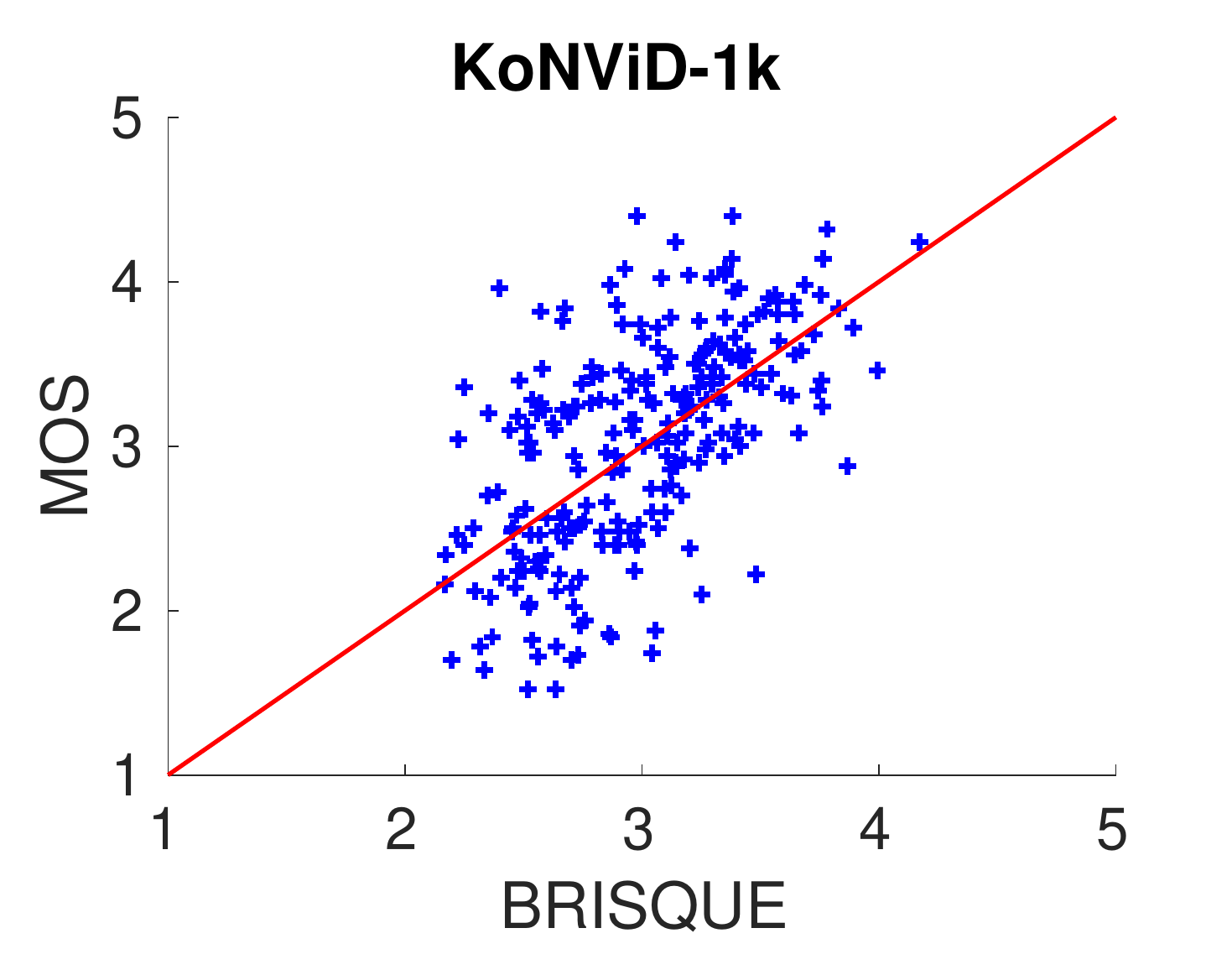}
 \hfill
 \includegraphics[height=.25\linewidth, width=.325\linewidth]{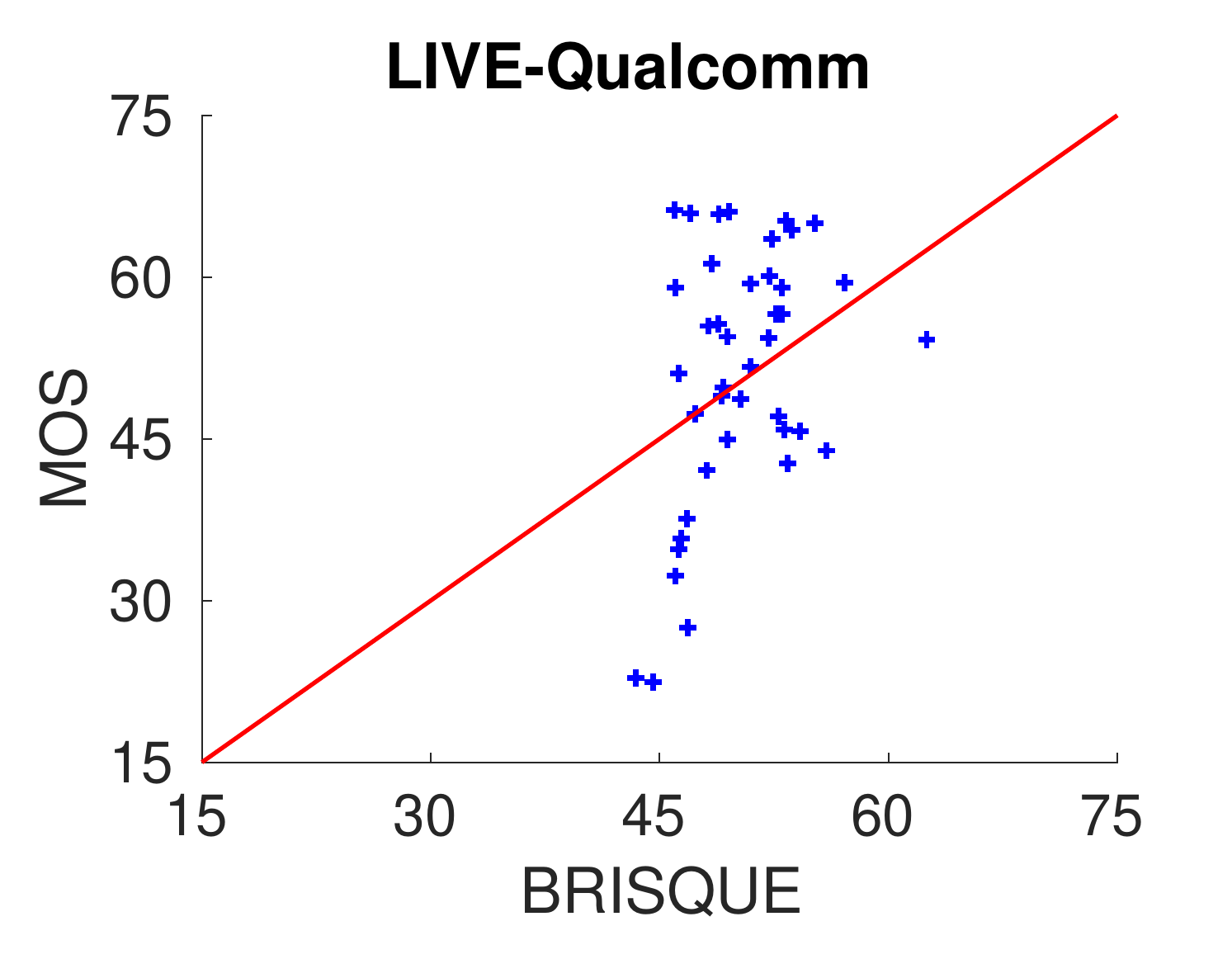}
 \hfill
 \includegraphics[height=.25\linewidth, width=.325\linewidth]{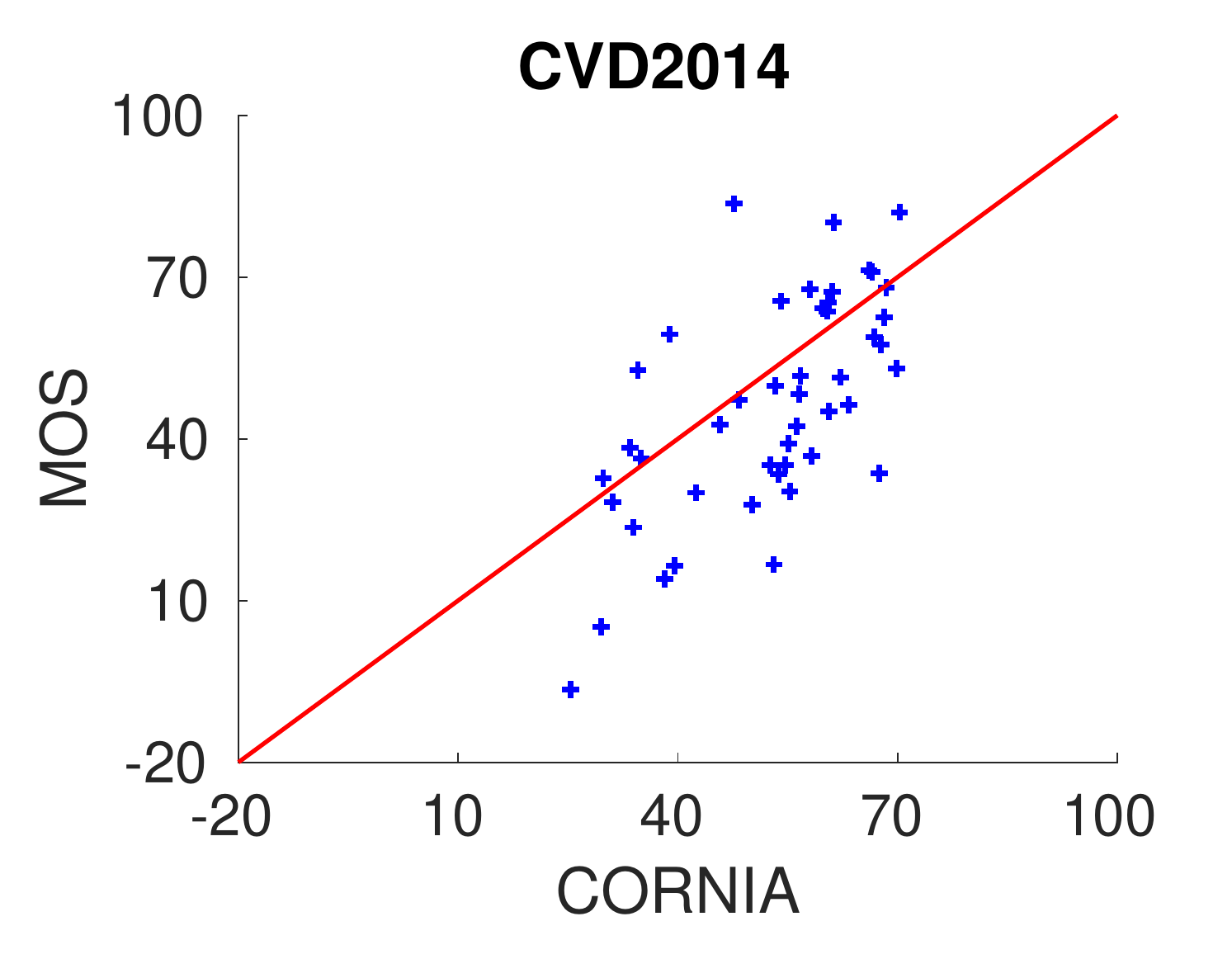}
 \hfill
 \includegraphics[height=.25\linewidth, width=.325\linewidth]{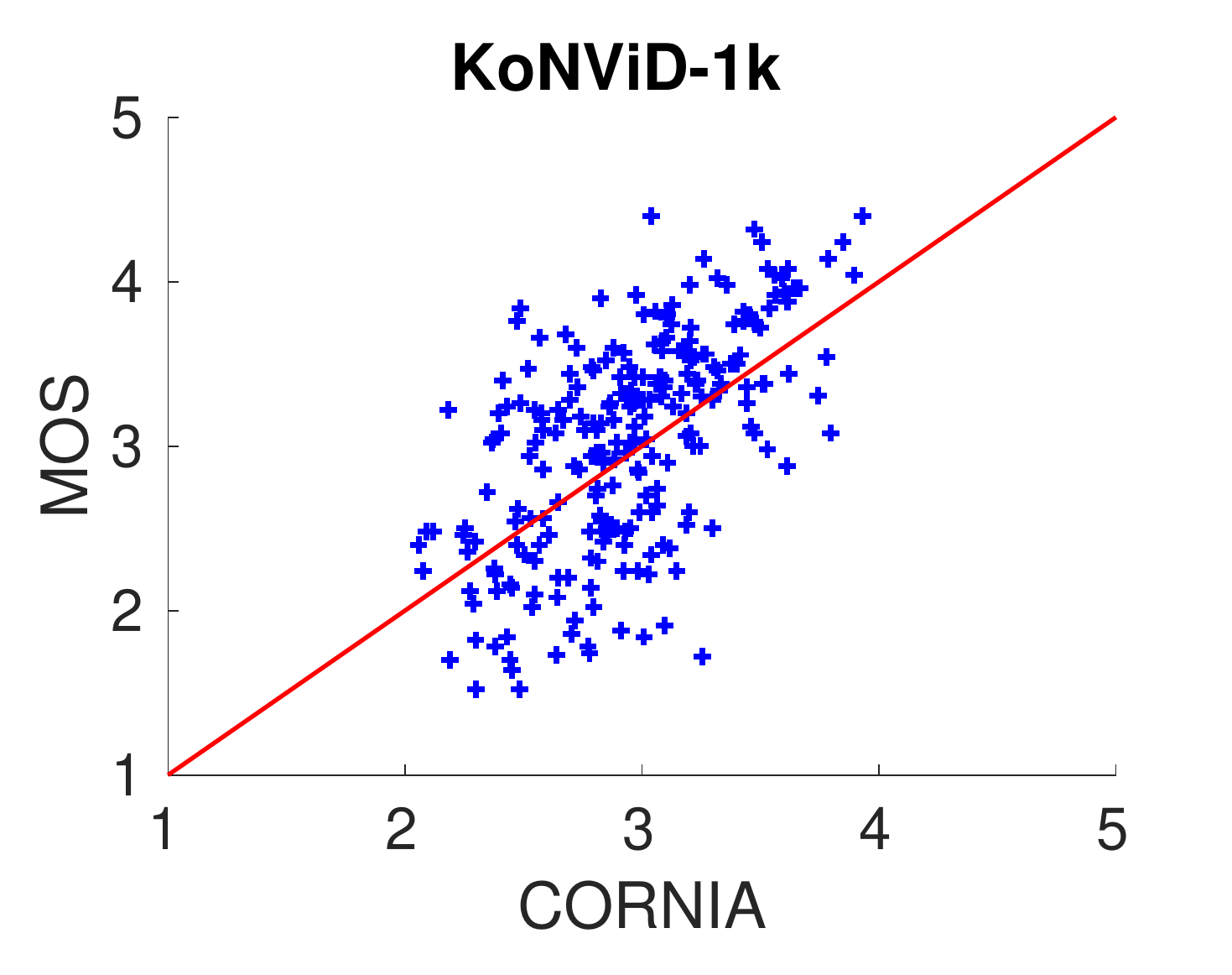}
 \hfill
 \includegraphics[height=.25\linewidth, width=.325\linewidth]{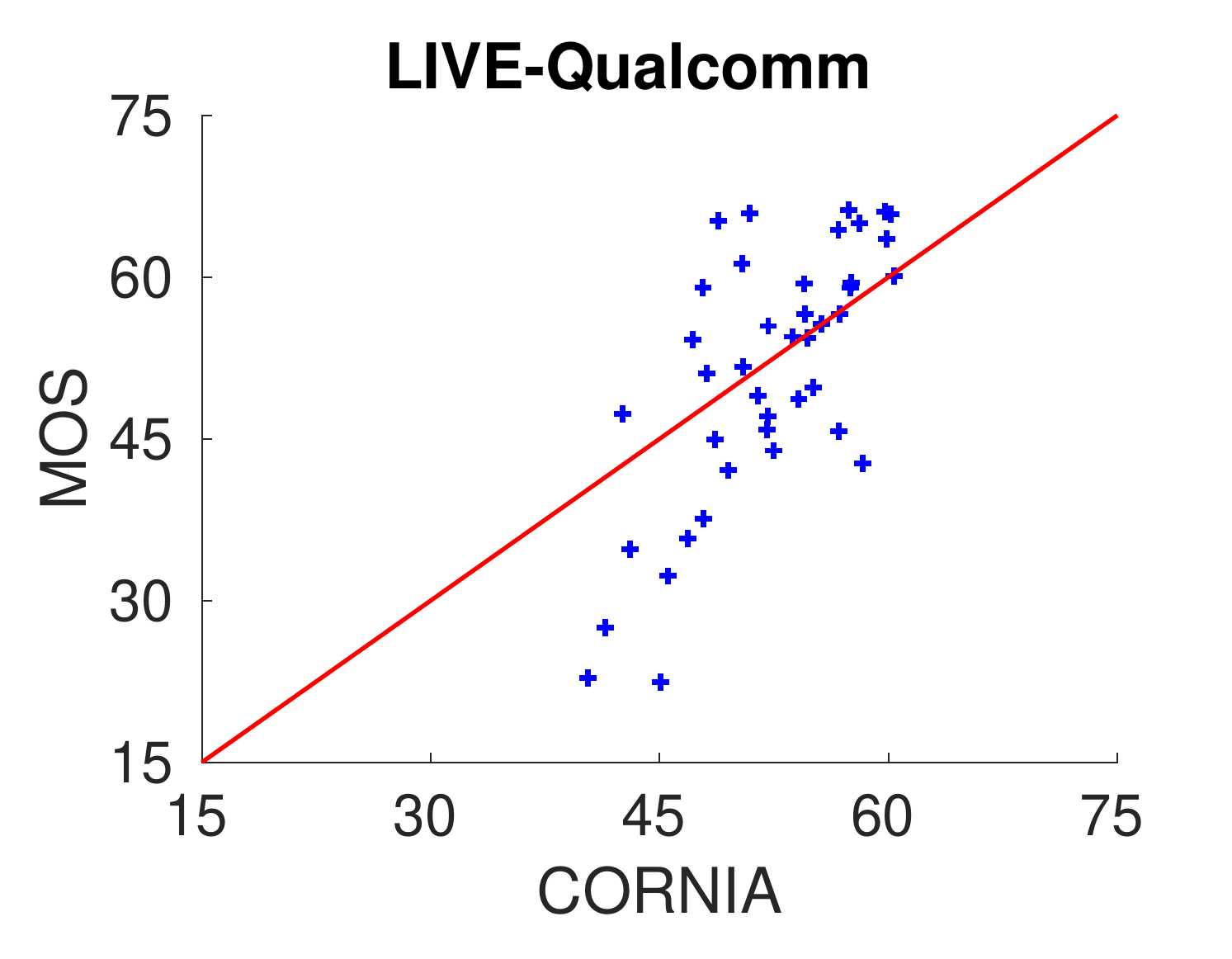}
 \hfill
 \includegraphics[height=.25\linewidth, width=.325\linewidth]{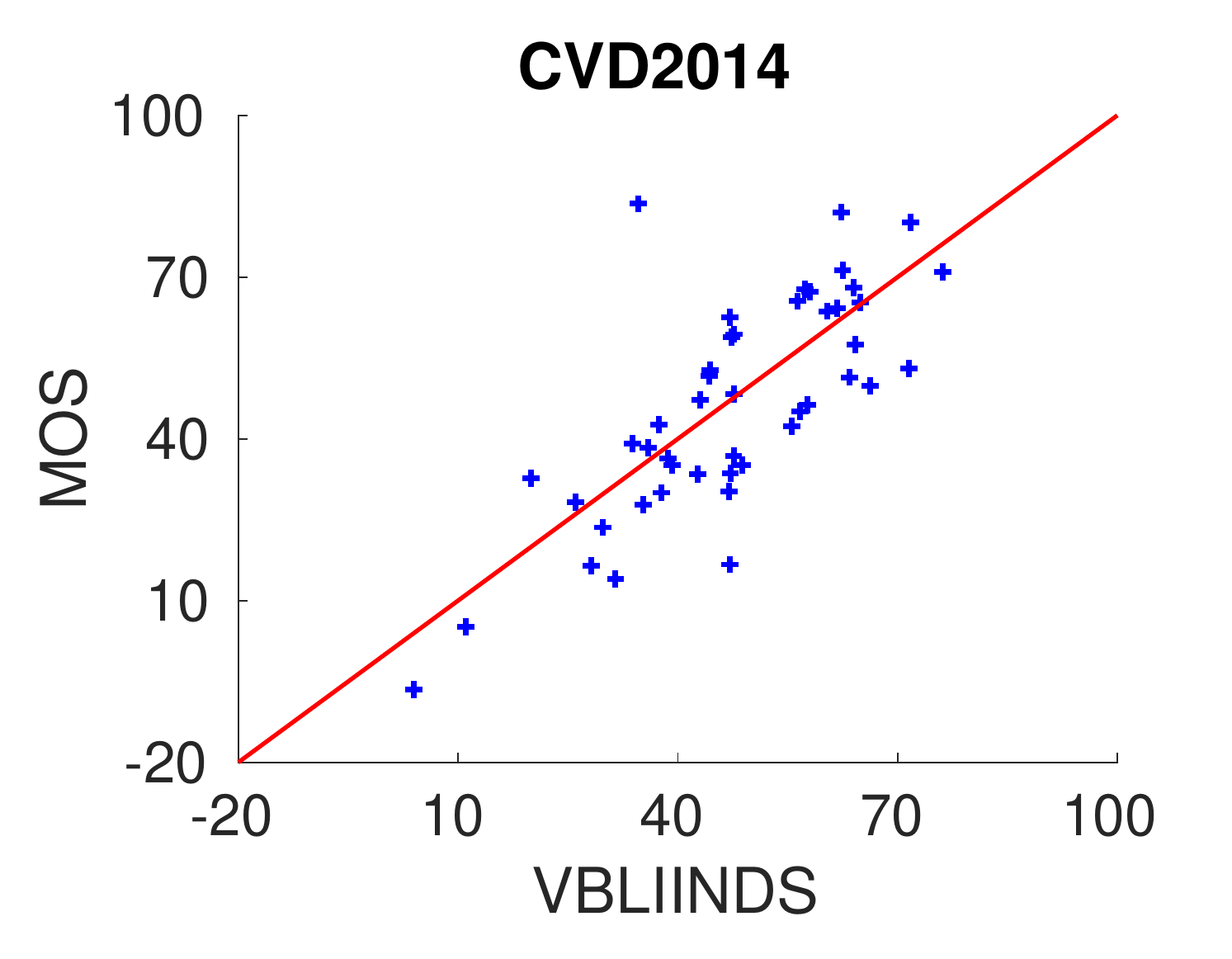}
 \hfill
 \includegraphics[height=.25\linewidth, width=.325\linewidth]{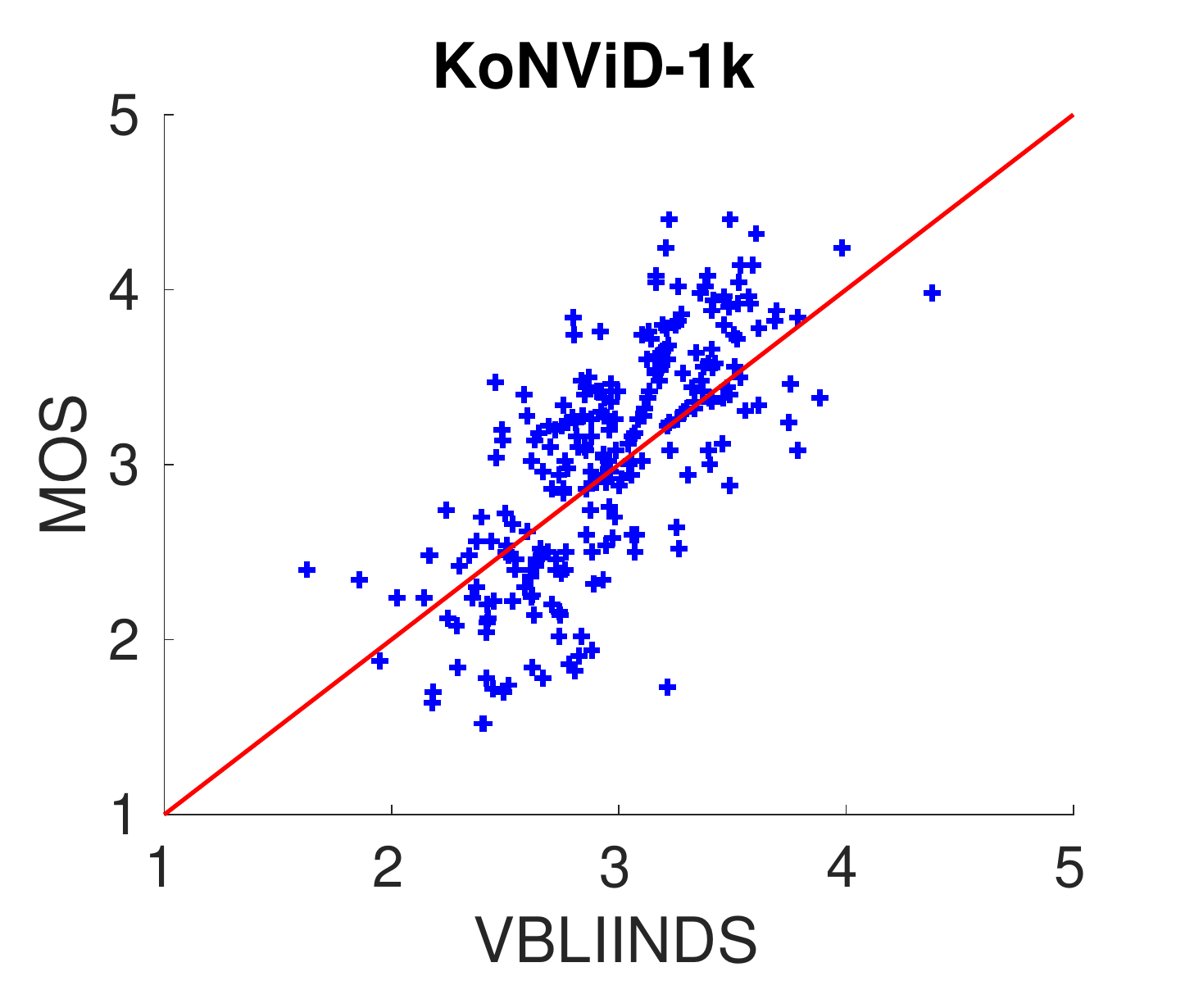}
 \hfill
 \includegraphics[height=.25\linewidth, width=.325\linewidth]{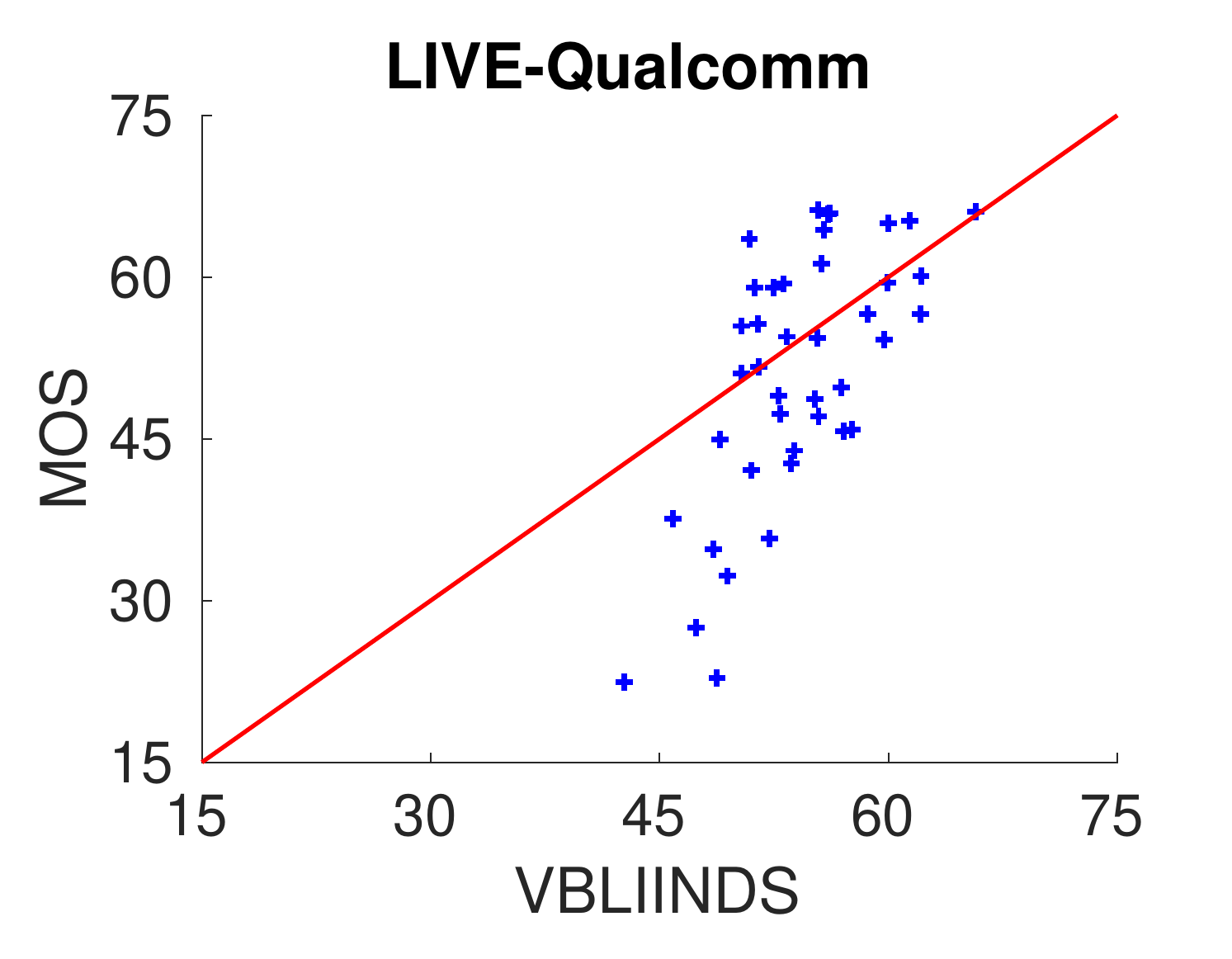}
 \hfill
 \includegraphics[height=.25\linewidth, width=.325\linewidth]{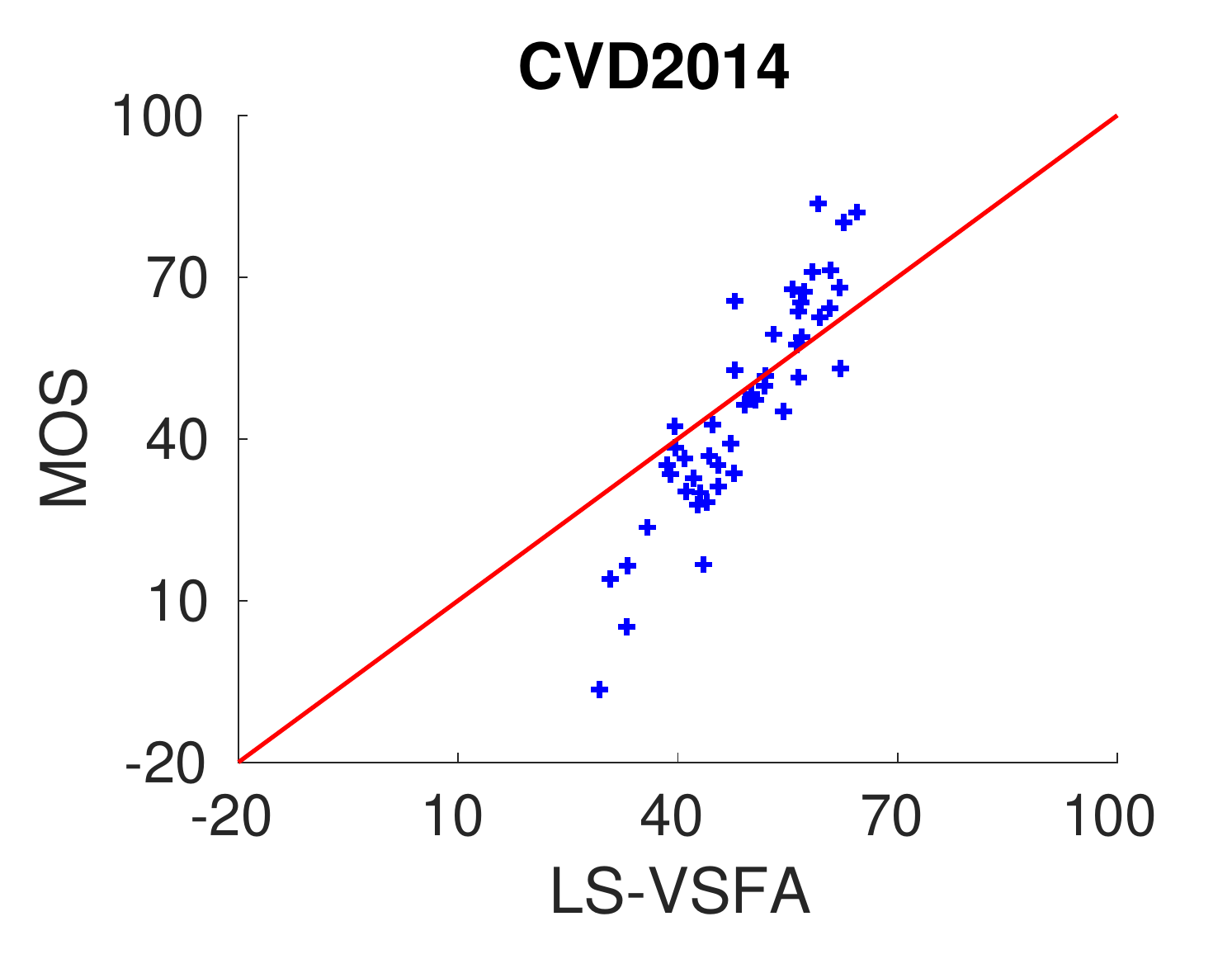}
 \hfill
 \includegraphics[height=.25\linewidth, width=.325\linewidth]{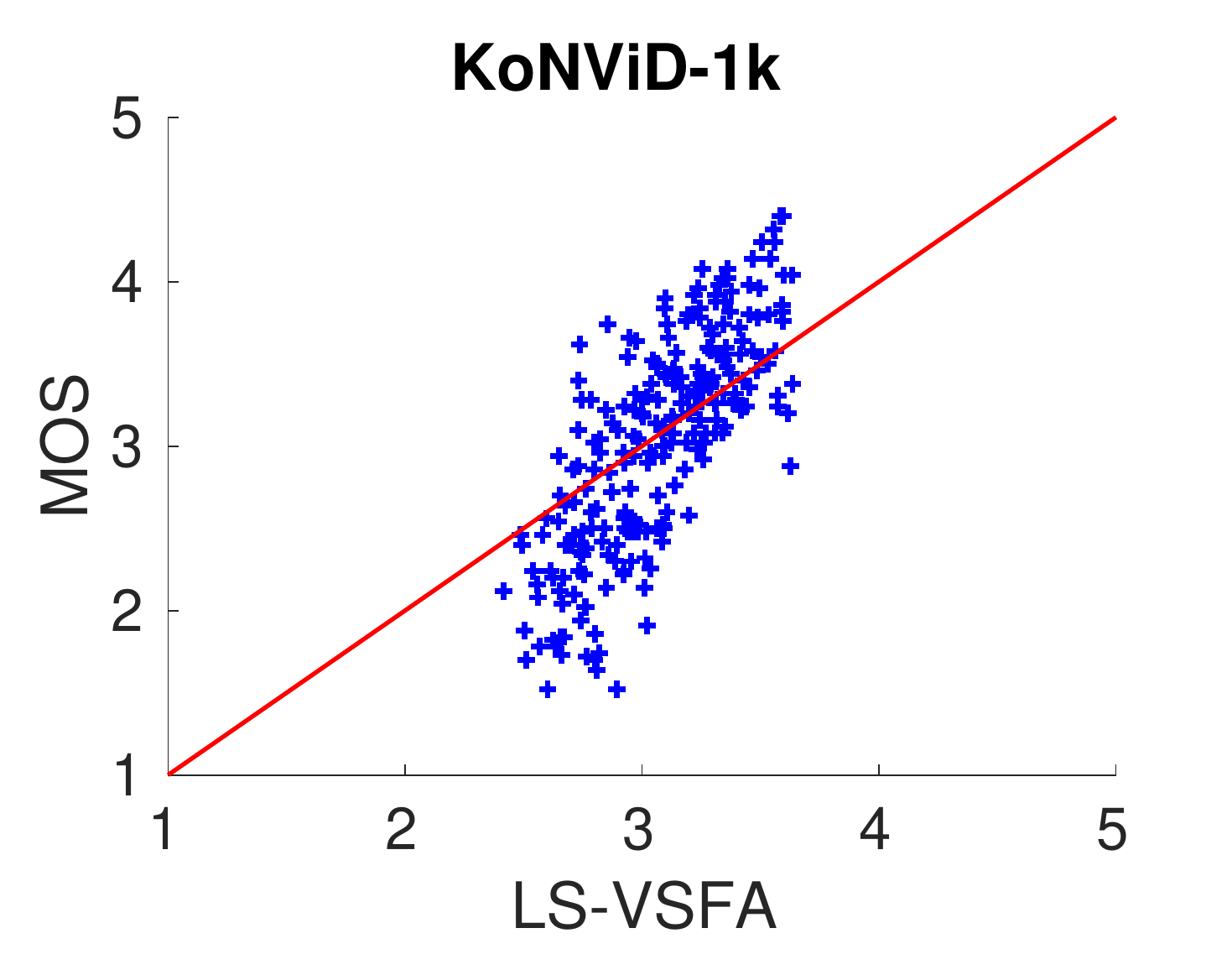}
 \hfill
 \includegraphics[height=.25\linewidth, width=.325\linewidth]{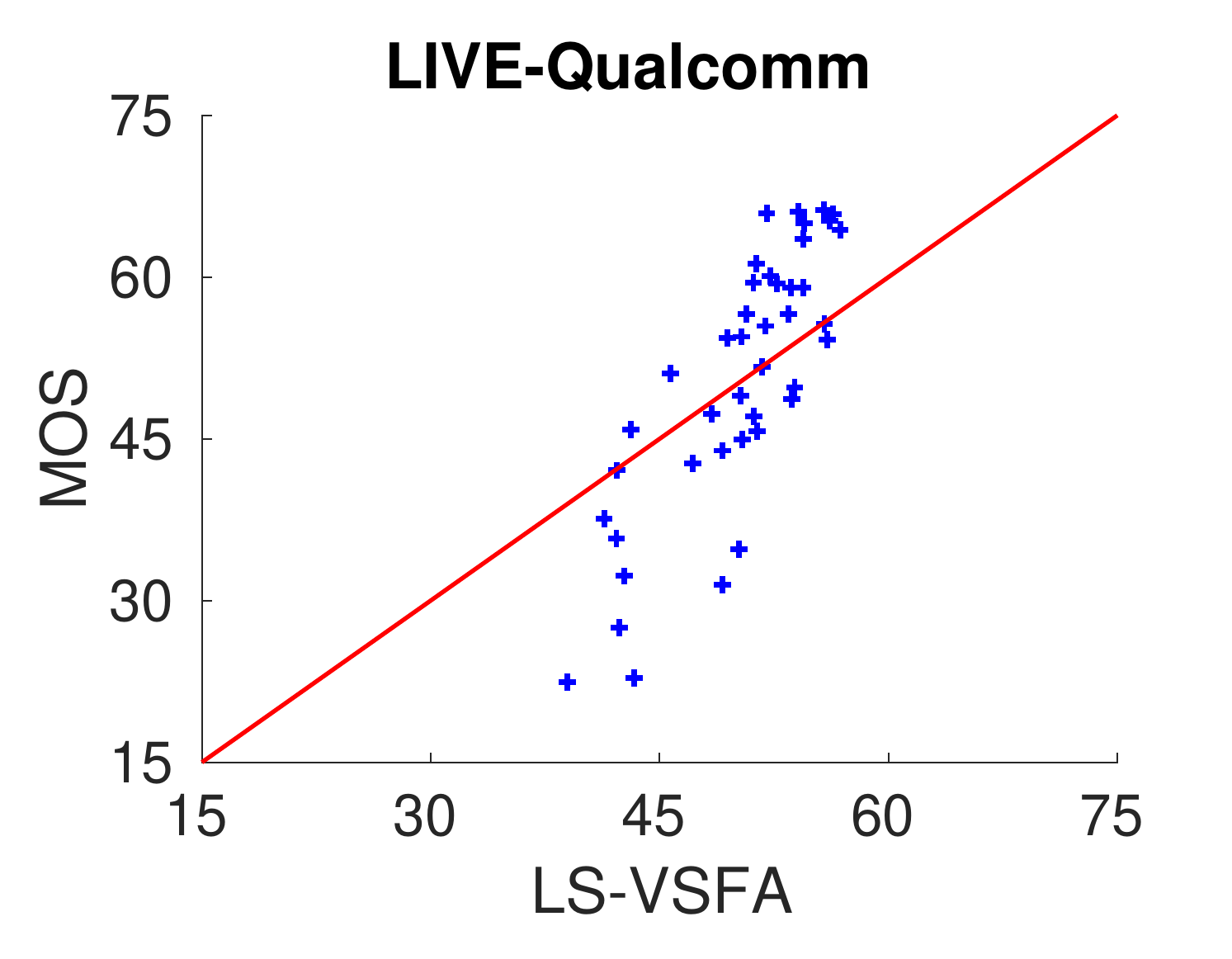}
 \hfill
 \includegraphics[height=.25\linewidth, width=.325\linewidth]{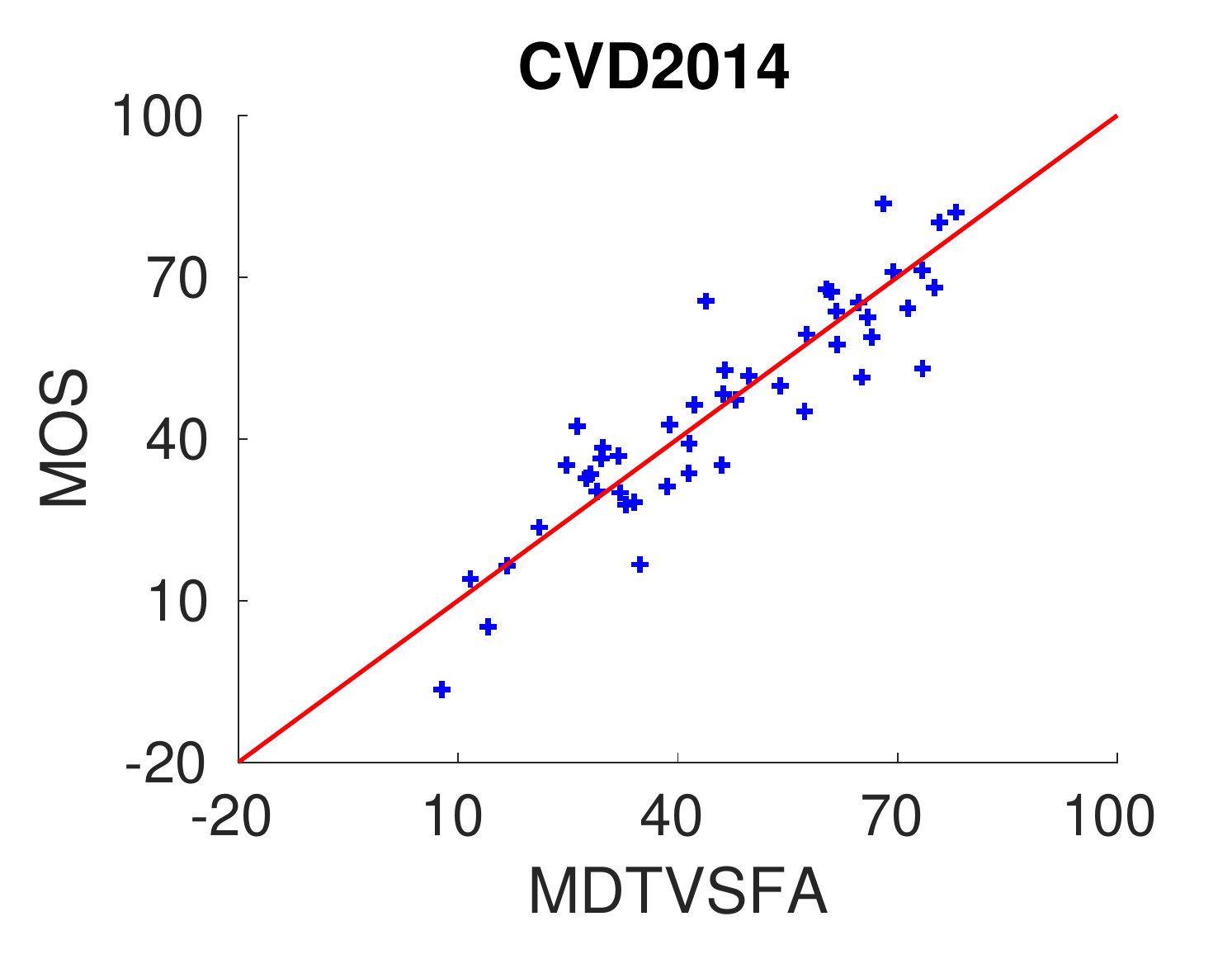}
 \hfill
 \includegraphics[height=.25\linewidth, width=.325\linewidth]{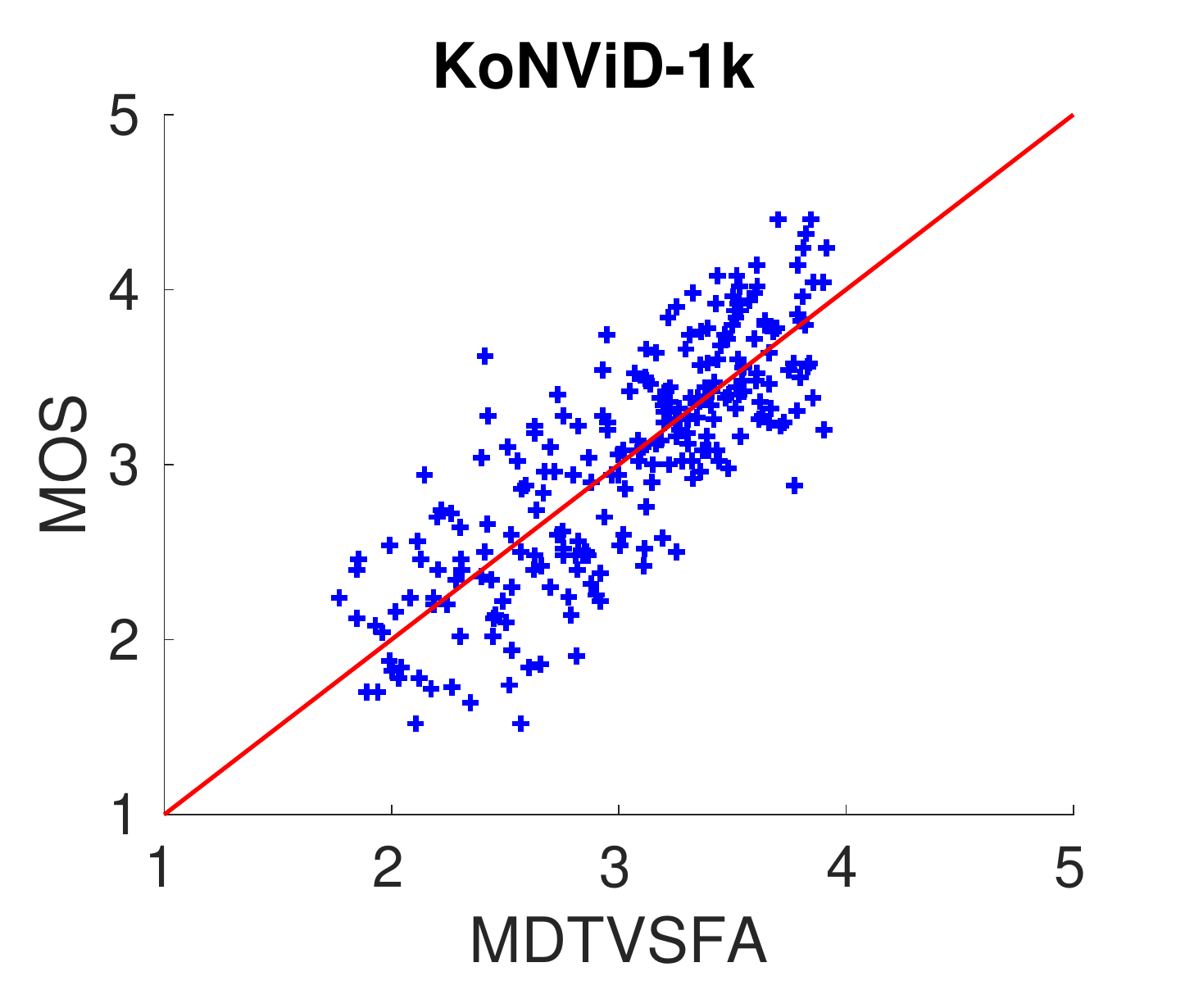}
 \hfill
 \includegraphics[height=.25\linewidth, width=.325\linewidth]{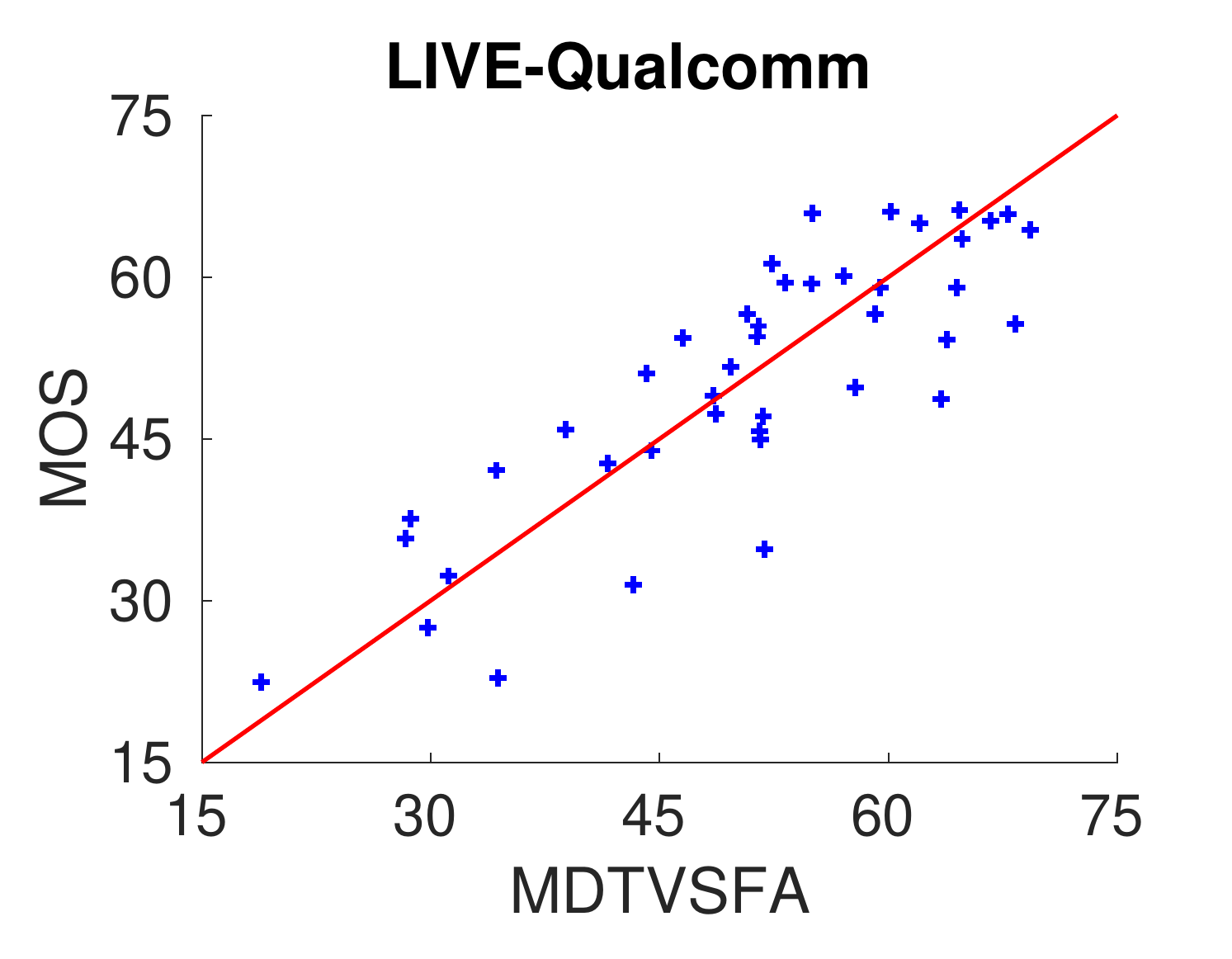}
 \hfill
\end{center}
  \caption{Scatter plots for BRISQUE, CORNIA, VBLIINDS, LS-VSFA, and MDTVSFA on CVD2014, KoNViD-1k, and LIVE-Qualcomm datasets. The predictions of MDTVSFA shows the best correlation with the mean opinion scores (MOSs) across the datasets.} 
 \label{fig:scatter plot}
\end{figure*}

\textbf{Scatter plot and qualitative examples}. 
To have an intuitive feeling, in Fig.~\ref{fig:scatter plot}, we visualize the scatter plots between the subjective scores and predicted scores for the five best-performed methods (excluding TLVQM, since we do not have its raw predictions) in the 10-th run. 
Each row shows the scatter plots for a method. 
From top to down, the methods are BRISQUE, CORNIA, VBLIINDS, LS-VSFA, and MDTVSFA. 
The first, second, and third column show the scatter plots on CVD2014, KoNViD-1k, and LIVE-Qualcomm, respectively.
In each sub-figure, the x-axis indicates the predicted score by the method while y-axis indicates the MOS. 
The scatter points are expected to be located at the diagonal line.
We can see that the scatter plots for BRISQUE and CORNIA are more dispersive than the ones for VBLIINDS and our method.
The scatter points for our method are more densely clustered around and centered at the diagonal line than the others.

\begin{figure}[!htb]
\begin{center}
  \subfloat[Three representative frames of video A on KoNViD-1k]{\includegraphics[width=.325\columnwidth]{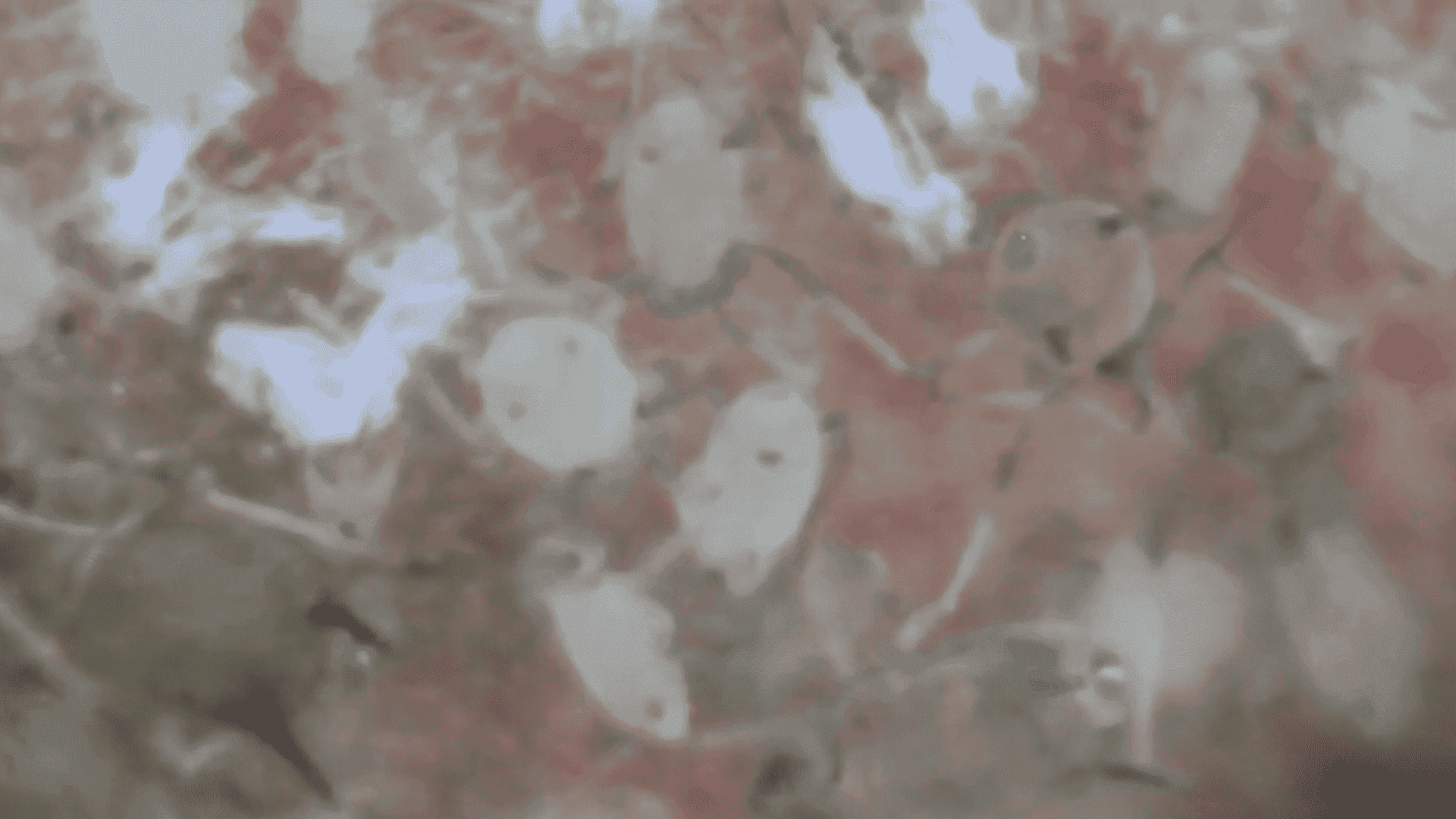} \hfill \includegraphics[width=.325\columnwidth]{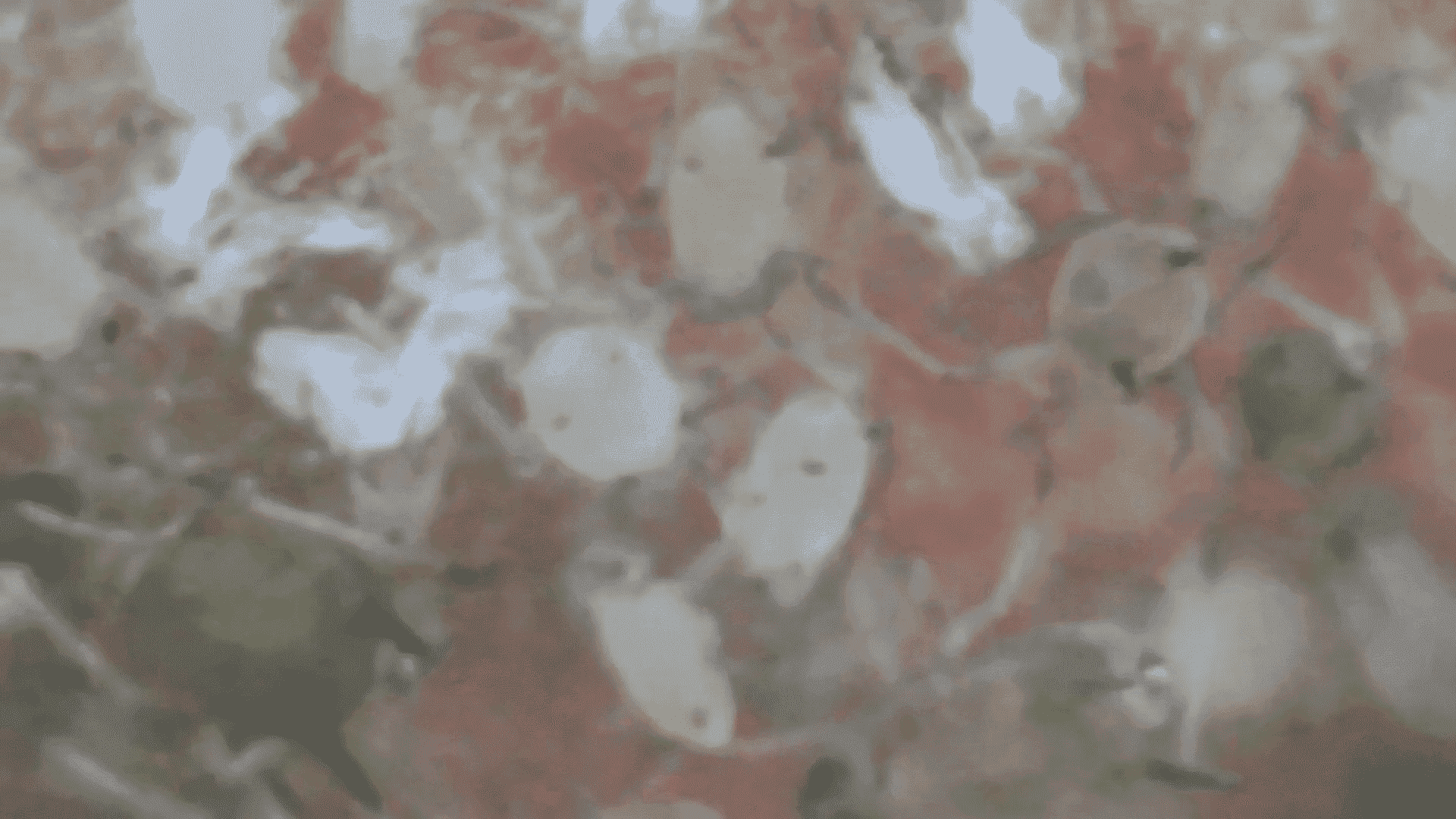} \hfill \includegraphics[width=.325\columnwidth]{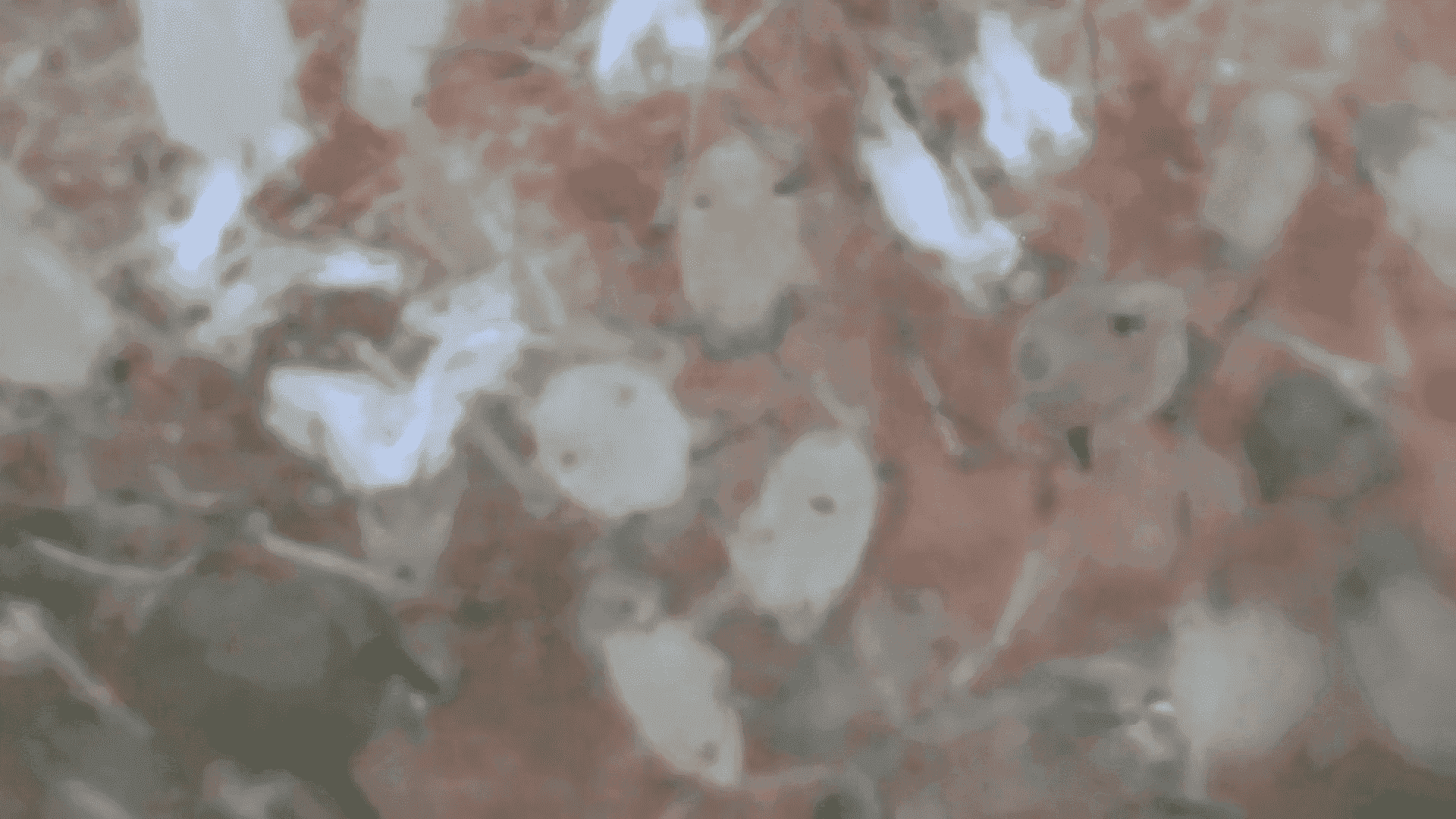}%
  \label{fig:KoNViD-1k-1}}
\hfill  
  \subfloat[Three representative frames of video B on KoNViD-1k]{\includegraphics[width=.325\columnwidth]{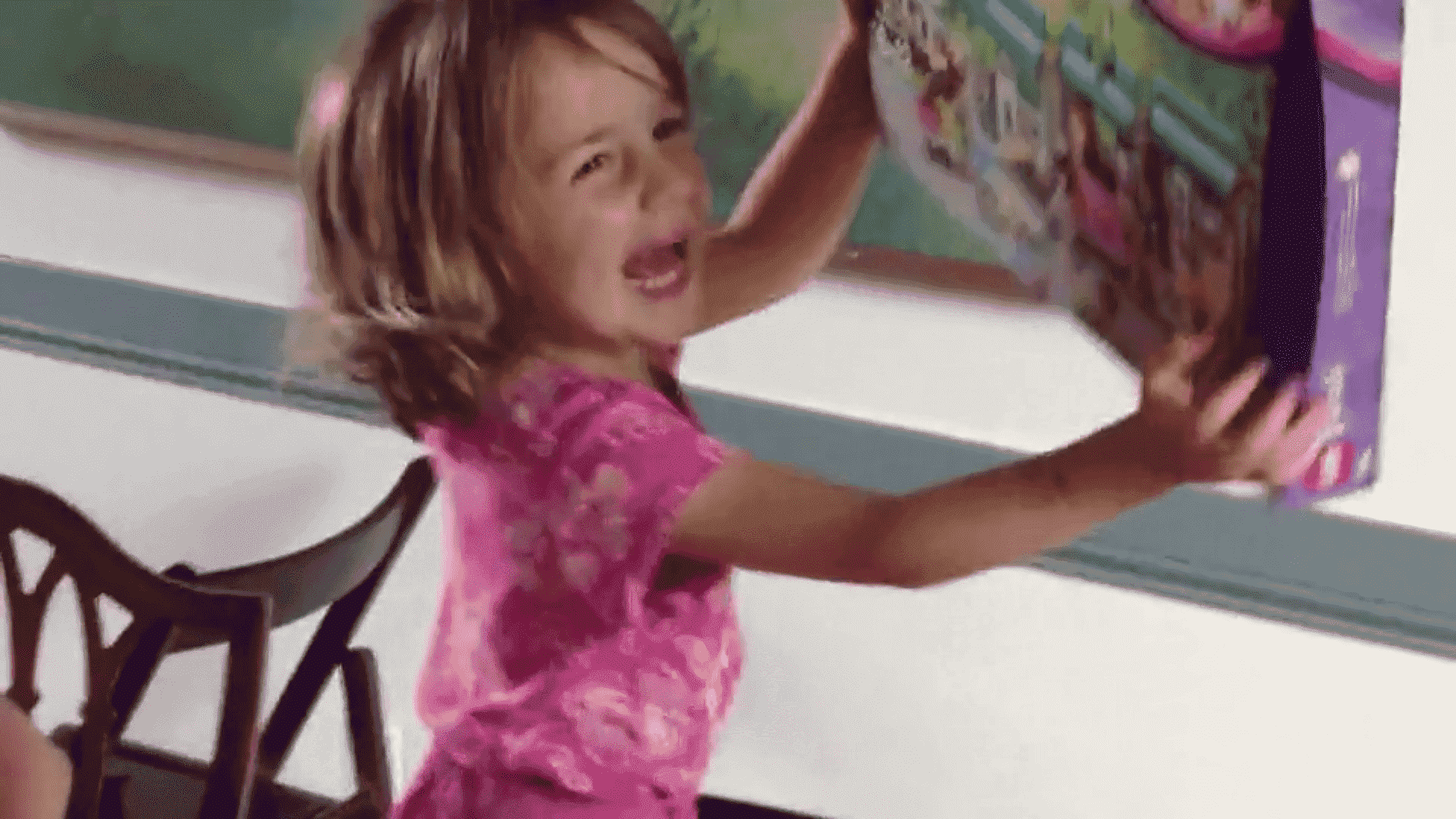} \hfill \includegraphics[width=.325\columnwidth]{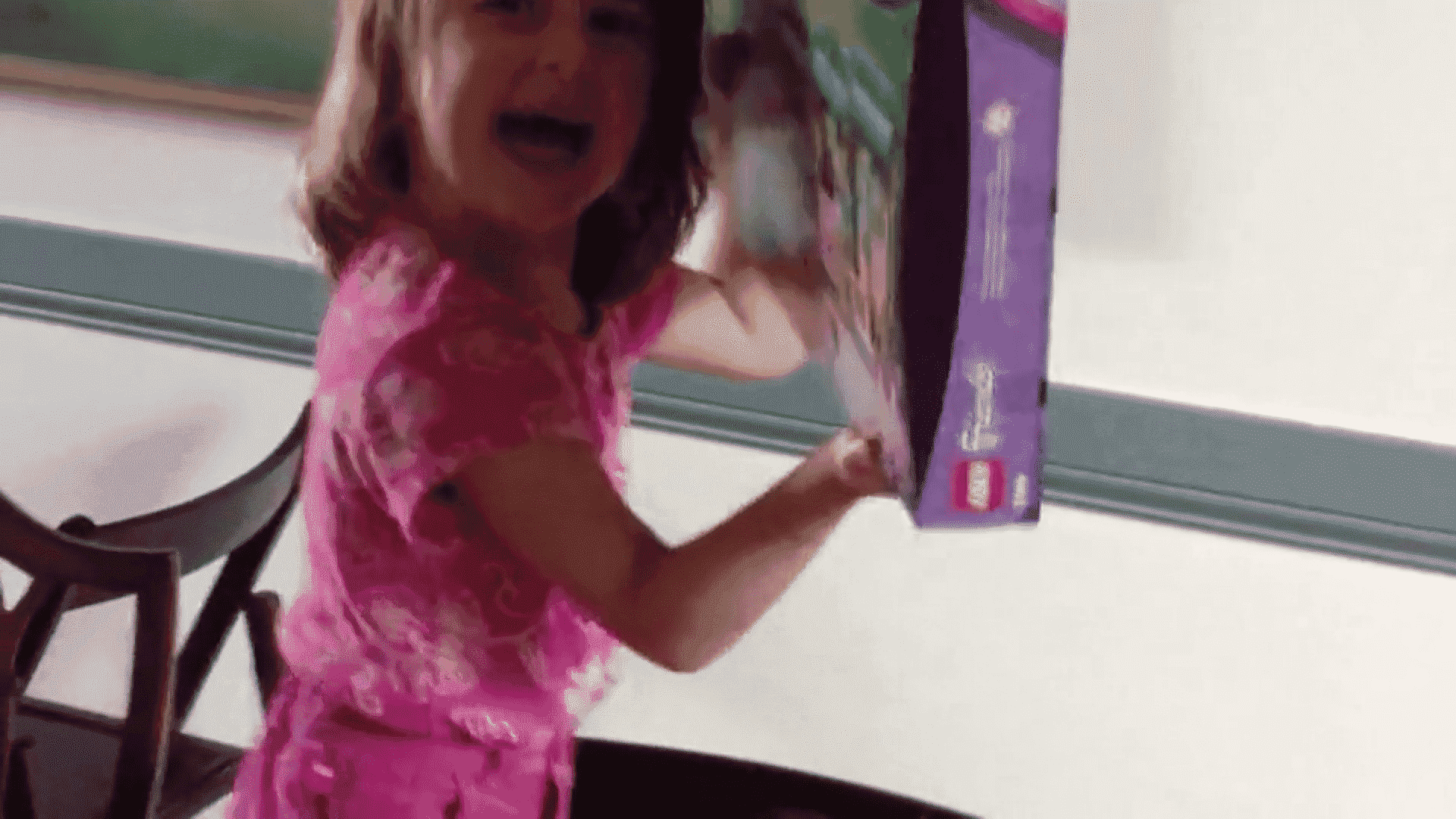} \hfill \includegraphics[width=.325\columnwidth]{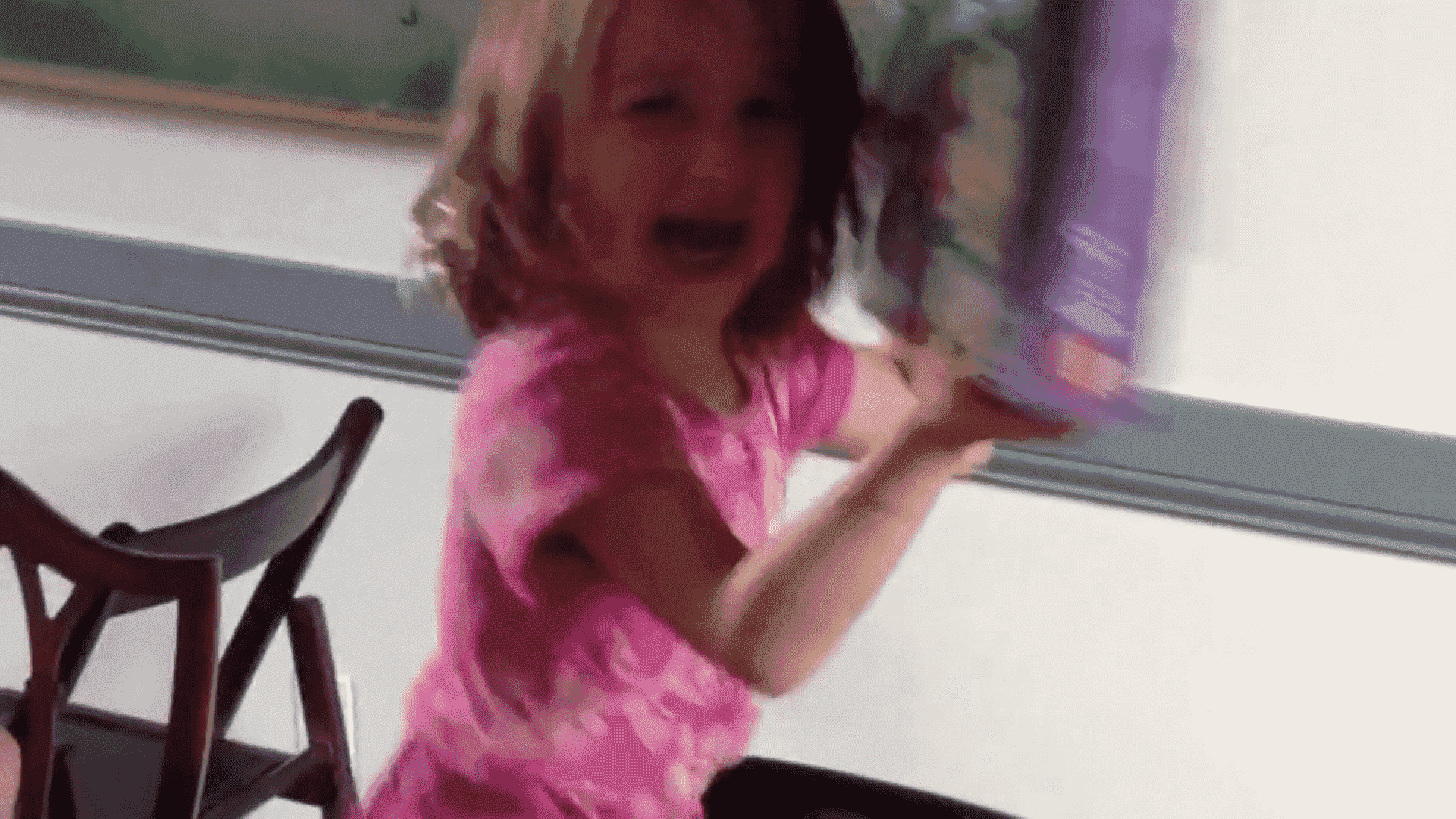}%
  \label{fig:KoNViD-1k-2}}
\hfill  
  \subfloat[Three representative frames of video C on KoNViD-1k]{\includegraphics[width=.325\columnwidth]{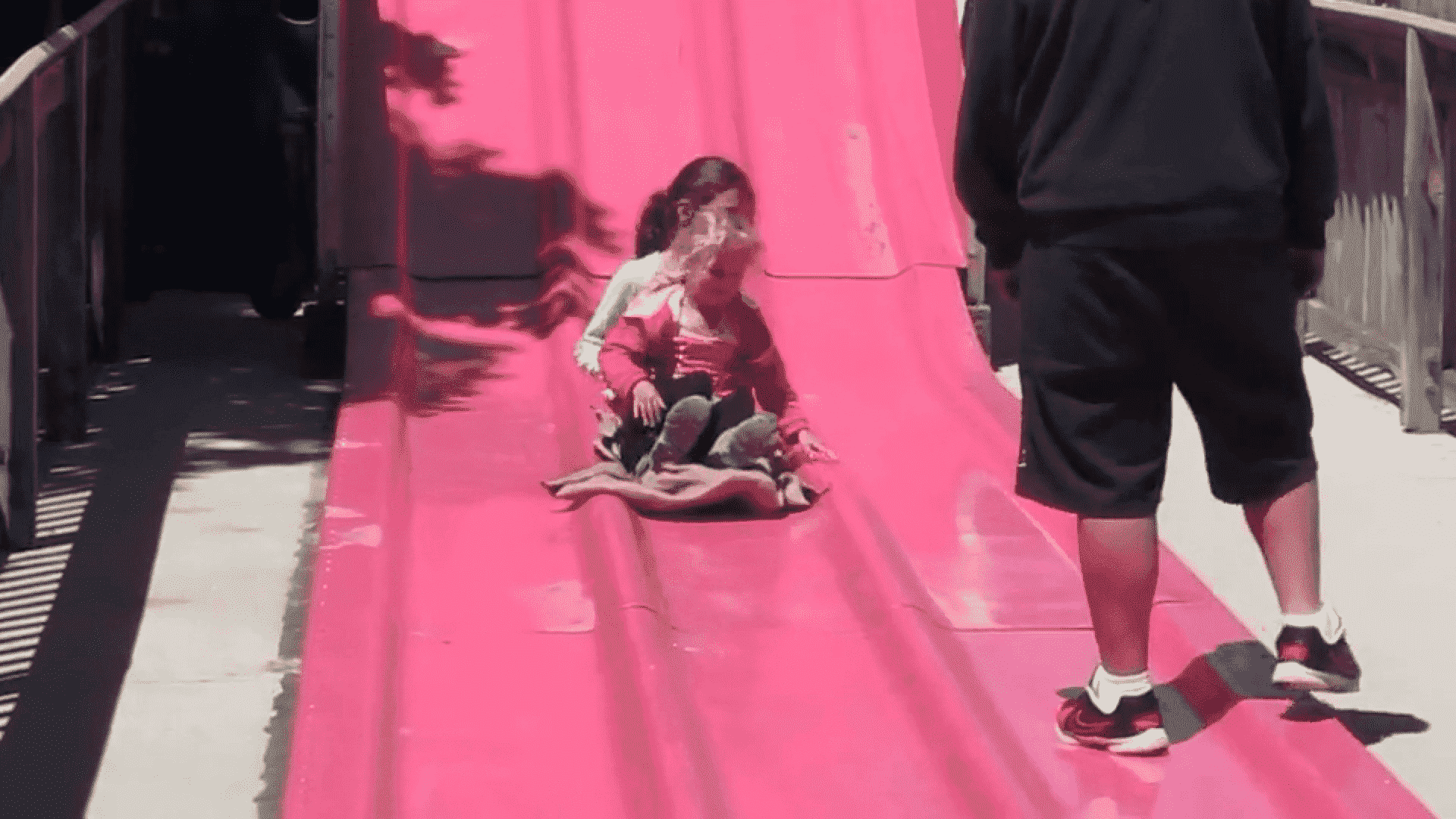} \hfill \includegraphics[width=.325\columnwidth]{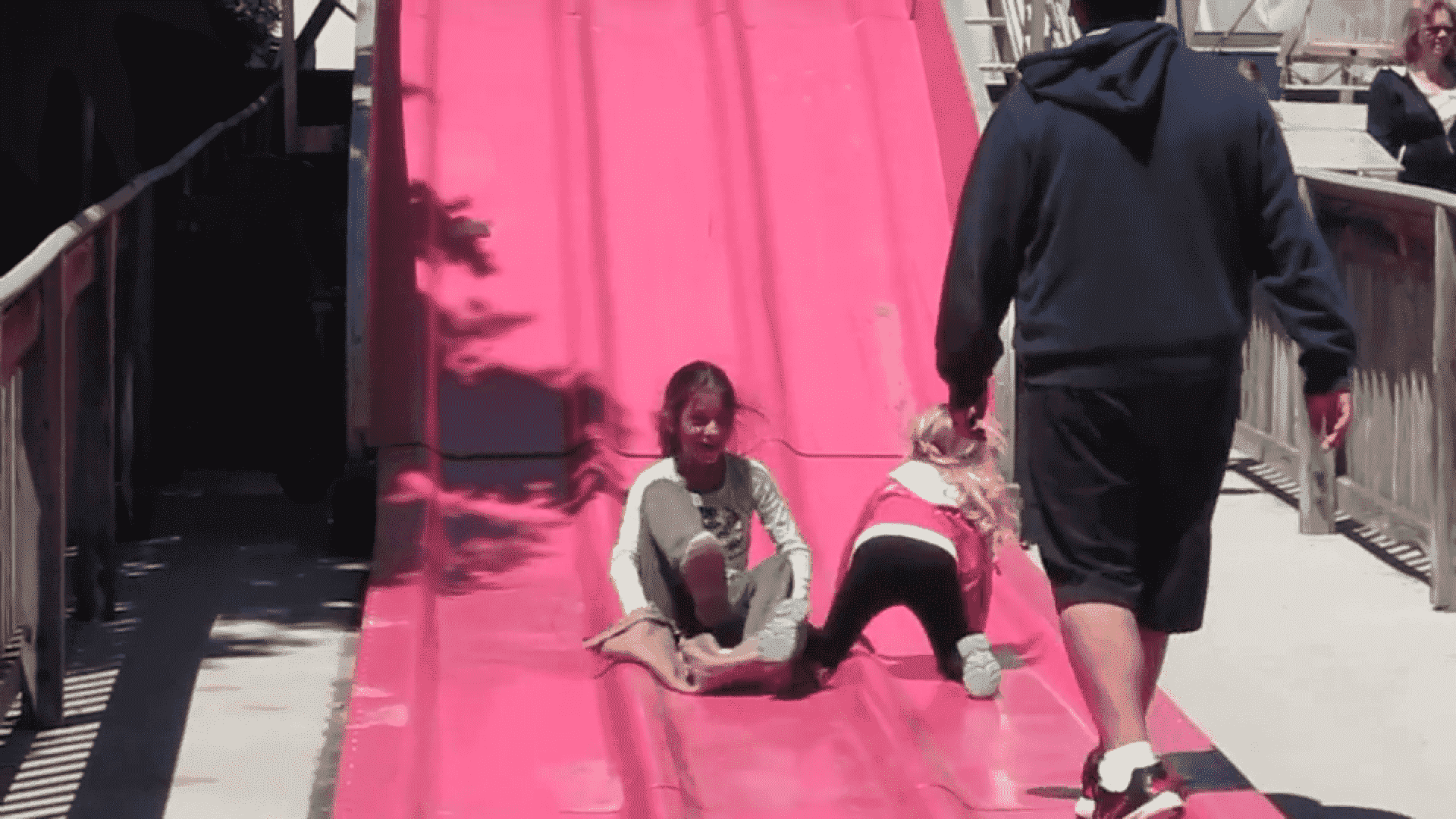} \hfill \includegraphics[width=.325\columnwidth]{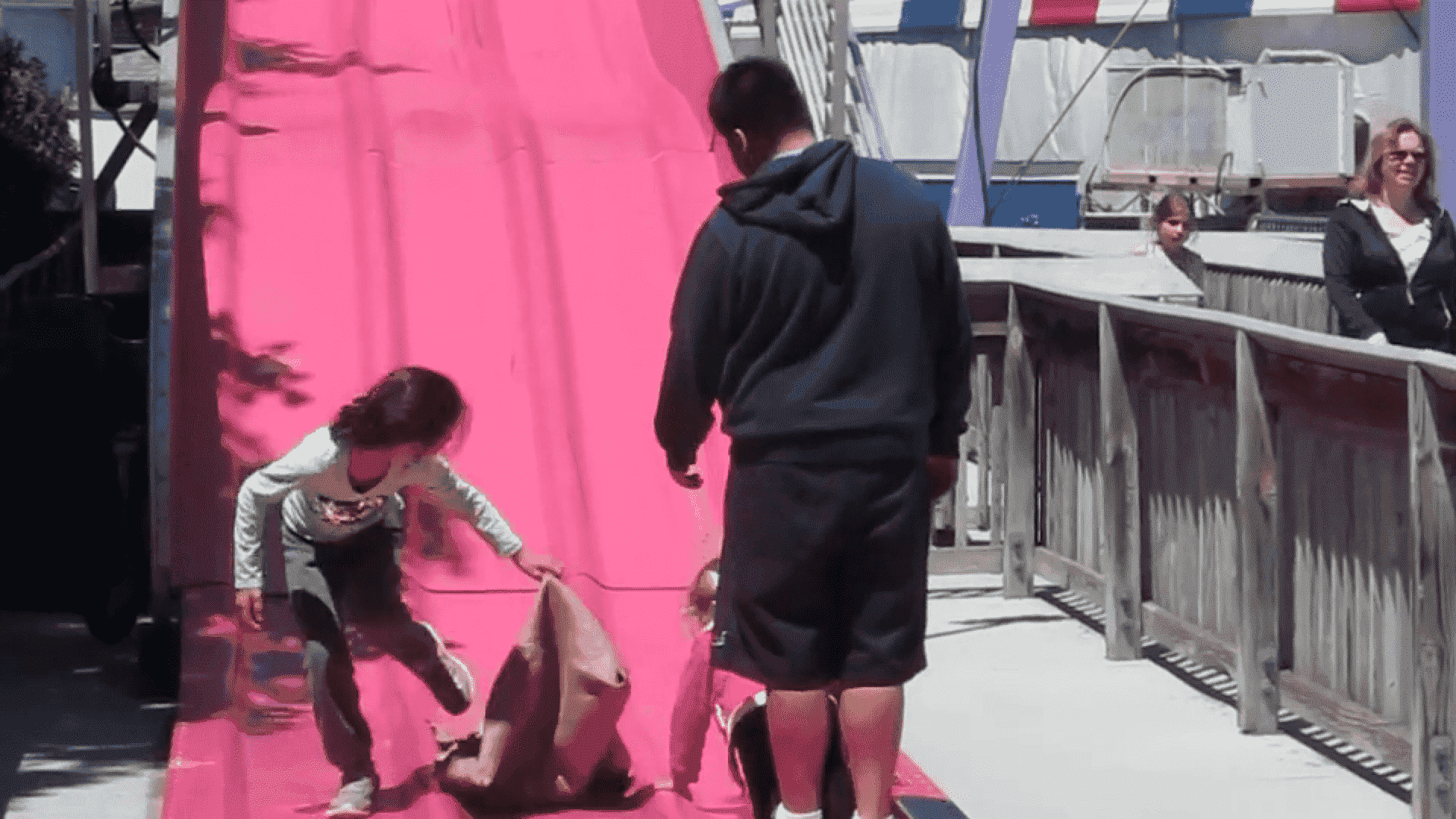}%
  \label{fig:KoNViD-1k-3}}
\hfill  
  \subfloat[Three representative frames of video D on KoNViD-1k]{\includegraphics[width=.325\columnwidth]{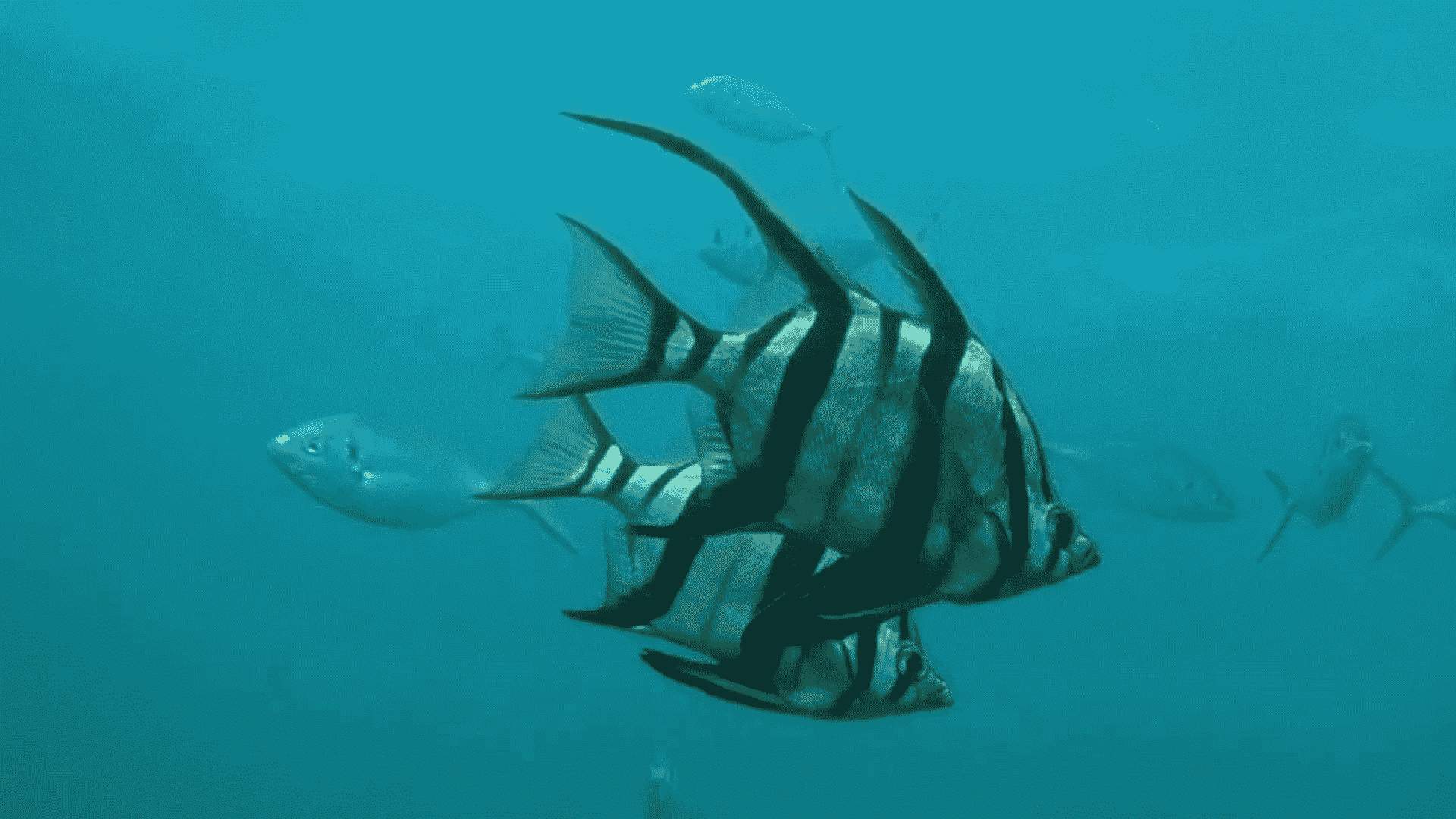} \hfill \includegraphics[width=.325\columnwidth]{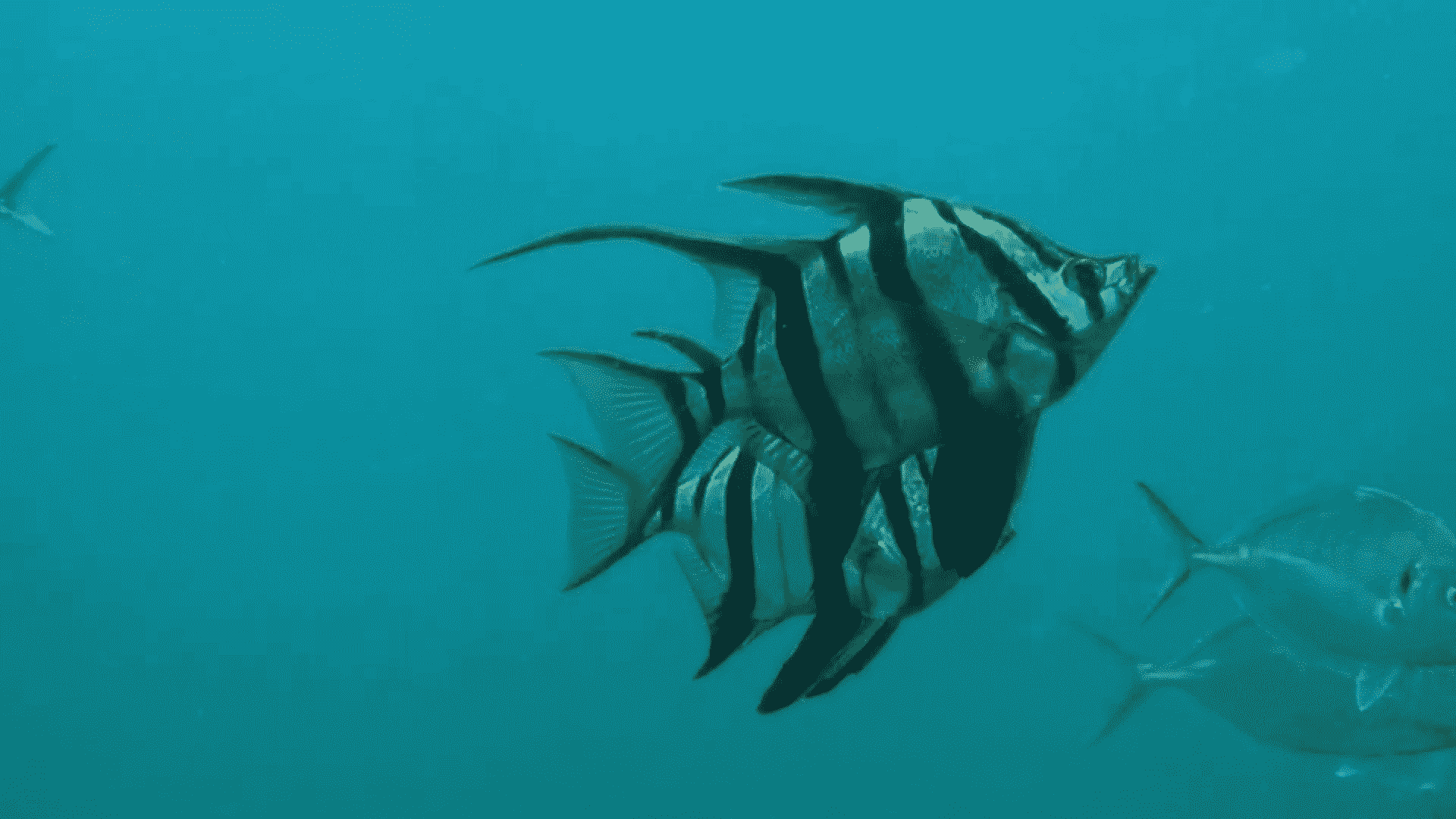} \hfill \includegraphics[width=.325\columnwidth]{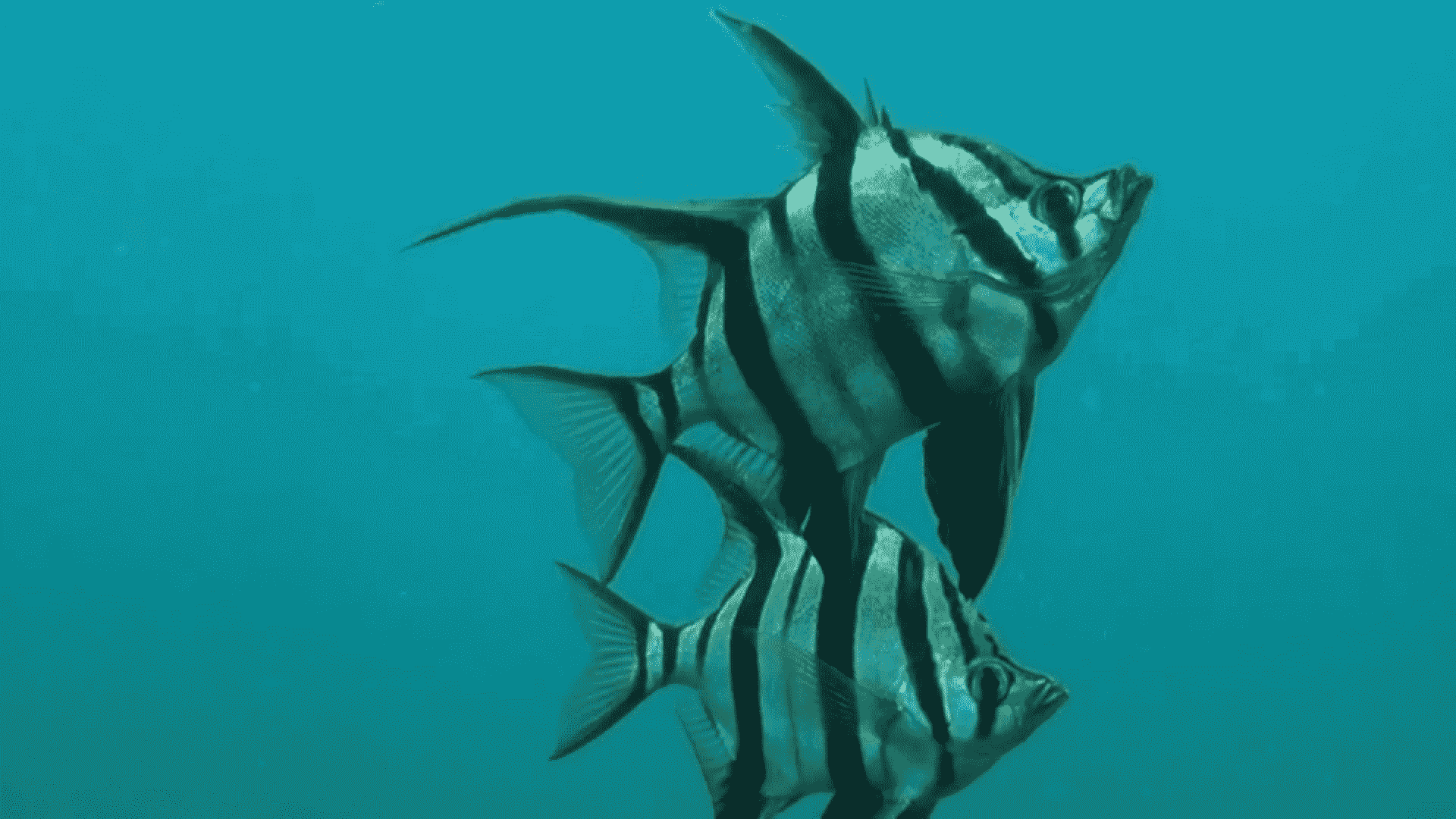}%
  \label{fig:KoNViD-1k-4}}
\hfill
\end{center}
  \caption{Qualitative example on KoNViD-1k test set. The quality rankings provided by MOS and MDTVSFA are both A$<$B$<$C$<$D, but LS-VSFA gives a quality ranking of A$<$C$<$B$<$D. Full-resolution videos are provided in \url{https://bit.ly/3csmHYk}.} 
 \label{fig:examples1}
\end{figure}

\begin{figure}[!htb]
\begin{center}
  \subfloat[Three representative frames of video E on LIVE-VQC]{\includegraphics[width=.325\linewidth]{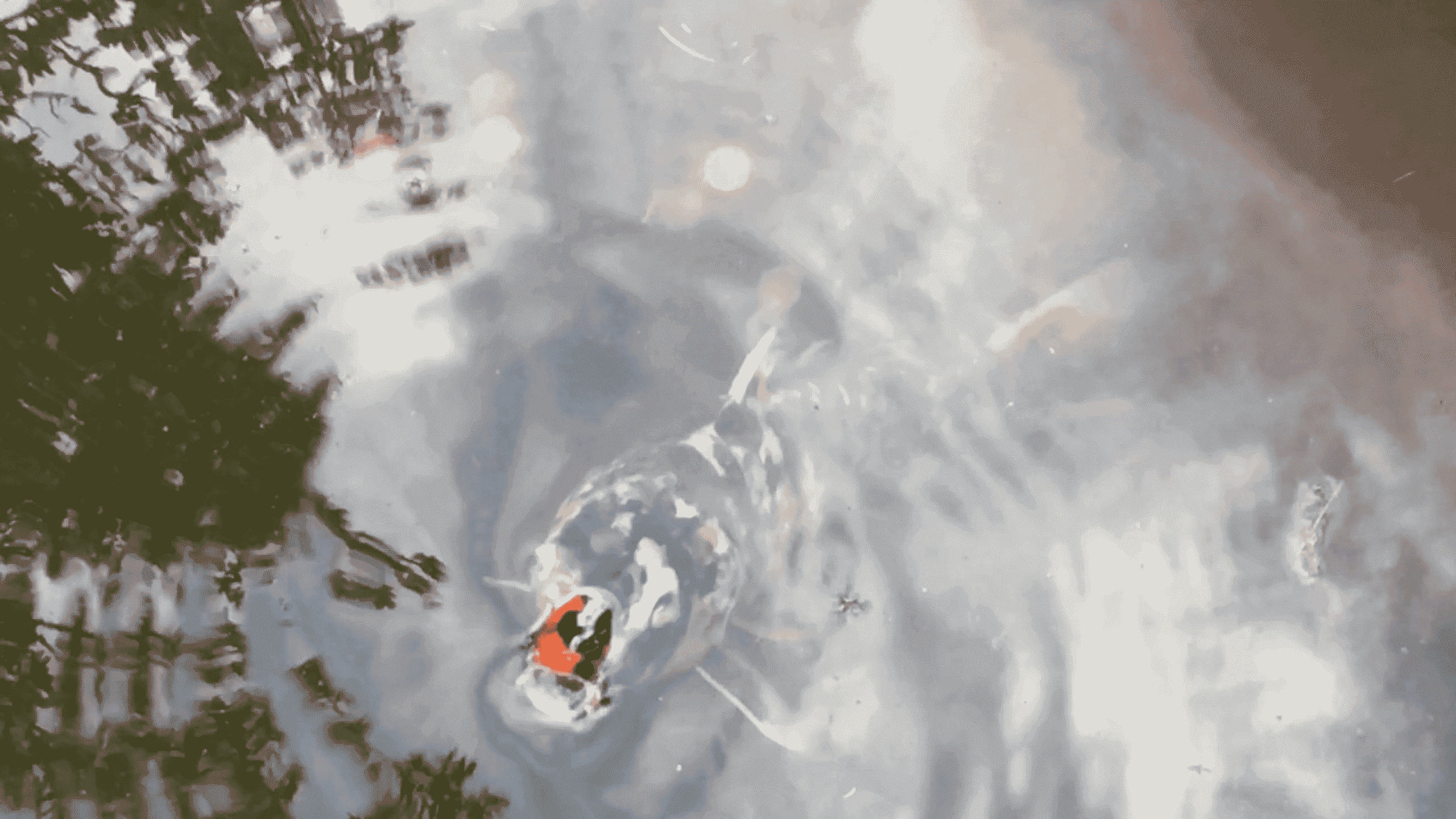} \hfill \includegraphics[width=.325\linewidth]{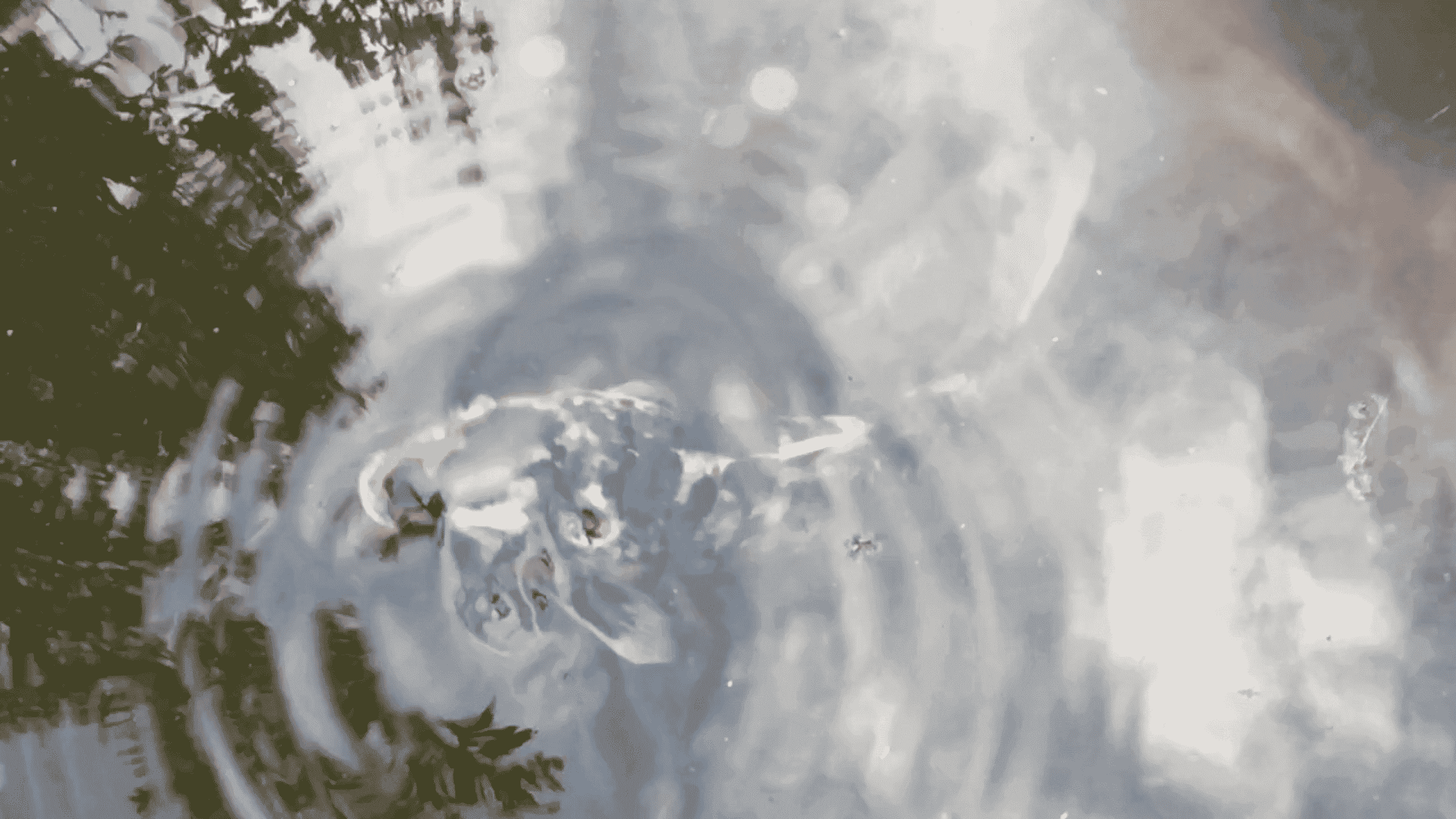} \hfill \includegraphics[width=.325\linewidth]{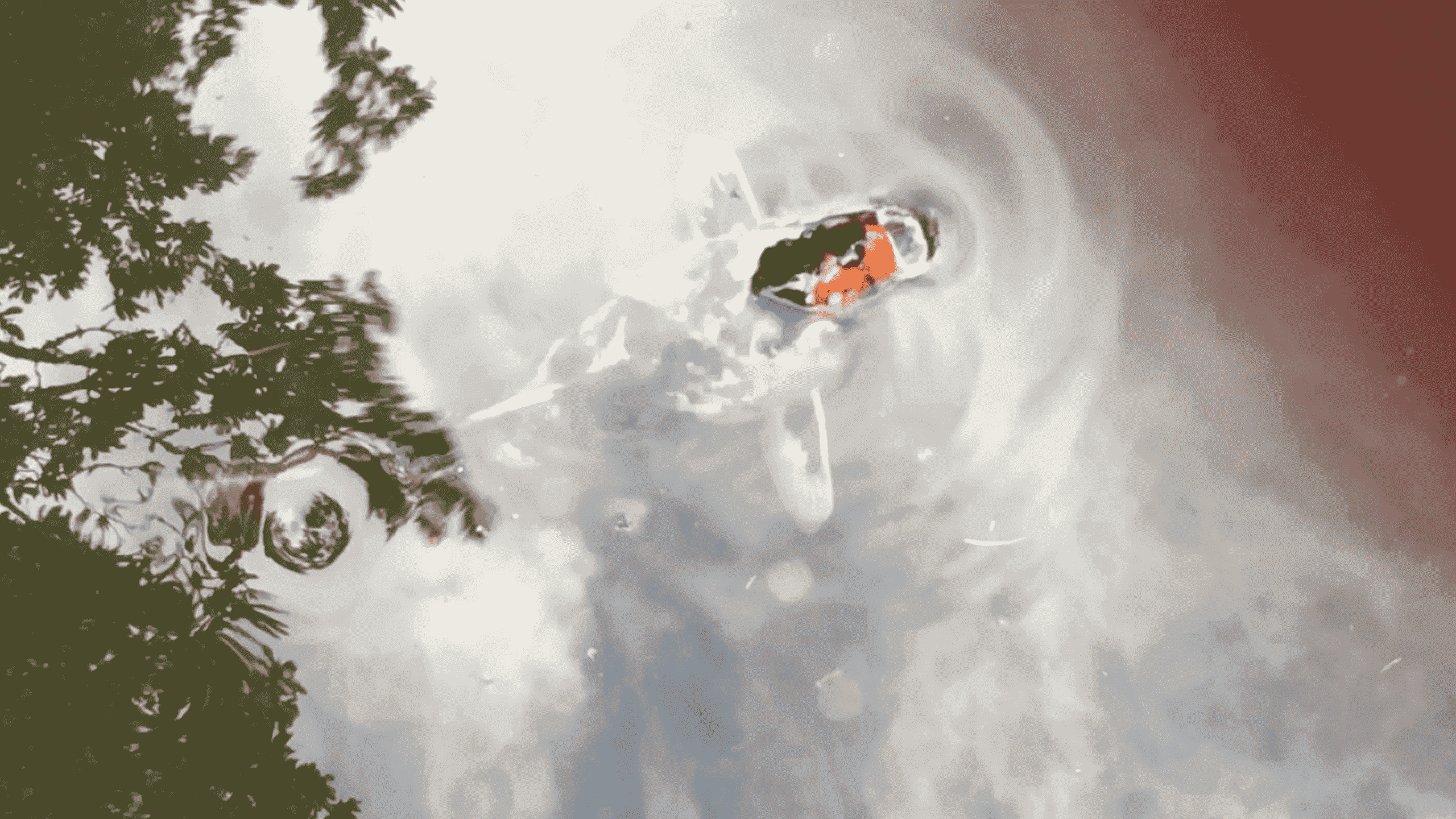}%
  \label{fig:LIVE-VQC-1}}
\hfill  
  \subfloat[Three representative frames of video F on LIVE-VQC]{\includegraphics[width=.325\linewidth]{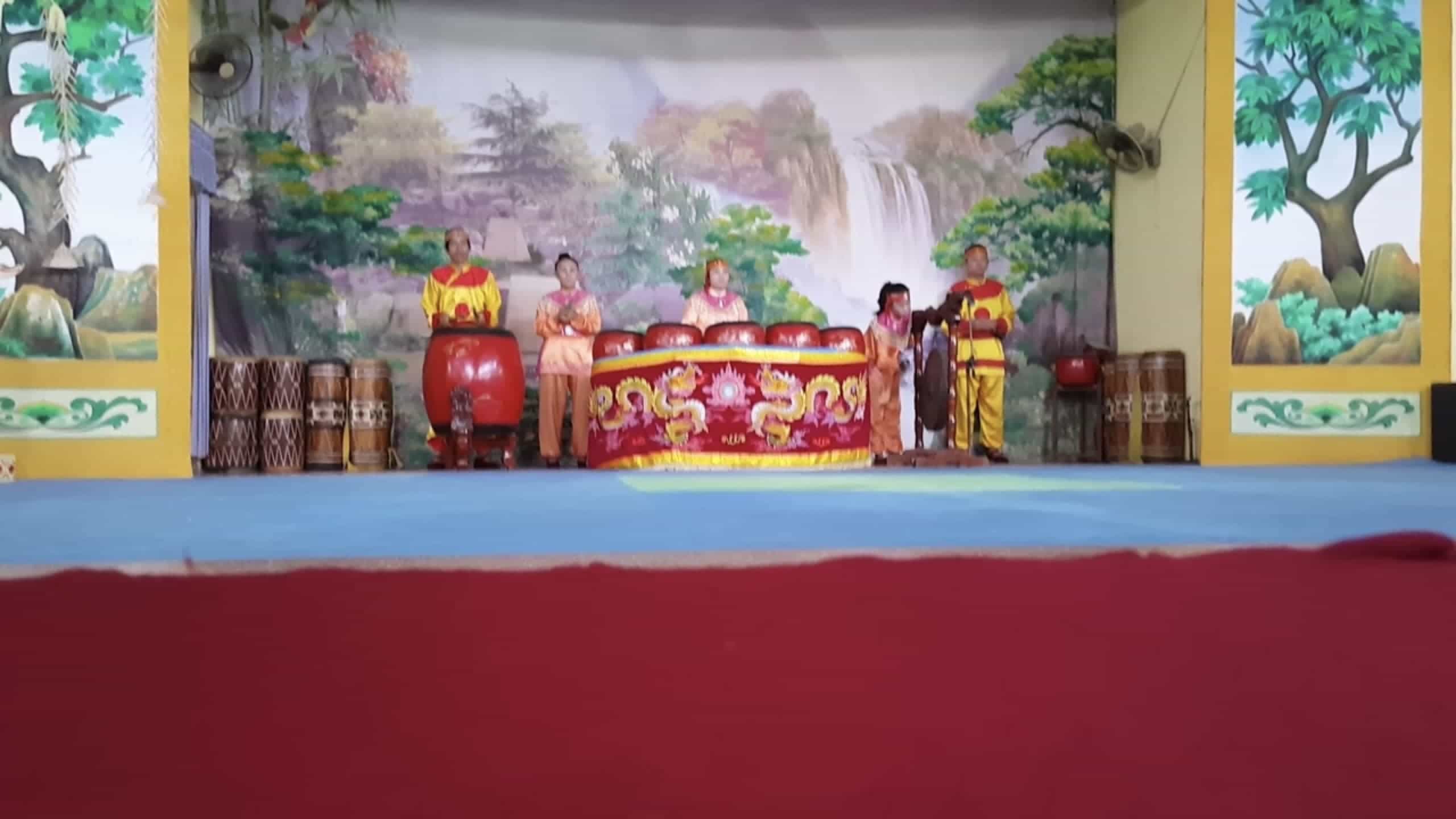} \hfill \includegraphics[width=.325\linewidth]{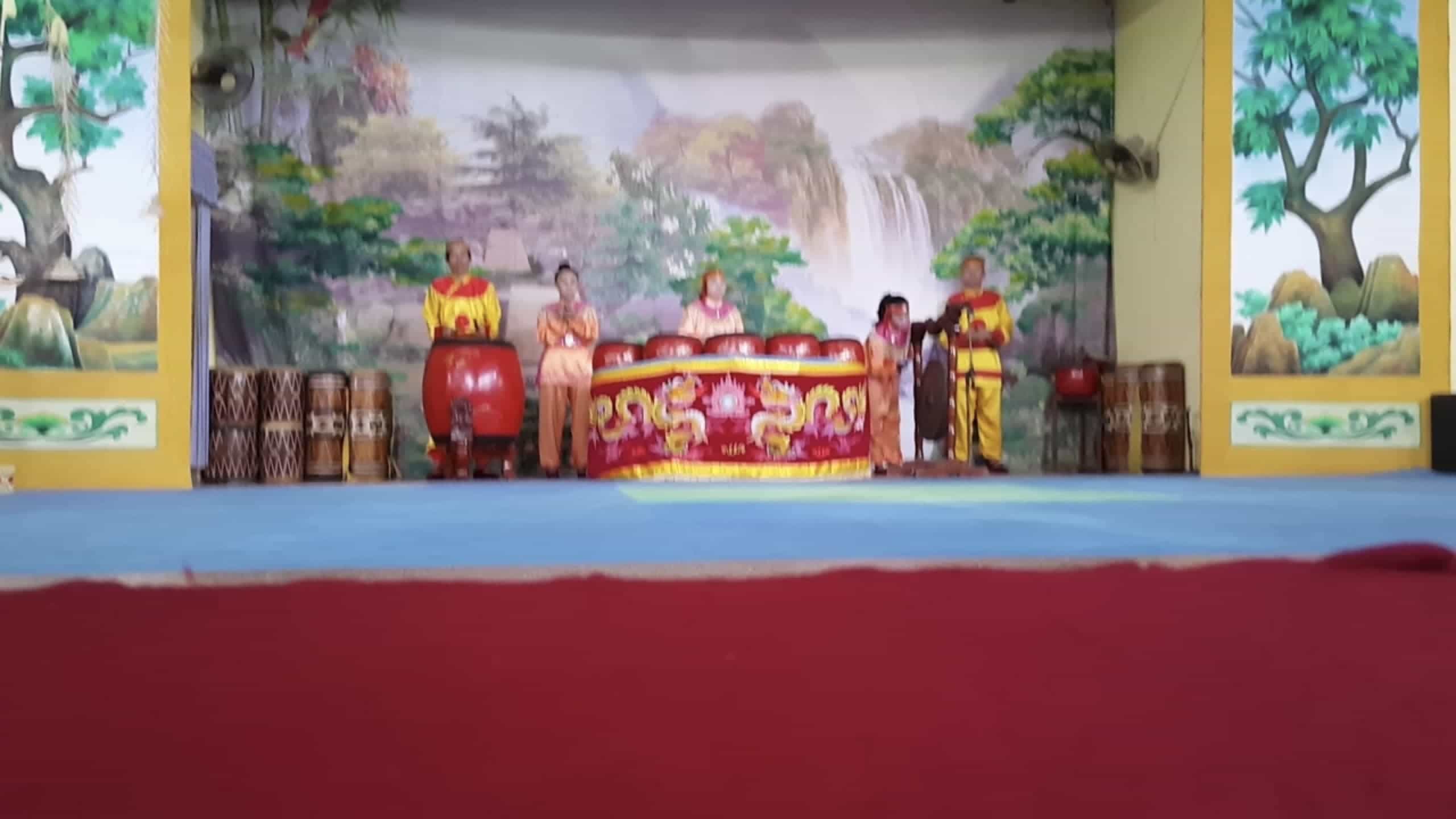} \hfill \includegraphics[width=.325\linewidth]{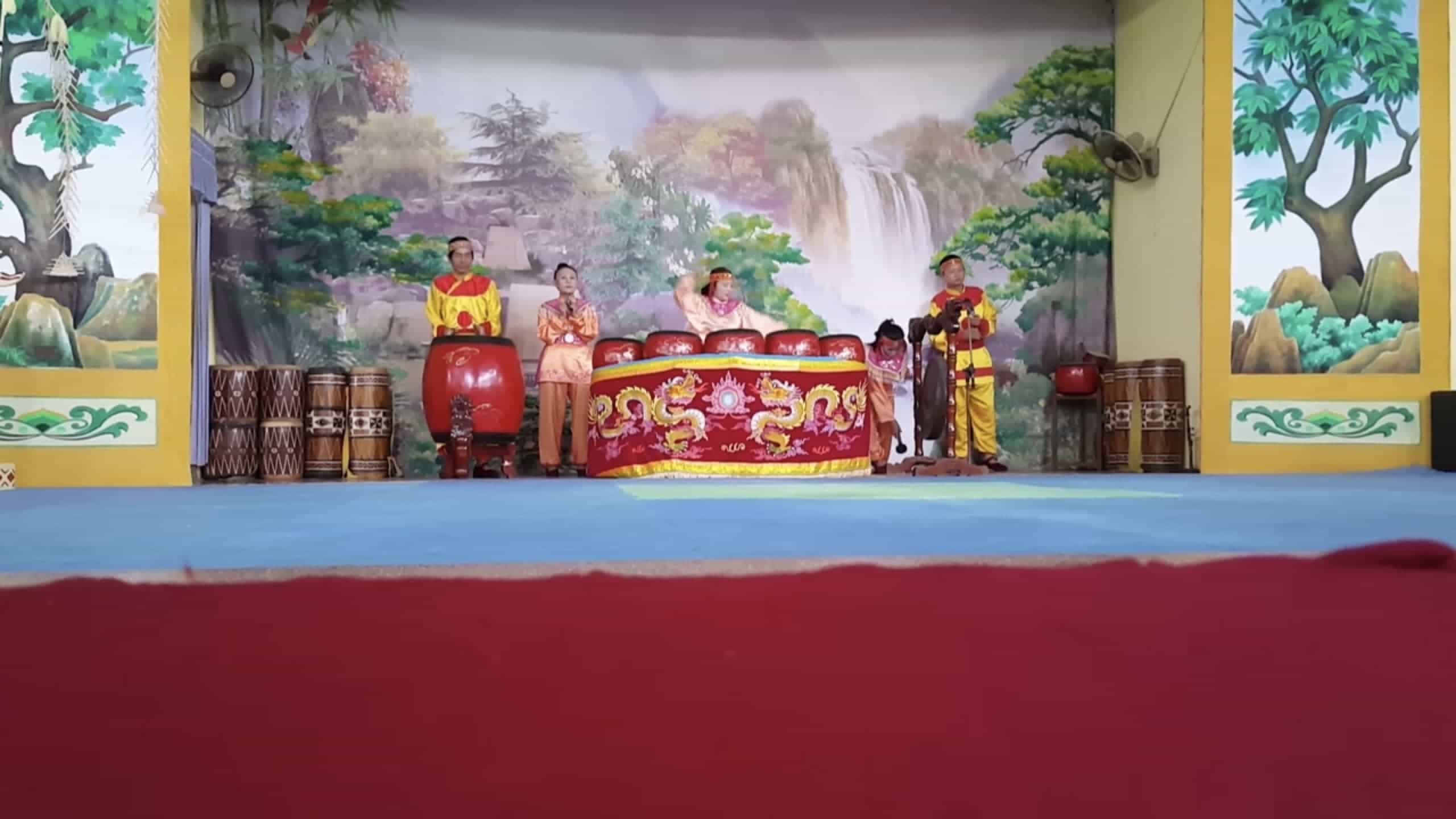}%
  \label{fig:LIVE-VQC-2}}
\hfill
\end{center}
  \caption{Qualitative example on LIVE-VQC. The quality rankings provided by MOS and MDTVSFA are both E$>$F, but LS-VSFA gives a quality ranking of E$<$F. Full-resolution videos are provided in \url{https://bit.ly/3csmHYk}.} 
 \label{fig:examples2}
\end{figure}

\begin{figure}[!htb]
\begin{center} 
  \subfloat[Three representative frames of video G on LIVE-VQC]{\includegraphics[width=.325\linewidth]{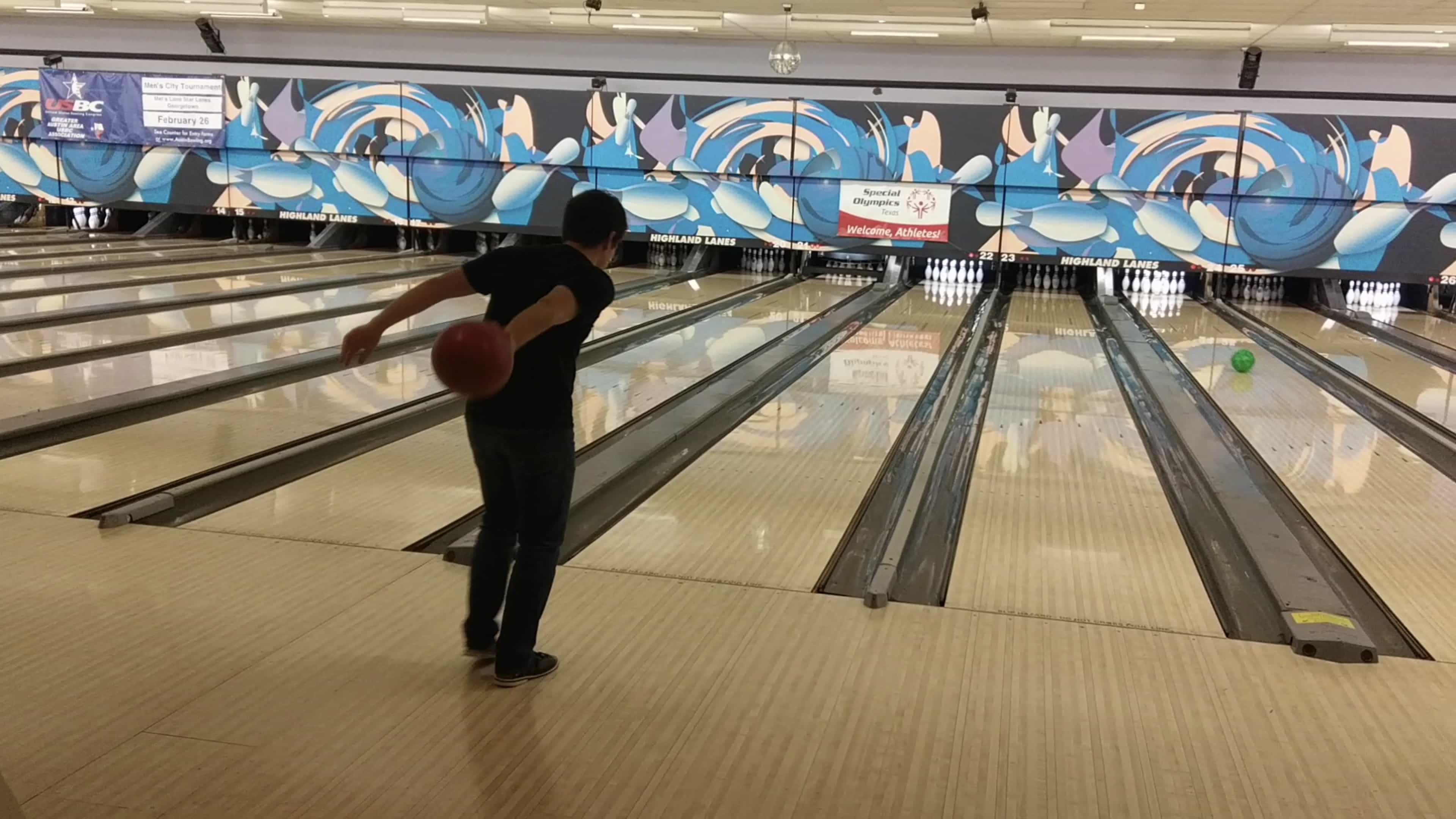} \hfill \includegraphics[width=.325\linewidth]{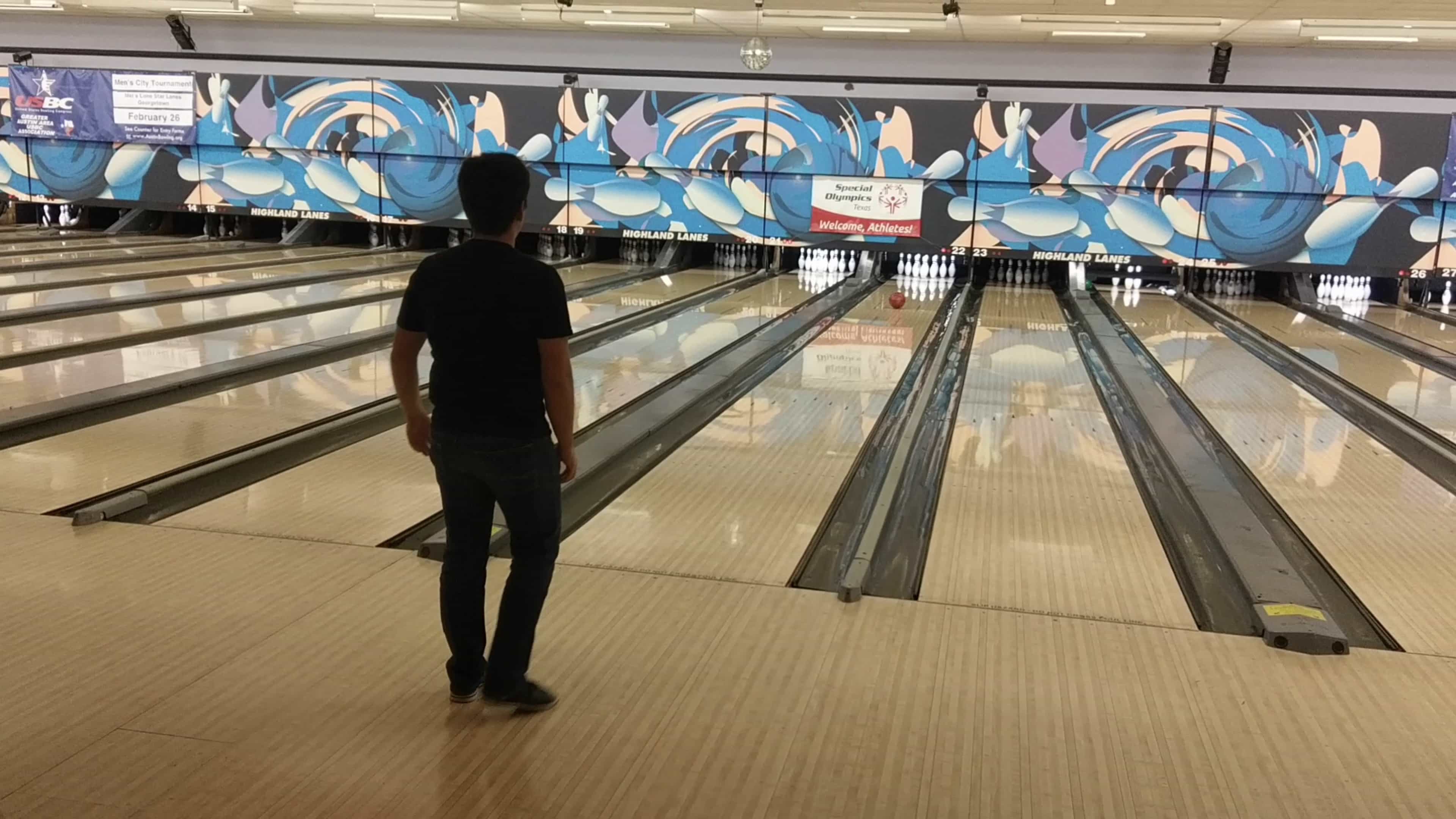} \hfill \includegraphics[width=.325\linewidth]{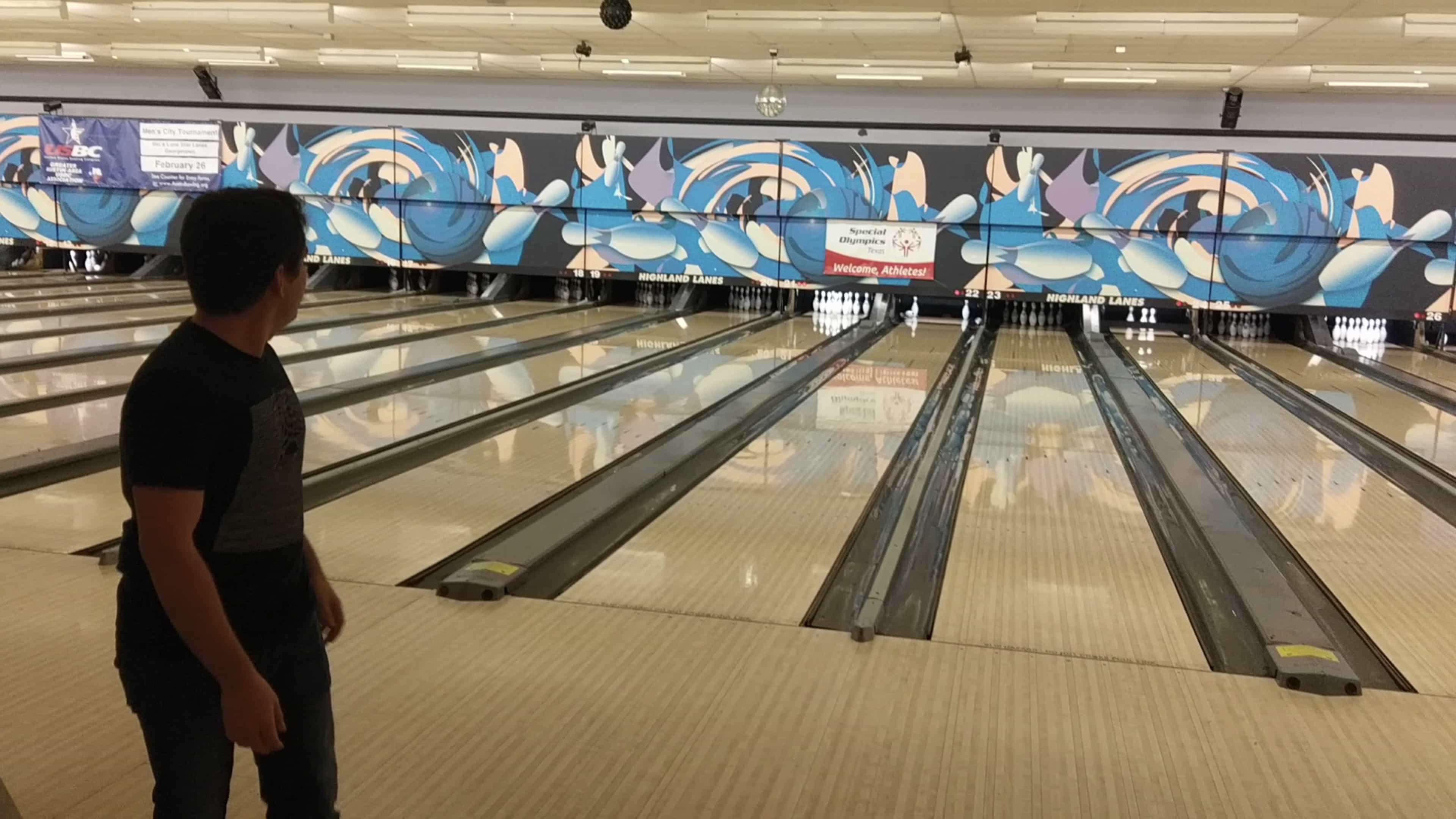}%
  \label{fig:LIVE-VQC-3}}
\hfill  
  \subfloat[Three representative frames of video H on LIVE-VQC]{\includegraphics[width=.325\linewidth]{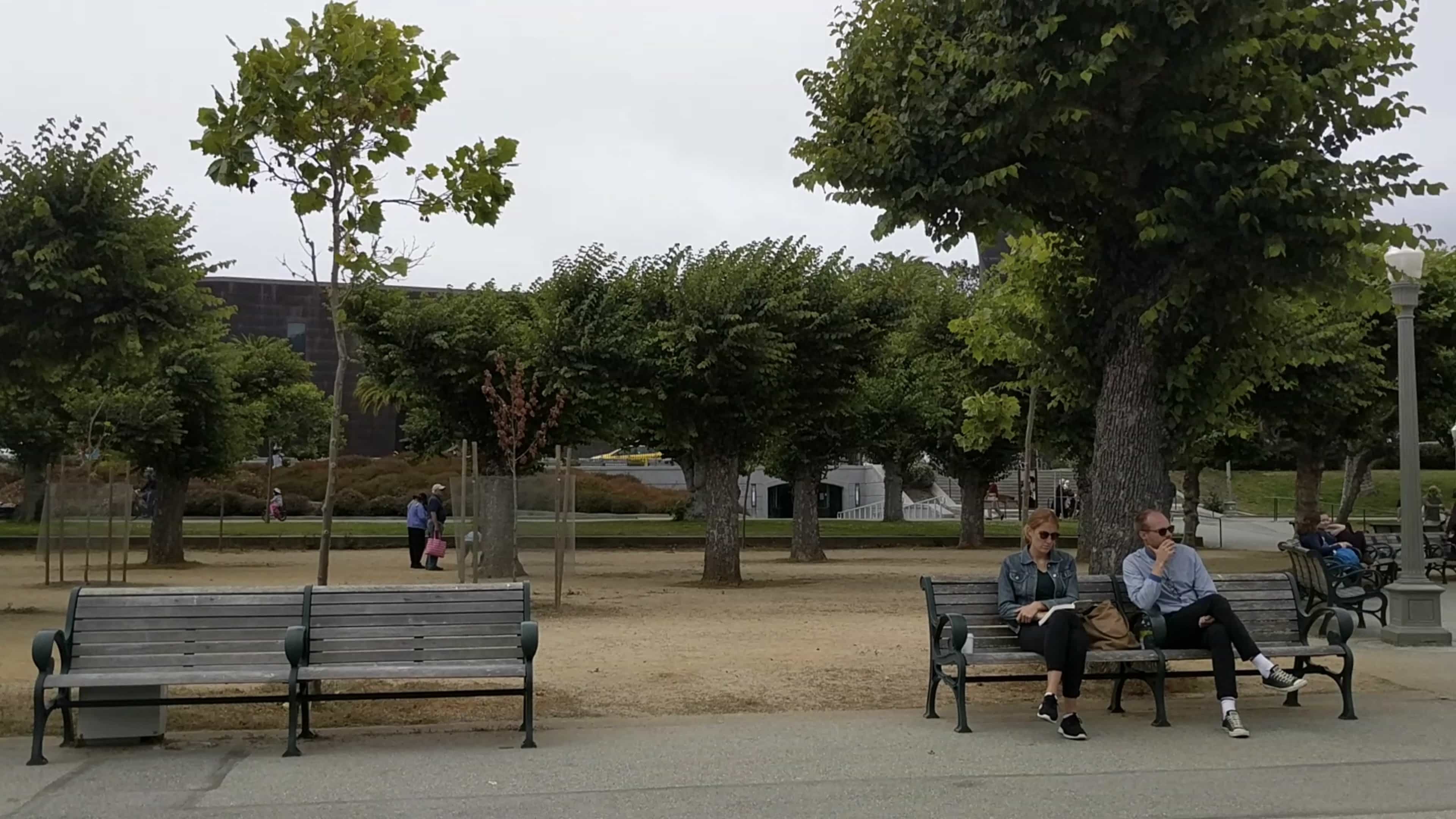} \hfill \includegraphics[width=.325\linewidth]{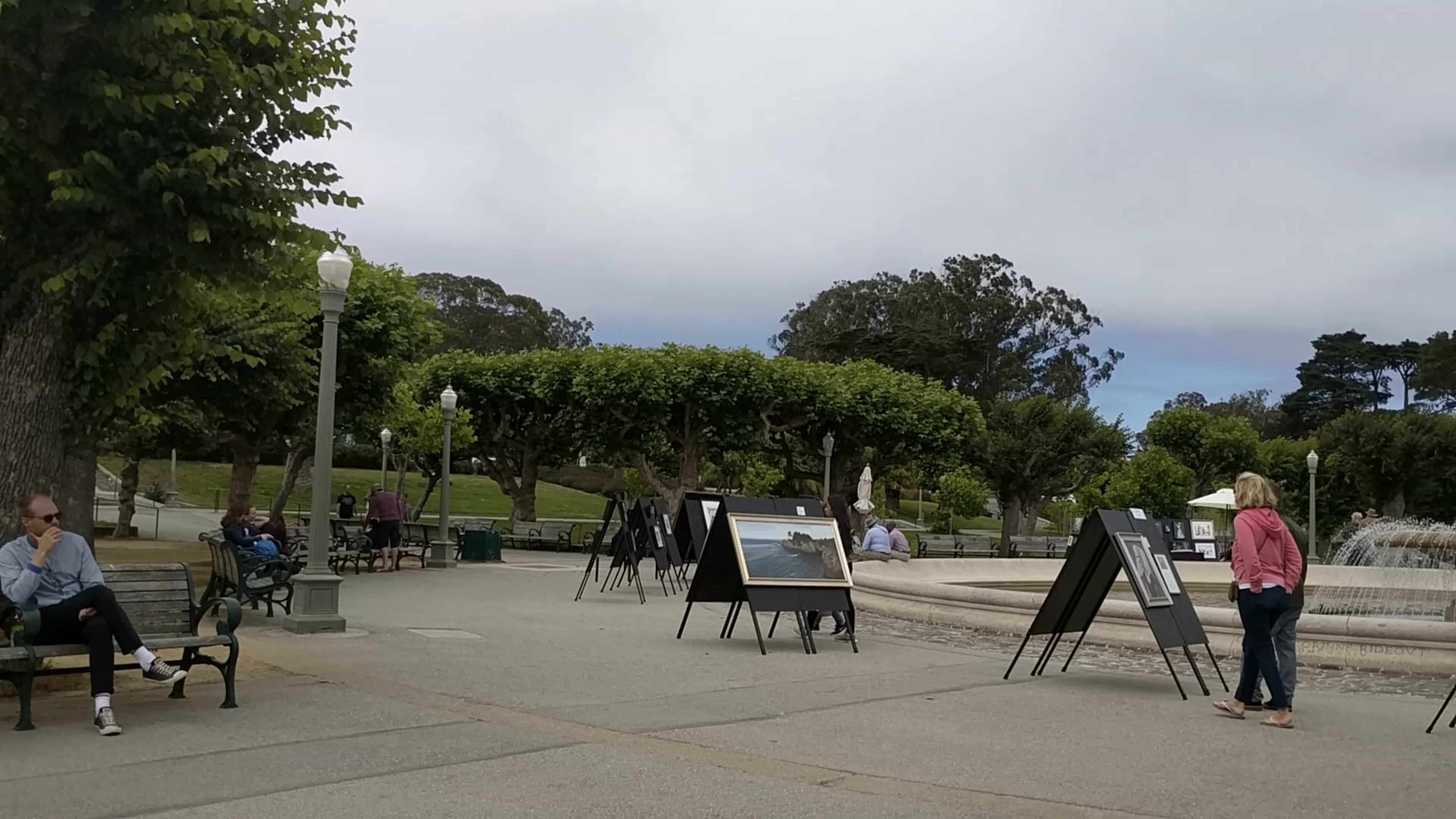} \hfill \includegraphics[width=.325\linewidth]{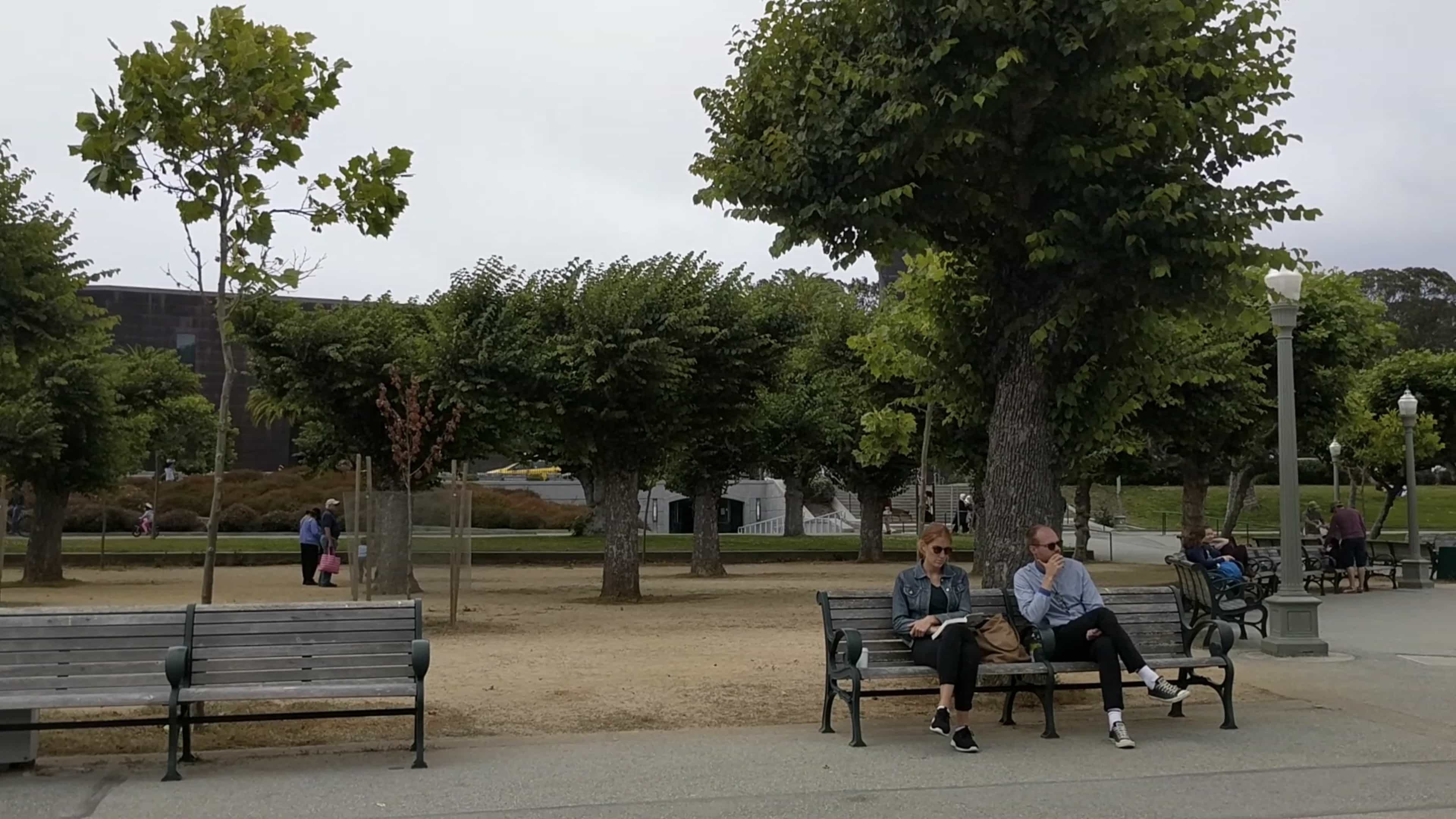}%
  \label{fig:LIVE-VQC-4}}
\hfill
\end{center}
  \caption{Another qualitative example on LIVE-VQC. The quality rankings provided by LS-VSFA and MDTVSFA are both G$<$H, but MOS gives a quality ranking of G$>$H. Note that the scenes change fast in video H, where full-resolution videos are provided in \url{https://bit.ly/3csmHYk}.} 
 \label{fig:examples3}
\end{figure}

In Fig.~\ref{fig:examples1}, \ref{fig:examples2} and~\ref{fig:examples3}, we show several success and failure cases of our method.
Fig.~\ref{fig:examples1} and Fig.~\ref{fig:examples2} show the success cases of MDTVSFA, which means the predictions of MDTVSFA model is consistent with MOS.
LS-VSFA has more failure cases than MDTVSFA since the linear re-scaling strategy disturbs the training process.
We also show two failure cases of LS-VSFA in Fig.~\ref{fig:examples1} and Fig.~\ref{fig:examples2}.
Besides, there is still a large room for improving the performance of MDTVSFA, and we show a failure case of both MDTVSFA and LS-VSFA in Fig.~\ref{fig:examples3}.
Such failure may be due to the fact that our models extract frame-level features and not fully exploit the motion and spatial-temporal information.
For example, our methods do not account for the discomfort caused by suddenly and fast scene change.

\textbf{Performance on individual datasets}. 
Besides the overall performance reported in last part, we report performance on individual datasets in this part.
Our method is trained by mixing the four datasets while other methods are trained on individual datasets. 
Table~\ref{tab:performance} summarizes the performance values on the four datasets individually. 
The results provided by our method are based only on a single unified model while the results provided by other methods are based on different models trained for different datasets.
The natural scene statistics (NSS)-based NR-IQA methods (such as BRISQUE) outperform VIIDEO.
This may be owing to the fact that VIIDEO is based only on temporal scene statistics and cannot model the complex distortions. 
VBLIINDS and TLVQM rely on a lot of carefully-designed handcrafted features that capture the spatial and temporal distortions, and thus they achieve a better performance than the NR-IQA methods and VIIDEO. 
Our method achieves the best performance in terms of prediction monotonicity (SROCC) and prediction accuracy (PLCC) on the three datasets (LIVE-VQC, LIVE-Qualcomm, and KoNViD-1k).
On CVD2014, MDTVSFA slightly outperforms TLVQM in terms of SROCC, while it slightly underperforms TLVQM in terms PLCC. 
However, we should note that the results of our method is based only on one single model, which indicates our unified model performs well across datasets.

\newcommand\Tstrut{\rule{0pt}{2.6ex}}         
\newcommand\Bstrut{\rule[-0.9ex]{0pt}{0pt}}   
\begin{table*}[!hbt]
    \centering
    \caption{Performance comparison on the four VQA datasets individually. Mean and standard deviation (std) of the performance values in 10 runs are reported, \textit{i.e.}, mean ($\pm$ std). In each column, the best mean SROCC and PLCC values are marked in boldface, and the second-best performance values are underlined.}
    \label{tab:performance}

    \begin{small}
    
    \resizebox{\textwidth}{!}{
    \begin{tabular}{l|cc|cc}
    \toprule
    \multirow{2}{*}{Method} & \multicolumn{2}{c|}{LIVE-VQC~\citep{sinno2019large}} & \multicolumn{2}{c}{LIVE-Qualcomm~\citep{ghadiyaram2018capture}} \\
    & SROCC$\uparrow$ & PLCC$\uparrow$ & SROCC$\uparrow$ & PLCC$\uparrow$ \Bstrut\\
    \hline
    BRISQUE~\citep{mittal2012no} & 0.5687 ($\pm$ 0.0729) & 0.5868 ($\pm$ 0.0642)  & 0.5036 ($\pm$ 0.1470) & 0.5158 ($\pm$ 0.1274) \Tstrut\\
    NIQE~\citep{mittal2013making} & 0.5892 ($\pm$ 0.0538)  & 0.6112 ($\pm$ 0.0554) & 0.4628 ($\pm$ 0.1052) & 0.4638 ($\pm$ 0.1362) \\
    CORNIA~\citep{ye2012unsupervised} &  0.5953 ($\pm$ 0.0170) & 0.5926 ($\pm$ 0.0230) & 0.4598 ($\pm$ 0.1299) & 0.4941 ($\pm$ 0.1327) \\
    VIIDEO~\citep{mittal2016completely} &  0.1498 ($\pm$ 0.0995) & 0.2454 ($\pm$ 0.0740) & 0.1267 ($\pm$ 0.1368) & -0.0012 ($\pm$ 0.1062)\\
    VBLIINDS~\citep{saad2014blind} &  \underline{0.7015} ($\pm$ 0.0483) & \underline{0.7120} ($\pm$ 0.0501) & 0.5659 ($\pm$ 0.0780) & 0.5676 ($\pm$ 0.0885)  \\
    ST-Naturalness~\citep{sinno2019spatio} & 0.5994$^*$ & 0.6069$^*$ & - & -\\
    3D-CNN+LSTM~\citep{you2019deep} & - & - & 0.687$^*$ & 0.792$^*$ \\
    FRIQUEE~\citep{ghadiyaram2017perceptual} & - & - & 0.6795$^*$ & 0.7349$^*$ \\
    TLVQM~\citep{korhonen2019two} & - & -  & \underline{0.78} ($\pm$ 0.07)$^*$ & \underline{0.81} ($\pm$ 0.06)$^*$  \Bstrut\\
    \hline
    \textbf{MDTVSFA} & \textbf{0.7382} ($\pm$ 0.0357) & \textbf{0.7728} ($\pm$ 0.0351) & \textbf{0.8019} ($\pm$ 0.0295) & \textbf{0.8218} ($\pm$ 0.0374) \Tstrut\\
    \bottomrule
    \end{tabular}
    }

    \bigskip
    
    \begin{tabular}{l|cc|cc}
    \toprule
    \multirow{2}{*}{Method} & \multicolumn{2}{c|}{KoNViD-1k~\citep{hosu2017konstanz}} & \multicolumn{2}{c}{CVD2014~\citep{nuutinen2016cvd2014}} \\
    & SROCC$\uparrow$ & PLCC$\uparrow$ & SROCC$\uparrow$ & PLCC$\uparrow$ \Bstrut\\
    \hline
    BRISQUE~\citep{mittal2012no}  & 0.6540 ($\pm$ 0.0418) & 0.6256 ($\pm$ 0.0407)  & 0.7086 ($\pm$ 0.0666) & 0.7154 ($\pm$ 0.0476) \Tstrut\\
    NIQE~\citep{mittal2013making}  & 0.5435 ($\pm$ 0.0396) & 0.5456 ($\pm$ 0.0376) & 0.4890 ($\pm$ 0.0908)  & 0.5931 ($\pm$ 0.0650) \\
    CORNIA~\citep{ye2012unsupervised}  & 0.6096 ($\pm$ 0.0343) & 0.6075 ($\pm$ 0.0318) & 0.6140 ($\pm$ 0.0754) & 0.6178 ($\pm$ 0.0792) \\
    VIIDEO~\citep{mittal2016completely}  & 0.2976 ($\pm$ 0.0522) & 0.3026 ($\pm$ 0.0486)  & 0.0228 ($\pm$ 0.1216) & -0.0249 ($\pm$ 0.1439) \\
    VBLIINDS~\citep{saad2014blind} & 0.6947 ($\pm$ 0.0239) & 0.6576 ($\pm$ 0.0254)  & 0.7458 ($\pm$ 0.0564) & 0.7525 ($\pm$ 0.0528) \\
    FC Model~\citep{men2017empirical} & 0.572$^*$ & 0.565$^*$ & - & -\\
    STFC Model~\citep{men2018spatiotemporal} & 0.606$^*$ & 0.639$^*$ & - & -\\
    STS-CNN200~\citep{yan2019no} & 0.735$^*$ & - & - & -\\
    TLVQM~\citep{korhonen2019two} & \textbf{0.78} ($\pm$ 0.02)$^*$ & \underline{0.77} ($\pm$ 0.02)$^*$  & \textbf{0.83} ($\pm$ 0.04)$^*$ & \textbf{0.85} ($\pm$ 0.04)$^*$  \Bstrut\\
    \hline
    \textbf{MDTVSFA} & \textbf{0.7812} ($\pm$ 0.0278) & \textbf{0.7856} ($\pm$ 0.0240) & \textbf{0.8314} ($\pm$ 0.0416) & \underline{0.8407} ($\pm$ 0.0296) \Tstrut\\
    \bottomrule
    \end{tabular}
    
    \vspace{1mm}
    
    $^*$The reported results in their original papers are shown here for reference.
    \end{small}
\end{table*}

\begin{table*}[!thb]
    \centering
    \caption{Performance comparison in terms of median SROCC between the single models trained by mixing all three datasets (CVD2014, KoNViD-1k, and LIVE-Qualcomm) and the models trained on one of the datasets. Overall performance indicates the dataset-size weighted median SROCC values in 10 runs. For each column, the largest value is marked in boldface.}
    \label{tab:vs individual}

    \begin{small}
    
    \begin{tabular}{llccccc}
    \toprule
    \multirow{2}{*}{Model} & \multirow{2}{*}{Train data} & \multirow{2}{*}{Mixed datasets training} & \multicolumn{3}{c}{Test dataset} & Overall \\
    & & & CVD2014 & KoNViD-1k & LIVE-Qualcomm & Performance\\
    \midrule
    \multirow{4}{*}{BRISQUE} & CVD2014 & No & 0.7582 & 0.5574 & 0.4632 & 0.5794   \\
    & KoNViD-1k & No & 0.5388 & 0.6191 & 0.3019 & 0.5621   \\
    & LIVE-Qualcomm & No & 0.3930 & 0.2341 & 0.5023 & 0.2973   \\ 
    & All three datasets & Linear re-scaling & 0.7356 & 0.6300 & 0.3809 & 0.6107  \\ 
    \midrule
    \multirow{4}{*}{VBLIINDS} & CVD2014 & No &  0.7892 & 0.5787 & 0.4170 & 0.5864 \\
    & KoNViD-1k & No &  0.5681 & 0.7078 & 0.4583 & 0.6544   \\
    & LIVE-Qualcomm & No &  0.5027 & 0.5432 & 0.6018 & 0.5544   \\ 
    & All three datasets & Linear re-scaling &  0.6749 & 0.6890 & 0.4684 & 0.6640   \\ 
    \midrule
    \multirow{4}{*}{TLVQM} & CVD2014 & No &  0.83$^*$ & 0.54$^*$ & 0.38$^*$ & - \\
    & KoNViD-1k & No &  $<$0.62$^*$ & \textbf{0.78}$^*$ & $<$0.49$^*$ &  -  \\
    & LIVE-Qualcomm & No & $<$0.36$^*$ & $<$0.38$^*$ & 0.788$^*$ &  -  \\ 
    & All three datasets & Linear re-scaling &  - & - & - & 0.77$^*$   \\ 
    \midrule
    \multirow{4}{*}{Our model} & CVD2014 & No &  \textbf{0.8747} & 0.6051 & 0.3919 & 0.6165   \\
    & KoNViD-1k & No &  0.6474 & \textbf{0.7809} & 0.6732 & 0.7483   \\
    & LIVE-Qualcomm & No &  0.5879 & 0.6128 & 0.7538 & 0.6271  \\ 
    & All three datasets & Our strategy & 0.8412 & 0.7659 & \textbf{0.8157} & \textbf{0.7829}  \\ 
    \bottomrule
    \end{tabular}
    
    \vspace{1mm}
    
    $^*$The reported SROCC results in the original paper~\citep{korhonen2019two} are shown here for reference. 
    
    The ``$<$" relation is inferred from the Table VII of ~\citet{korhonen2019two}. ``-" indicates that the results are not reported.
    \end{small}
\end{table*}

We further prove the above statement by conducting experiments to compare the models trained by mixing CVD2014, KoNViD-1k, and LIVE-Qualcomm datasets with the models trained on one of the datasets. 
Table~\ref{tab:vs individual} shows the median SROCC of different models on the three datasets.
We can see that, no matter which model it is, the unified model trained by mixing all datasets achieves better overall performance than the model trained on one of the datasets.
And our model trained with our proposed strategy achieves better overall performance across the datasets than the other models (VBLIINDS, BRISQUE, and TLVQM) trained with the linear re-scaling strategy. 
Among these datasets, the size of LIVE-Qualcomm dataset is the smallest one. And our model trained only on LIVE-Qualcomm dataset suffered from over-fitting problem. In such situation, mixed datasets training helps alleviating the problem to some extent. So a performance improvement of the proposed model with mixed dataset training is found on LIVE-Qualcomm dataset.
This verifies the necessity of mixed datasets training and the effectiveness of our mixed datasets training strategy.

\subsection{Computational efficiency}

\begin{figure}[!thb]
    \centering
    \subfloat[Video@resolution 640$\times$480, 364 frames]{\includegraphics[width=.95\columnwidth]{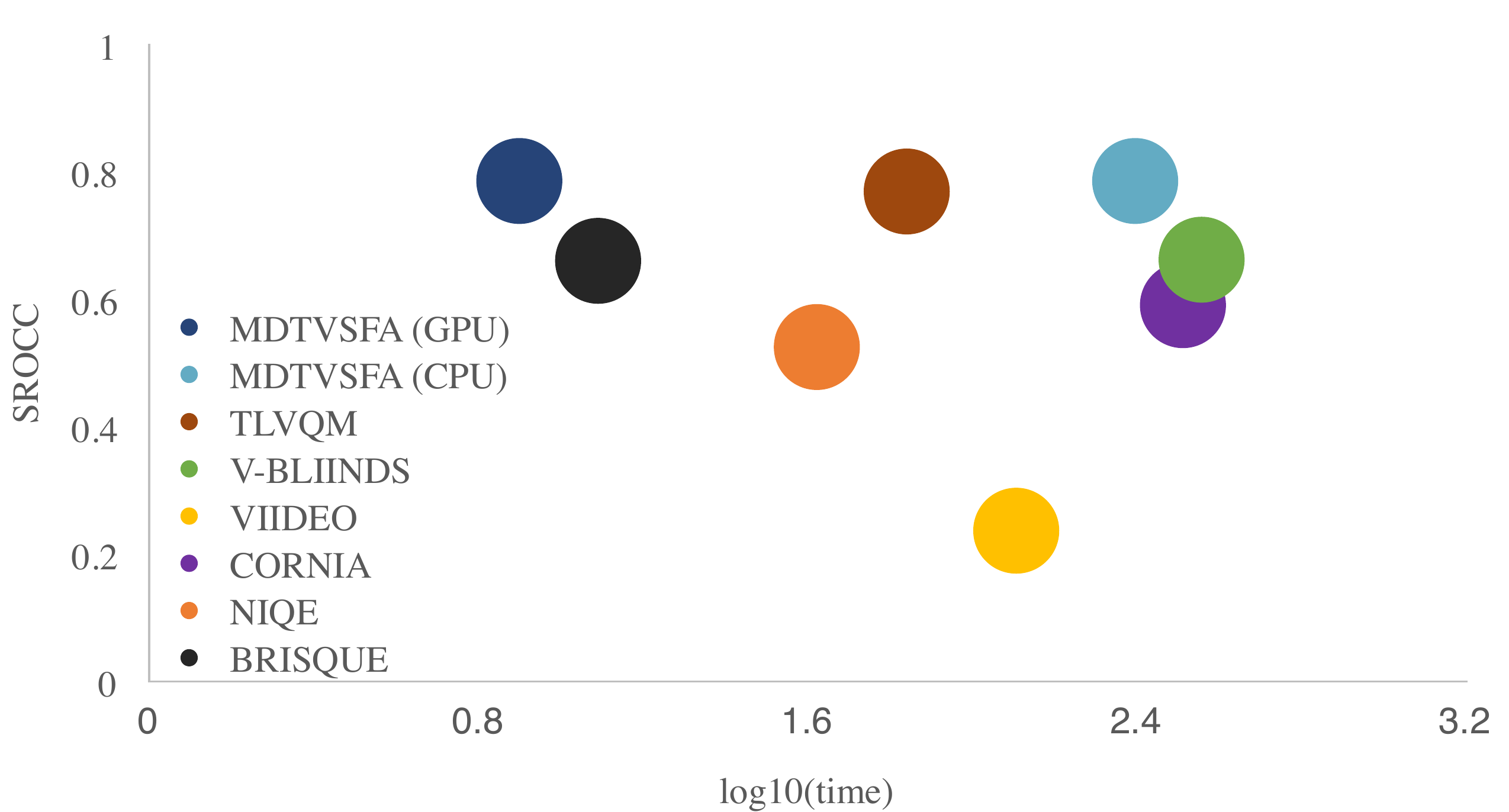}}
    
    \subfloat[Video@resolution 1280$\times$720, 467 frames]{\includegraphics[width=.95\columnwidth]{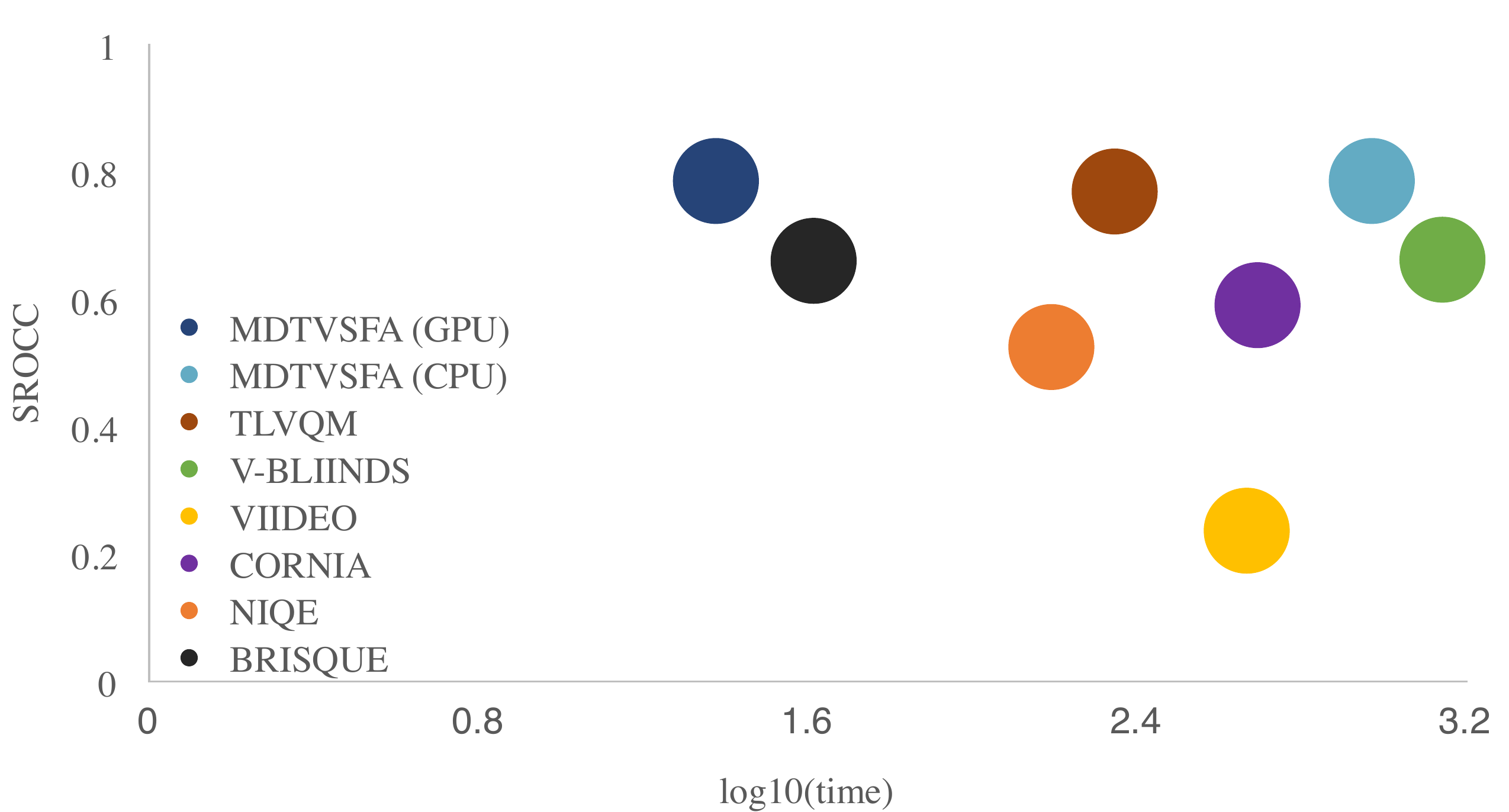}}
    \caption{Bubble charts with the overall performance (mean SROCC values in Table~\ref{tab:overall performance}) and the logarithm of average computation time (seconds) on videos with different resolutions and different lengths}
    \label{fig:time}
\end{figure}

Besides the performance, computational efficiency is also crucial for NR-VQA methods. 
To provide a fair comparison for the computational efficiency of different methods, all tests are carried out on the same desktop computer with Intel Core i7-6700K CPU@4.00 GHz, 12G NVIDIA TITAN Xp GPU, and 64 GB RAM. 
The operating system is Ubuntu 14.04. 
The compared methods are implemented with MATLAB R2016b while our method is implemented with Python 3.6. 
We use the default settings of the original codes without any modification. 
We select two videos with different lengths and different resolutions for testing. 
The tests are run in a separate environment and repeated ten times to avoid any influence. 
The logarithm (with base 10) of the average computation time (seconds) for each method is shown in Fig.~\ref{fig:time}. 
The point near the left is the fast one, and the point near the top is the good-performed one. 
Our method (CPU version) is faster than VBLIINDS---the method with the third-best performance. 
TLVQM, the second-best performed method, considers two-level features, \textit{i.e.}, low-complexity features for all frames and high-level features for only selected representative frames.  
It achieves a good trade-off between the performance and computational efficiency.
It is worth mentioning that our method can be accelerated to \textbf{30x faster or more} (The larger resolution and length the video has, the faster acceleration is) by simply switching the CPU mode to the GPU mode. 
With the GPU available, our method (GPU version) is at the upper-left, and thus it is the fastest one as well as the best-performed one.
To further improve the computational efficiency, we may resort to the light-weight networks.

\section{Conclusion and Future Work}
\label{sec:conclusion}
In this work, we propose a novel unified NR-VQA framework with a mixed datasets training strategy for in-the-wild videos. 
The backbone model is a deep neural network designed for characterizing the two eminent effects of HVS, \textit{i.e.}, content-dependency and temporal-memory effects. 
We enable mixed datasets training by designing two losses (monotonicity-induced loss, linearity-induced loss) for predicting relative quality and perceptual quality, and assigning dataset-specific perceptual scale alignment layers for predicting subjective quality.
Our proposed method is compared with the state-of-the-art methods on four publicly available in-the-wild VQA datasets (CVD2014, KoNViD-1k, LIVE-Qualcomm, and LIVE-VQC).
Experiments show the superior performance of our method and also verify the effectiveness of our unified VQA model with the mixed datasets training strategy.

However, our mixed datasets training strategy needs to re-train the unified VQA model every time when a new dataset is added to the training data. 
This will increase the burden of training. 
In the further study, we will explore lifelong learning for this task. 
Also, besides video capture, we intend to provide a unified and efficient VQA framework that can handle the whole chain-flow of video production. 
Moreover, some meta information that is crucial for the video quality, like video resolution, can be used as extra features for improving the model performance.
Finally, we intend to apply our unified VQA model for practical computer vision applications such as video enhancement.

\begin{acknowledgements}
This work was partially supported by the Natural Science Foundation of China under contracts 61572042,  61520106004,  and 61527804. 
This work was also supported in part by National Key R\&D Program of China (2018YFB1403900).
We acknowledge the High-Performance Computing Platform of Peking University for providing computational resources. 
\end{acknowledgements}

%
\section*{Conflict of interest}
The authors declare no conflict of interest.

\bibliographystyle{spbasic}      
\bibliography{MDTVSFA}   

\end{document}